\documentclass{article}
\usepackage{iclr2027_conference,times}

\usepackage{amsmath,amsfonts,bm}

\def\eqref#1{equation~\ref{#1}}

\def\1{\bm{1}}

\DeclareMathAlphabet{\mathsfit}{\encodingdefault}{\sfdefault}{m}{sl}
\SetMathAlphabet{\mathsfit}{bold}{\encodingdefault}{\sfdefault}{bx}{n}

\usepackage{hyperref}
\hypersetup{pdftitle={LADDER: A Long-Horizon Robot Manipulation Benchmark},
  pdfauthor={Petko Petkov, Giuliano Albanese, Nikolay Nikolov, Luc Van Gool, Jonathan Tremblay, Danda Pani Paudel}}
\usepackage{url}
\usepackage{graphicx}
\usepackage{booktabs}
\usepackage{longtable}
\usepackage{array}
\usepackage{amsmath,amssymb}
\usepackage{xcolor}
\usepackage{subcaption}
\usepackage{multirow}
\usepackage{microtype}
\usepackage[section]{placeins}
\usepackage{float}
\usepackage{tcolorbox}
\tcbuselibrary{breakable}
\newtcolorbox{promptbox}{breakable, colback=black!4, colframe=black!40, boxrule=0.4pt, arc=1pt, left=4pt, right=4pt, top=2pt, bottom=2pt, fontupper=\small, before skip=3pt, after skip=6pt}
\ifdefined\LaTeXML\renewenvironment{promptbox}{\begin{quote}\small}{\end{quote}}\fi
\usepackage{tikz}
\usetikzlibrary{arrows.meta,positioning,calc,fit,backgrounds}

\DeclareRobustCommand{\benchname}{LADDER}
\DeclareRobustCommand{\robotarena}{\mbox{RobotArena-$\infty$}}
\DeclareRobustCommand{\capx}{\mbox{CaP-X}}
\DeclareRobustCommand{\tiptop}{TiPToP}
\DeclareRobustCommand{\harness}{\mbox{PSP-Agent}}
\DeclareRobustCommand{\gptpolicy}{GPT-Policy}
\DeclareRobustCommand{\robolab}{RoboLab}

\DeclareRobustCommand{\pifive}{$\pi_{0.5}$}
\newcommand{\smallpm}[1]{{\scriptsize$\pm$#1}}
\definecolor{bestlearned}{RGB}{0,128,0}
\definecolor{bestscripted}{RGB}{0,80,200}

\newcommand{\periodicStepsBridge}{27}
\newcommand{\periodicStepsRobolab}{250}
\newcommand{\budgetStepsPerGoal}{100}
\newcommand{\budgetRobolabMin}{900}
\newcommand{\budgetRobolabMax}{4950}
\newcommand{\shareSixPlus}{36}
\newcommand{\tasksIncremental}{138}
\newcommand{\tasksTotalAll}{145}
\newcommand{\taskTypeCount}{8}
\newcommand{\tasksSixPlus}{52}
\newcommand{\episodesPerEval}{2{,}024}
\newcommand{\tasksSimplerAll}{54}
\newcommand{\tasksRaAll}{47}
\newcommand{\tasksRobolabAll}{44}
\newcommand{\medianOurs}{4.7}
\newcommand{\cmpRobolabMedian}{1.9}
\newcommand{\cmpRobolabSixPlus}{4}
\newcommand{\cmpVoloMedian}{2.0}
\newcommand{\cmpVoloSixPlus}{2}
\newcommand{\cmpLiberoMedian}{1.6}
\newcommand{\cmpLiberoSixPlus}{0}
\newcommand{\gridDirectSimpler}{8.7}
\newcommand{\gridDirectRa}{10.9}
\newcommand{\gridDirectRobolab}{19.6}
\newcommand{\gridVoloSimpler}{25.5}
\newcommand{\gridCapxSimpler}{28.7}
\newcommand{\gridCapxRa}{16.2}
\newcommand{\gridCapxRobolab}{20.7}
\newcommand{\gridHarnessSimpler}{37.1}
\newcommand{\gridHarnessRa}{39.9}
\newcommand{\gridHarnessRobolab}{51.4}
\newcommand{\gridGptSimpler}{60.3}
\newcommand{\gridGptRa}{66.8}
\newcommand{\gridGptRobolab}{62.8}
\newcommand{\gridVoloRa}{21.3}
\newcommand{\gridVoloRobolab}{29.0}
\newcommand{\gridOursSimpler}{25.2}
\newcommand{\gridOursRa}{24.2}
\newcommand{\gridOursRobolab}{25.6}
\newcommand{\gridOracleSimpler}{29.5}
\newcommand{\gridOracleRa}{25.5}
\newcommand{\gridOracleRobolab}{34.4}
\newcommand{\gridCapxAll}{25.0}
\newcommand{\gridTiptopAll}{9.4}
\newcommand{\gridHarnessAll}{40.1}
\newcommand{\gridGptAll}{62.0}
\newcommand{\gridOracleAll}{29.6}
\newcommand{\budgetSecondsPerGoal}{20}
\newcommand{\budgetRobolabMinS}{60}
\newcommand{\budgetRobolabMaxS}{330}
\newcommand{\decompBenchStateN}{562}
\newcommand{\effHalfVerifier}{42.1}
\newcommand{\effEndVerifier}{49.5}
\newcommand{\effUsdVerifier}{0.045}
\newcommand{\effUsdVolo}{0.17}
\newcommand{\effUsdCapx}{0.043}
\newcommand{\effUsdTiptop}{0.007}
\newcommand{\effUsdHarness}{0.13}
\newcommand{\effUsdGpt}{4.02}
\newcommand{\effUsdSuccHarness}{0.33}
\newcommand{\effUsdGptTotal}{8{,}134}
\newcommand{\effUsdSuccGpt}{6.49}
\newcommand{\wallMsRtxLo}{51}
\newcommand{\wallMsRtxHi}{56}
\newcommand{\wallMsAhLo}{130}
\newcommand{\wallMsAhHi}{177}
\newcommand{\bootDiffGptHarness}{21.8 [18.1, 25.8]}
\newcommand{\bootDiffHarnessOracle}{10.5 [7.4, 13.7]}
\newcommand{\bootDiffOracleVer}{4.5 [2.6, 6.6]}
\newcommand{\bootDiffOracleVolo}{4.2 [2.3, 6.3]}
\newcommand{\bootDiffVerTrigger}{3.4 [1.4, 5.3]}
\newcommand{\bootDiffVerFlat}{14.0 [9.8, 18.7]}
\newcommand{\bootDiffVoloVer}{0.3 [$-$1.8, 2.6]}
\newcommand{\bootDiffCapxVer}{0.0 [$-$3.8, 3.6]}
\newcommand{\bootDiffFlatTiptop}{1.6 [$-$1.3, 4.7]}
\newcommand{\stallLastMedian}{28}
\newcommand{\tiptopNoPlanOne}{5.2}
\newcommand{\tiptopNoPlanFive}{74.0}
\newcommand{\successBudgetLo}{36}
\newcommand{\successBudgetHi}{45}
\newcommand{\tasksOrderDependent}{47}
\newcommand{\dupTasks}{20}
\newcommand{\dupTasksVolo}{11}
\newcommand{\latestRlVoloEarly}{31.0}
\newcommand{\decompBenchMeanFlash}{80.8}
\newcommand{\tokCoverMin}{98.6}
\newcommand{\decompBenchTrackFlash}{70.8}
\newcommand{\decompBenchExactFlash}{93.1}
\newcommand{\compEpisodesTotal}{21{,}296}
\newcommand{\compEpisodesCosmos}{1{,}056}
\newcommand{\latestRlFlat}{19.6}
\newcommand{\latestRlOracle}{34.4}
\newcommand{\latestRlVolo}{29.0}
\newcommand{\latestRlCosmos}{11.1}
\newcommand{\latestRlCosmosVolo}{25.6}
\newcommand{\latestRlCosmosOracle}{22.2}
\newcommand{\realFlatSingle}{66.7}
\newcommand{\realFlatMulti}{1.2}
\newcommand{\realVerMulti}{31.0}
\newcommand{\frankaFlatSuccess}{5.6}
\newcommand{\frankaVerSuccess}{58.3}
\newcommand{\faFailures}{14{,}824}
\newcommand{\catRevisitAll}{2.3}
\newcommand{\introFlatOne}{85.4}
\newcommand{\introVerOne}{80.2}
\newcommand{\introFlatTwo}{26.9}
\newcommand{\introFlatFive}{0.0}
\newcommand{\introVerTwo}{58.4}
\newcommand{\introVerFive}{8.3}
\newcommand{\introHarnessFive}{27.3}
\newcommand{\introOracleFive}{11.1}
\newcommand{\introFlatLong}{0.5}
\newcommand{\introFlatRa}{5.2}
\newcommand{\introFlatRl}{15.8}
\newcommand{\introVerSimpler}{22.7}
\newcommand{\introVerRa}{20.6}
\newcommand{\faRaSparseFrom}{7}
\newcommand{\faRaSparseTasks}{one or two}
\newcommand{\faGeomAgreement}{73.0}
\newcommand{\faGeomFalsePos}{11.1}
\newcommand{\faGeomFalseNeg}{15.9}
\newcommand{\faGeomGoals}{504}
\newcommand{\faConstraint}{2.5}
\newcommand{\faVerPrecSe}{89.9}
\newcommand{\faVerRecSe}{62.4}
\newcommand{\faVerPrecRa}{88.5}
\newcommand{\faVerRecRa}{56.5}
\newcommand{\faVerPrecRl}{84.0}
\newcommand{\faVerRecRl}{62.7}
\newcommand{\faVerifierPrecision}{88.6}
\newcommand{\faVerifierRecall}{61.8}
\newcommand{\faVerifierCalls}{23{,}145}
\newcommand{\faConstraintOrdered}{32.3}
\newcommand{\fateNoPlanTiptop}{50.9}
\newcommand{\decMovedPeriodic}{43.4}
\newcommand{\decMovedTrigger}{32.6}
\newcommand{\decMovedHarness}{16.2}
\newcommand{\decMovedVerifier}{3.4}
\newcommand{\decPlanCoverMin}{98.6}
\newcommand{\decAllVerifier}{7.1}
\newcommand{\decStuckVerifierRa}{4.4}
\newcommand{\decAllVolo}{13.8}
\newcommand{\fateNoPlanTiptopSe}{66.6}
\newcommand{\fateLostVerifierRa}{66.3}
\newcommand{\fateLostGptRa}{16.6}
\newcommand{\fateLostHarnessRa}{49.8}
\newcommand{\fateMissFlatRa}{10.2}
\newcommand{\fateNeverCmdRa}{38.9}
\newcommand{\fateLostHarnessRl}{18.4}
\newcommand{\fateMissCapxRl}{2.3}
\newcommand{\fateWrongFlatRl}{15.4}
\newcommand{\fateRegraspOrchRl}{61.8}
\newcommand{\fateRegraspFlatRl}{10.5}
\newcommand{\fateOutOfTimeCapxRl}{35.0}
\newcommand{\fateSlipCosmosRl}{9.5}
\newcommand{\fateSlipPiRl}{1.7}
\newcommand{\fateNearMissWx}{60.4}
\newcommand{\fateWrongDestPi}{90.8}
\newcommand{\fateWrongDestHarness}{13.3}

\title{\centering\benchname{}: A Long-Horizon\\Robot Manipulation Benchmark}

\author{\makebox[\dimexpr\textwidth-2\tabcolsep\relax]{\begin{tabular}[t]{@{}c@{}}\bf Petko Petkov$^{1}$ \quad Giuliano Albanese$^{1}$ \quad Nikolay Nikolov$^{1}$ \\
\textbf{Luc Van Gool$^{1}$ \quad Jonathan Tremblay$^{2,\dagger}$ \quad Danda Pani Paudel$^{1,\dagger}$} \\
\mdseries $^{1}$INSAIT, Sofia University ``St.\ Kliment Ohridski'' \quad $^{2}$NVIDIA \quad $^{\dagger}$Equal supervision
\end{tabular}}}

\iclrfinalcopy

\begin{document}

\maketitle
\lhead{Preprint}
\vspace*{-6pt}

\providecommand{\cardsper}{2}%
\begin{figure}[h!]
\centering
\definecolor{cSuiteA}{HTML}{0072B2}
\definecolor{cSuiteB}{HTML}{CC79A7}
\definecolor{cSuiteC}{HTML}{009E73}
\definecolor{cFlat}{HTML}{999999}
\definecolor{cVolo}{HTML}{56B4E9}
\definecolor{cCapx}{HTML}{0072B2}
\definecolor{cTiptop}{HTML}{CC79A7}
\definecolor{cHarness}{HTML}{CC79A7}
\definecolor{cOurs}{HTML}{D55E00}
\definecolor{cOracle}{HTML}{009E73}
\definecolor{cGpt}{HTML}{000000}
\newcommand{\band}[4]{%
  \fill[black!8, rounded corners=1.6pt] (#1,#2) rectangle ++(#3,-0.40);
  \node[anchor=west, font=\footnotesize\bfseries, text=black!80] at (#1+0.13,#2-0.205) {#4};}
\newcommand{\hbars}[3]{%
  \pgfmathsetmacro{\gx}{6.58+#1*2.43}%
  \foreach \col/\v [count=\j from 0] in {#3}{%
    \pgfmathsetmacro{\bx}{\gx+\j*0.33}%
    \fill[\col] (\bx,\zero) rectangle ++(0.295,{\v*\uy});
    \node[anchor=south, font=\fontsize{4.6}{4.6}\selectfont, text=black!70, inner sep=0.8pt]
      at (\bx+0.1475,{\zero+\v*\uy}) {\v};}%
  \node[anchor=north, tiny, text=black!70] at (\gx+1.1375,\zero-0.05) {#2};}
\newcommand{\stat}[3]{%
  \node[anchor=base, font=\large\bfseries, text=black!85] at (#1,#2) {#3};}
\ifdefined\LaTeXML\includegraphics[width=\linewidth]{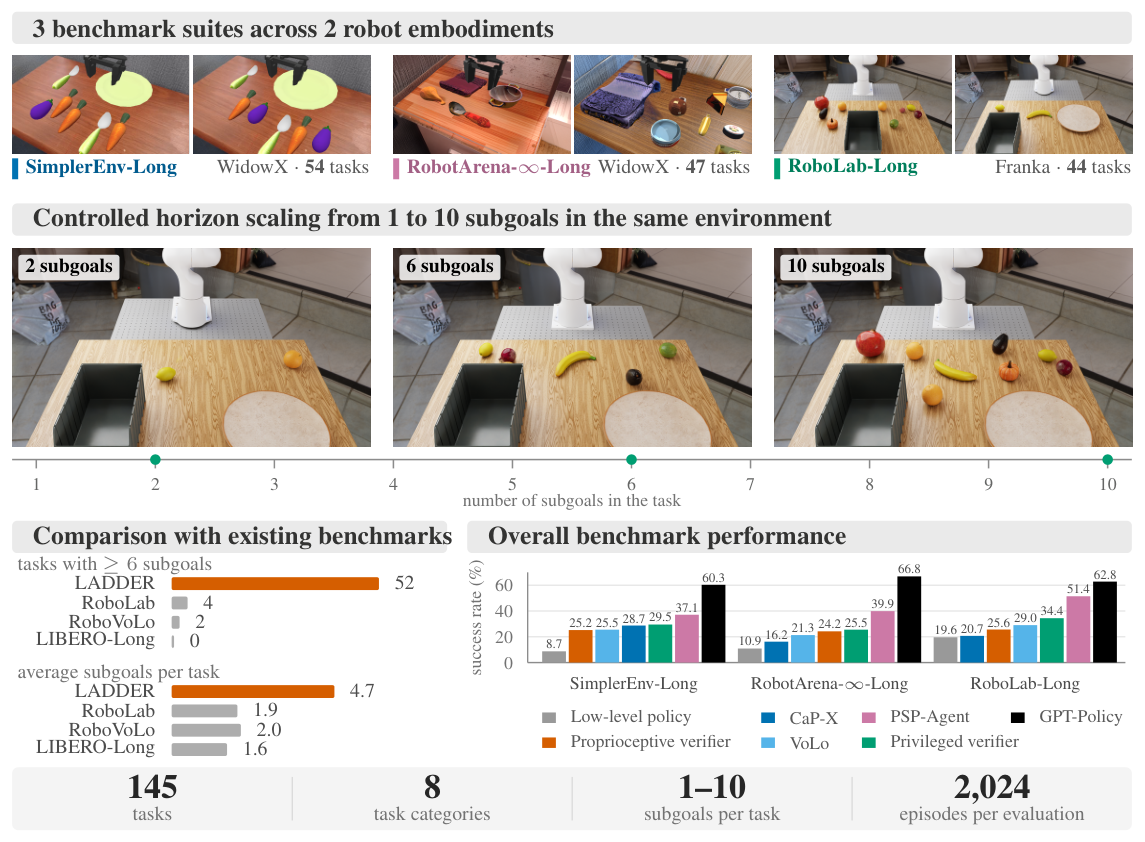}\else\begin{tikzpicture}[scale=0.93, transform shape, x=1cm, y=1cm, font=\small,
  ph/.style   = {draw=black!30, line width=0.3pt, inner sep=0pt},
  gl/.style   = {text=black!55, font=\scriptsize},
  tiny/.style = {font=\fontsize{6.3}{6.9}\selectfont},
  chip/.style = {rounded corners=2pt, draw=cOurs!55, fill=cOurs!8, inner xsep=2.6pt,
                 inner ysep=1.5pt, font=\fontsize{6.3}{6.6}\selectfont, text=black!70},
]
\ifnum\cardsper=2 \useasboundingbox (0,0.06) rectangle (13.90,-10.22);
\else \useasboundingbox (0,0.06) rectangle (13.90,-11.46); \fi

\ifnum\cardsper=2 \def\rowAdrop{1.24}\else \def\rowAdrop{0}\fi
\begin{scope}[yshift=1.00cm]
\band{0}{-1.00}{13.9}{3 benchmark suites across 2 robot embodiments}

\foreach \i/\name/\col/\imga/\imgb/\imgc/\imgd/\rob/\nt/\cat in {%
  0/SimplerEnv-Long/cSuiteA/card_simpler/sceneA2/sceneA3/sceneA4/WidowX/\tasksSimplerAll/{exclusion},
  1/{RobotArena-$\infty$-Long}/cSuiteB/card_ra/sceneB2/sceneB3/sceneB4/WidowX/\tasksRaAll/{transfer},
  2/RoboLab-Long/cSuiteC/card_robolab/sceneC2/sceneC3/sceneC4/Franka/\tasksRobolabAll/{single destination}}{
  \begin{scope}[shift={(\i*4.73,0)}]
    \ifnum\cardsper=1
      \node[ph, anchor=north west] at (0,-1.54)
        {\ifcase\i\relax\includegraphics[width=4.44cm]{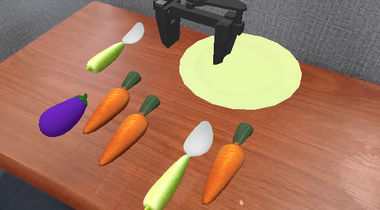}\or\includegraphics[width=4.44cm]{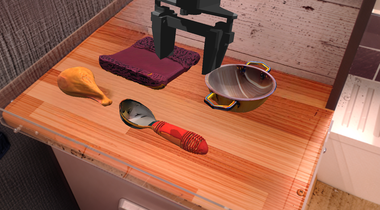}\or\includegraphics[width=4.44cm]{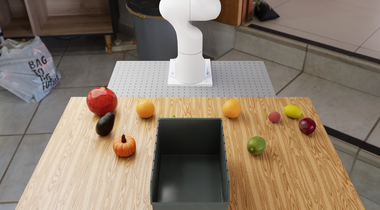}\fi};
      \node[anchor=south west, fill=white, fill opacity=0.85, text opacity=1, inner xsep=2.4pt,
            inner ysep=1.4pt, font=\fontsize{6.3}{6.6}\selectfont, text=black!70, rounded corners=1pt]
        at (0.07,-3.92) {\cat};
    \fi
    \ifnum\cardsper=2
      \node[ph, anchor=north west] at (0,-1.54)
        {\ifcase\i\relax\includegraphics[width=2.19cm]{figures/panels/card_simpler.png}\or\includegraphics[width=2.19cm]{figures/panels/card_ra.png}\or\includegraphics[width=2.19cm]{figures/panels/card_robolab.png}\fi};
      \node[ph, anchor=north west] at (2.25,-1.54)
        {\ifcase\i\relax\includegraphics[width=2.19cm]{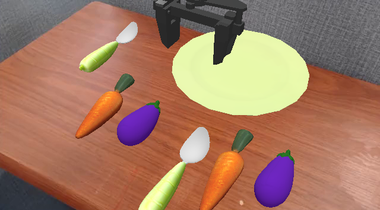}\or\includegraphics[width=2.19cm]{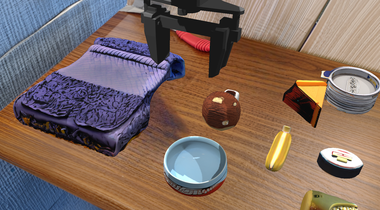}\or\includegraphics[width=2.19cm]{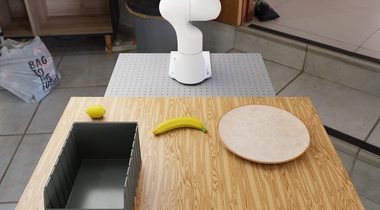}\fi};
    \fi
    \fill[\col] (0,{-4.06+\rowAdrop}) rectangle ++(0.075,-0.26);
    \node[anchor=north west, font=\scriptsize\bfseries, text=\col!80!black, inner sep=0pt]
      at (0.16,{-4.06+\rowAdrop}) {\name};
    \node[anchor=north east, font=\scriptsize, text=black!65, inner sep=0pt]
      at (4.44,{-4.06+\rowAdrop}) {\rob{} $\cdot$ \textbf{\nt{}} tasks};
  \end{scope}}

\begin{scope}[yshift=\rowAdrop cm]
\band{0}{-4.62}{13.9}{Controlled horizon scaling from 1 to 10 subgoals in the same environment}

\foreach \i/\g/\f in {0/2/ladder_g2, 1/6/ladder_g6, 2/10/ladder_g10}{
  \begin{scope}[shift={(\i*4.73,0)}]
    \node[ph, anchor=north west] at (0,-5.18)
      {\ifcase\i\relax\includegraphics[width=4.44cm]{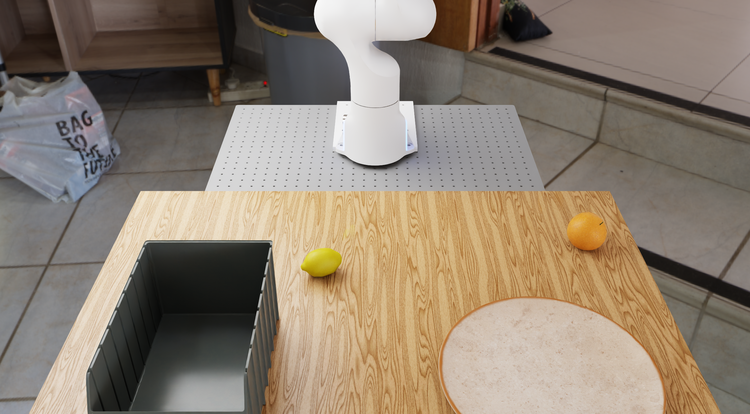}\or\includegraphics[width=4.44cm]{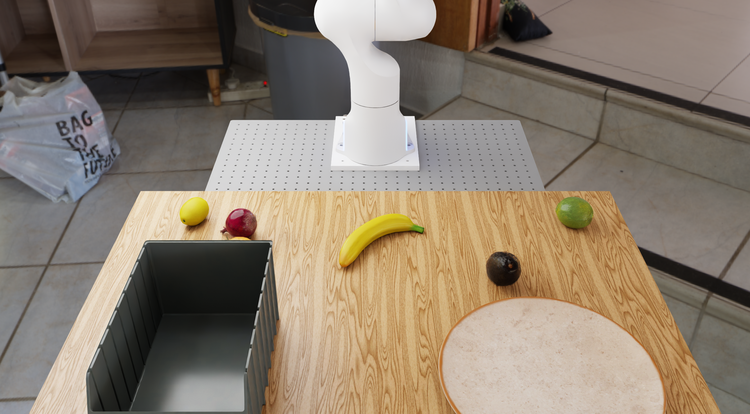}\or\includegraphics[width=4.44cm]{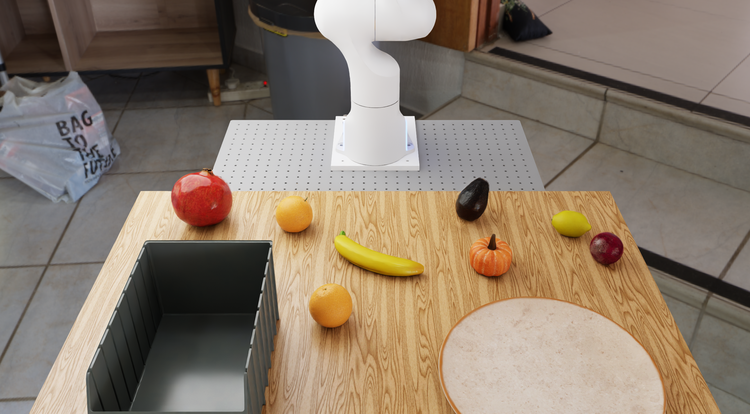}\fi};
    \node[anchor=north west, fill=white, fill opacity=0.85, text opacity=1, inner xsep=2.4pt,
          inner ysep=1.4pt, font=\scriptsize\bfseries, rounded corners=1pt]
      at (0.07,-5.25) {\g{} subgoals};
  \end{scope}}

\draw[black!45, line width=0.5pt] (0,-7.80) -- (13.90,-7.80);
\foreach \g in {1,...,10}{
  \pgfmathsetmacro{\xx}{0.30+(\g-1)*1.4778}
  \draw[black!45, line width=0.5pt] (\xx,-7.80) -- (\xx,-7.91);
  \node[anchor=north, tiny, text=black!60] at (\xx,-7.90) {\g};}
\foreach \g in {2, 6, 10}{
  \pgfmathsetmacro{\xx}{0.30+(\g-1)*1.4778}
  \fill[cSuiteC] (\xx,-7.80) circle (0.065);}
\node[anchor=north, tiny, text=black!55] at (6.95,-8.10) {number of subgoals in the task};

\band{0}{-8.56}{5.40}{Comparison with existing benchmarks}
\node[anchor=north west, gl, inner sep=0pt] at (0.06,-9.00) {tasks with $\geq 6$ subgoals};
\foreach \i/\lbl/\v/\c in {0/\benchname/\tasksSixPlus/cOurs, 1/RoboLab/\cmpRobolabSixPlus/black!32,
                           2/RoboVoLo/\cmpVoloSixPlus/black!32,
                           3/LIBERO-Long/\cmpLiberoSixPlus/black!32}{
  \pgfmathsetmacro{\yy}{-9.42-\i*0.24}
  \node[anchor=east, font=\scriptsize, text=black!70] at (1.90,\yy+0.09) {\lbl};
  \pgfmathsetmacro{\w}{max(\v*0.0495,0.03)}
  \fill[\c, rounded corners=0.6pt] (1.98,\yy) rectangle ++(\w,0.16);
  \node[anchor=west, font=\scriptsize, text=black!70] at (1.98+\w+0.07,\yy+0.093) {\v};}
\node[anchor=north west, gl, inner sep=0pt] at (0.06,-10.34) {average subgoals per task};
\foreach \i/\lbl/\v/\c in {0/\benchname/\medianOurs/cOurs, 1/RoboLab/\cmpRobolabMedian/black!32,
                           2/RoboVoLo/\cmpVoloMedian/black!32,
                           3/LIBERO-Long/\cmpLiberoMedian/black!32}{
  \pgfmathsetmacro{\yy}{-10.76-\i*0.24}
  \node[anchor=east, font=\scriptsize, text=black!70] at (1.90,\yy+0.09) {\lbl};
  \pgfmathsetmacro{\w}{max(\v*0.4300,0.03)}
  \fill[\c, rounded corners=0.6pt] (1.98,\yy) rectangle ++(\w,0.16);
  \node[anchor=west, font=\scriptsize, text=black!70] at (1.98+\w+0.07,\yy+0.093) {\v};}

\band{5.65}{-8.56}{8.25}{Overall benchmark performance}

\pgfmathsetmacro{\zero}{-10.32}
\pgfmathsetmacro{\uy}{0.0160}
\foreach \t in {0,20,40,60}{
  \pgfmathsetmacro{\ty}{\zero+\t*\uy}
  \draw[black!12, line width=0.35pt] (6.40,\ty) -- (13.82,\ty);
  \node[anchor=east, tiny, text=black!55] at (6.34,\ty) {\t};}
\draw[black!45, line width=0.45pt] (6.40,\zero) -- (6.40,{\zero+70*\uy});
\draw[black!45, line width=0.45pt] (6.40,\zero) -- (13.82,\zero);
\node[anchor=south, tiny, text=black!55, rotate=90] at (5.98,{\zero+35*\uy}) {success rate (\%)};

\hbars{0}{SimplerEnv-Long}{cFlat/\gridDirectSimpler,
  cOurs/\gridOursSimpler, cVolo/\gridVoloSimpler, cCapx/\gridCapxSimpler,
  cOracle/\gridOracleSimpler, cHarness/\gridHarnessSimpler, cGpt/\gridGptSimpler}
\hbars{1}{RobotArena-$\infty$-Long}{cFlat/\gridDirectRa,
  cCapx/\gridCapxRa, cVolo/\gridVoloRa, cOurs/\gridOursRa, cOracle/\gridOracleRa,
  cHarness/\gridHarnessRa, cGpt/\gridGptRa}
\hbars{2}{RoboLab-Long}{cFlat/\gridDirectRobolab,
  cCapx/\gridCapxRobolab, cOurs/\gridOursRobolab, cVolo/\gridVoloRobolab,
  cOracle/\gridOracleRobolab, cHarness/\gridHarnessRobolab, cGpt/\gridGptRobolab}

\foreach \lx/\ly/\lbl/\col in {6.58/-11.07/{Low-level policy}/cFlat,
                          9.30/-11.07/{\capx{}}/cCapx, 10.55/-11.07/{\harness{}}/cHarness,
                          12.40/-11.07/{\gptpolicy{}}/cGpt,
                          6.58/-11.38/{Proprioceptive verifier}/cOurs,
                          9.30/-11.38/{VoLo}/cVolo, 10.55/-11.38/{Privileged verifier}/cOracle}{
  \fill[\col] (\lx,\ly) rectangle ++(0.17,0.13);
  \node[anchor=west, tiny, text=black!65] at (\lx+0.23,\ly+0.065) {\lbl};}

\end{scope}
\end{scope}

\begin{scope}[yshift=\rowAdrop cm]
\fill[black!4, rounded corners=2pt] (0,-10.62) rectangle (13.90,-11.40);
\foreach \i/\num/\lbl in {0/\tasksTotalAll/{tasks},
                          1/\taskTypeCount/{task categories},
                          2/{1--10}/{subgoals per task},
                          3/\episodesPerEval/{episodes per evaluation}}{
  \pgfmathsetmacro{\cx}{1.74+\i*3.475}
  \node[anchor=base, font=\large\bfseries, text=black!85] at (\cx,-11.00) {\num};
  \node[anchor=base, gl] at (\cx,-11.28) {\lbl};
  \ifnum\i<3 \draw[black!15, line width=0.4pt] (\cx+1.7375,-10.74) -- (\cx+1.7375,-11.28); \fi}

\end{scope}
\end{tikzpicture}\fi
\caption{\textbf{Overview of our benchmark and key findings.} 
Across all suites, orchestrating policy execution with sequential subgoal decomposition and visual verification substantially outperforms issuing the full instruction directly to the policy (bottom right). See Tables~\ref{tab:benchmarks} and~\ref{tab:comp-overall}.}
\label{fig:hero}
\end{figure}

\begin{abstract}
General-purpose robot manipulation policies learn short-horizon skills effectively, but degrade rapidly on long-horizon tasks.
Existing benchmarks cannot isolate failures because task horizon is either too short or entangled with scene and object variations.
We propose \benchname{}, a long-horizon robot manipulation benchmark scaling from 1 to 10 subgoals within the same environment over three suites (SimplerEnv-Long, \robotarena{}-Long, and RoboLab-Long), \tasksTotalAll{} tasks, eight categories, two embodiments, and \episodesPerEval{} simulated episodes per system. 
We evaluate two orchestration paradigms: text subgoals for learned low-level policies (VoLo) and tool selection orchestration (\capx{}, \gptpolicy{}, \harness{}, \tiptop{}). 
Given the full instruction directly, low-level policies achieve \introFlatOne{}\% on single-subgoal tasks, but drop to near zero on extended tasks.
While orchestration remedies this, it remains bounded by physical execution errors of the low-level policy rather than high-level planning and achieves only $\sim$25\%. 
Tool selection orchestration with \gptpolicy{} on GPT-6 Astra significantly outperforms learned-policy methods, but incurs significant time and cost budget and succeeds only $\sim$62\%.
Real-world experiments across two embodiments confirm the collapse of the low-level policy and the gain from orchestration.
Project website: \href{https://ladder.insait.ai}{ladder.insait.ai}
\end{abstract}

\section{Introduction}
\label{sec:intro}

General-purpose robot manipulation policies learn diverse skills from large datasets of teleoperated demonstrations~\citep{brohan2023rt2,walke2023bridgedata,khazatsky2024droid,intelligence2025pi05,cosmos3}. However, these demonstrations predominantly cover short-horizon, single-subgoal tasks. Because collecting demonstrations for every multi-subgoal combination is infeasible~\citep{mees2022calvin,liu2023libero}, generalist robot models rely on \emph{zero-shot skill composition}: using low-level policies trained on single-subgoal tasks to solve long-horizon tasks without task-specific fine-tuning.

When presented directly with full instructions involving multiple subgoals or conditional constraints, current policies degrade rapidly. Existing benchmarks offer limited ability to isolate why this failure occurs: suites such as CALVIN~\citep{mees2022calvin} and LIBERO-Long~\citep{liu2023libero} evaluate short horizons and often require in-domain training rather than testing zero-shot composition. Recent multi-task evaluations such as RoboLab~\citep{yang2026robolab} scale task complexity, but conflate task horizon with scene clutter and instruction phrasing. Systematically isolating long-horizon composition requires evaluating low-level policies zero-shot across extended horizons while holding the environment and the object sizes fixed.

In this paper, we introduce \benchname{}, a benchmark for long-horizon robot manipulation composed of three evaluation suites: SimplerEnv-Long, \robotarena{}-Long, and RoboLab-Long. Because evaluating thousands of multi-subgoal trials on physical hardware is infeasible at scale, we use physics-based simulation environments that closely match real setups and have demonstrated strong sim-to-real correlation~\citep{li2024simplerenv,jangir2025robotarena}. Rather than inflating task horizon through unstructured scene changes, our benchmark structures long-horizon manipulation across \textbf{eight spatial and logical task categories} (Figure~\ref{fig:hero}). Across all suites, task horizon scales systematically from 1 to 10 subgoals within the same environment, using explicit programmatic predicates to measure intermediate subgoal progress alongside full task completion.

The benchmark evaluates \tasksTotalAll{} tasks across two robot embodiments (WidowX and Franka) over \episodesPerEval{} simulated episodes per system. On single-subgoal control tasks, low-level policies achieve a high success rate: \introFlatOne{}\% in simulation and \realFlatSingle{}\% on the real WidowX. However, when given multi-subgoal tasks (2 to 10 subgoals), the low-level policy collapses, falling to \introFlatRa\% to \introFlatRl\% in simulation (\introFlatLong\% at 8 to 10 subgoals) and, in real-world experiments, to \realFlatMulti{}\% on WidowX and \frankaFlatSuccess{}\% on Franka.

To bridge the long-horizon gap, recent frameworks deploy high-level orchestrators across distinct execution backends: decomposing instructions into text subgoals for learned policies~\citep{chen2026volo}, synthesizing executable code over motion primitives~\citep{fu2026capx}, or generating symbolic plans for task-and-motion planners~\citep{shen2026tiptop}.  Although not specifically designed for long-horizon tasks, GPT-Policy~\citep{cheng2026incontextrobotlearningvlm} uses GPT-6 Astra to invoke robot tools and achieves the strongest performance on \benchname{}.
We therefore systematically study the orchestrator design space, characterizing four modular mechanisms: subgoal planning, execution triggering, visual verification, and failure recovery. Orchestration enables low-level policies to execute tasks up to 10 subgoals without fine-tuning, improving multi-subgoal success to \introVerRa\% to \introVerSimpler\% in simulation and \realVerMulti{}\% to \frankaVerSuccess\% in real-world experiments. 
Tool selection systems, which call motion primitives instead of a learned policy, go further. In all cases, \benchname{} remains challenging despite some methods spending impractical budgets. 
However, decomposition alone is insufficient: low-level policies frequently fail on individual subgoals. Offline, vision-language models decompose instructions accurately (up to \decompBenchExactFlash\%), but struggle with scene state tracking (at most \decompBenchTrackFlash\%).

Our contributions are as follows:
\begin{itemize}
    \item A long-horizon manipulation benchmark across three suites, spanning 1 to 10 subgoals and eight structured task categories with programmatic progress predicates (\tasksTotalAll{} tasks, \episodesPerEval{} simulated episodes per system).
    \item A systematic study of hierarchical orchestration over frozen low-level policies, characterizing the modular roles of planning, triggering, visual verification, and failure recovery, evaluated alongside representative alternative paradigms that achieve higher success at a higher cost.
    \item Real-world experiments across two robot platforms (WidowX and Franka) demonstrating that orchestration improves multi-subgoal execution.
\end{itemize}

\section{Related Work}
\label{sec:related}

\textbf{Manipulation benchmarks.} 
Existing robot manipulation benchmarks provide limited means to study how task horizon affects the performance of a system (see Table~\ref{tab:benchmarks} for an overview). 
CALVIN \citep{mees2022calvin} and LIBERO-Long \citep{liu2023libero} include multi-step tasks, but their standard evaluation protocol for learned policies involves fine-tuning on in-domain demonstrations. 
Their simulated environments also have limited alignment with real-world robot appearance and dynamics, so success in these benchmarks does not reliably transfer to the real-world. 
SimplerEnv \citep{li2024simplerenv} and \robotarena{} \citep{jangir2025robotarena} reduce this gap by matching simulated scenes and robot dynamics to real data, while RoboLab \citep{yang2026robolab} uses photorealistic simulation. 
However, the first two focus primarily on single-step tasks, while RoboLab and its long-horizon extension RoboVoLo \citep{chen2026volo} vary task horizon along other axes of complexity, such as scene background and out of distribution objects.
In contrast, building on SimplerEnv, \robotarena{}, and RoboLab, we propose a benchmark that isolates task horizon scaling by systematically increasing the number of subgoals of a task while holding the environment and underlying skill fixed.\\
\textbf{Hierarchical orchestrators.} 
Recent advances in the visual and semantic reasoning capabilities of foundation models have motivated hierarchical orchestration as an approach to long-horizon manipulation. Orchestration systems use frontier multi-modal models to break an instruction into subgoals, and delegate their execution to a robot-control backend. 
Prior work has used code over motion primitives \citep{liang2023codeaspolicies,fu2026capx}, task and motion planning (TAMP) \citep{shen2026tiptop}, and learned skills or policies \citep{ahn2022saycan,huang2022innermonologue,belkhale2024rth,shi2025hirobot,li2025hamster,chen2026volo} as backends.
However, it remains unclear how different design choices of these systems affect performance as task horizon systematically increases.

\begin{table}[!ht]
\centering
\setlength{\tabcolsep}{4pt}
\caption{\textbf{Comparison with existing manipulation benchmarks.} \emph{Sim-to-Real} indicates whether policy performance in the simulator transfers to or correlates with physical robots. CALVIN chains five instructions but issues each one separately, so each of its tasks has one subgoal. \emph{Avg Subgoals} is the average number of object placements per task. \emph{Incremental Subgoal Tasks} counts tasks that extend a shorter task of the same scene with more subgoals, or are extended by a longer one, to isolate the effect of task horizon.}
\label{tab:benchmarks}
\scriptsize
\begin{tabular}{l c c c c c c}
\toprule
Benchmark & Sim-to-Real & Tasks & \shortstack{Avg\\ Subgoals} & \shortstack{Max\\ Subgoals} & \shortstack{Tasks with\\ $\geq 6$ subgoals} & \shortstack{Incremental\\ Subgoal Tasks} \\
\midrule
CALVIN~(\citeauthor{mees2022calvin}) & $\times$ & 34 & 1.0 & 1 & 0 & 0 \\
LIBERO-Long~(\citeauthor{liu2023libero}) & $\times$ & 10 & 1.6 & 2 & 0 & 0 \\
SimplerEnv~(\citeauthor{li2024simplerenv}) & \checkmark & 4 & 1.0 & 1 & 0 & 0 \\
\robotarena{}~(\citeauthor{jangir2025robotarena}) & \checkmark & 74 & 1.0 & 2 & 0 & 0 \\
RoboLab~(\citeauthor{yang2026robolab}) & \checkmark & 120 & 1.9 & 11 & 4 & 56 \\
RoboVoLo~(\citeauthor{chen2026volo}) & \checkmark & 126 & 2.0 & 7 & 2 & 2 \\
\midrule
\textbf{\benchname{}} & \textbf{\checkmark} & \textbf{145} & \textbf{4.7} & \textbf{10} & \textbf{52} & \textbf{138} \\
\bottomrule
\end{tabular}
\end{table}

\FloatBarrier
\section{Long-Horizon Manipulation Benchmark}
\label{sec:benchmarks}

\begin{figure*}[!t]
\centering
\input{data/fig_anatomy.tex}
\definecolor{cSuiteA}{HTML}{0072B2}
\definecolor{cSuiteB}{HTML}{CC79A7}
\definecolor{cSuiteC}{HTML}{009E73}
\definecolor{cGo}{HTML}{D55E00}
\definecolor{cGoB}{HTML}{0072B2}
\definecolor{cNo}{HTML}{C1272D}
{\setlength{\fboxsep}{0.7pt}%
\ifdefined\LaTeXML\includegraphics[width=\linewidth]{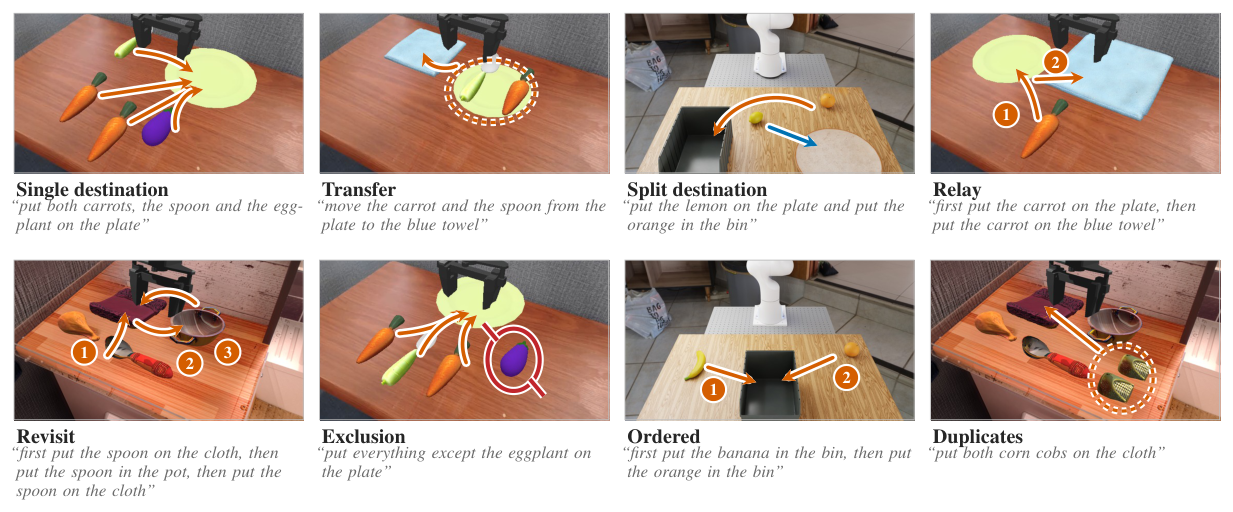}\else\begin{tikzpicture}[x=1cm, y=1cm, font=\small,
  gl/.style  = {text=black!55, font=\fontsize{5.9}{6.4}\selectfont},
  ins/.style = {text=black!58, font=\fontsize{5.6}{6.2}\selectfont\itshape},
  nm/.style  = {font=\fontsize{6.6}{7.0}\selectfont\bfseries},
]
\useasboundingbox (0,0.06) rectangle (13.90,-5.70);

\pgfmathsetmacro{\cw}{3.34}
\pgfmathsetmacro{\ch}{1.843}
\pgfmathsetmacro{\yperx}{\cw/\ch}

\newcommand{\arw}[6]{%
  \draw[white, line width=2.9pt, line cap=round,
        -{Stealth[length=4.3pt, width=4.3pt, line width=2.2pt]}]
    (#3,#4) to[bend left=#1] (#5,#6);
  \draw[#2, line width=1.2pt, line cap=round, -{Stealth[length=4.3pt, width=4.3pt]}]
    (#3,#4) to[bend left=#1] (#5,#6);}

\newcommand{\stp}[3]{\node[circle, fill=cGo, draw=white, line width=0.7pt, text=white,
                           inner sep=0pt, minimum size=2.9mm,
                           font=\fontsize{5.4}{5.4}\selectfont\bfseries] at (#1,#2) {#3};}

\newcommand{\nope}[4]{%
  \pgfmathsetmacro{\nxi}{0.62*#3}\pgfmathsetmacro{\nyi}{0.62*#4}
  \pgfmathsetmacro{\nxo}{1.16*#3}\pgfmathsetmacro{\nyo}{1.16*#4}
  \draw[white, line width=2.7pt] (#1,#2) ellipse (#3 and #4);
  \draw[cNo, line width=1.35pt] (#1,#2) ellipse (#3 and #4);
  \foreach \s in {-1,1}{
    \draw[white, line width=2.7pt] ({#1+\s*\nxi},{#2+\s*\nyi}) -- ({#1+\s*\nxo},{#2+\s*\nyo});
    \draw[cNo, line width=1.35pt] ({#1+\s*\nxi},{#2+\s*\nyi}) -- ({#1+\s*\nxo},{#2+\s*\nyo});}}

\newcommand{\ring}[5]{%
  \draw[white, dash pattern=on 1.7pt off 1.3pt, line width=2.4pt] (#1,#2) ellipse (#3 and #4);
  \draw[#5, dash pattern=on 1.7pt off 1.3pt, line width=1.05pt] (#1,#2) ellipse (#3 and #4);}

\input{data/fig_overlays.tex}

\foreach \i/\key/\name/\example/\suite in \anTypeCards{
  \pgfmathsetmacro{\col}{mod(\i,4)}
  \pgfmathsetmacro{\row}{floor(\i/4)}
  \pgfmathsetmacro{\x}{\col*(\cw+0.1867)}
  \pgfmathsetmacro{\y}{-\row*2.85}
  \begin{scope}[shift={(\x,\y)}]
    \node[anchor=north west, inner sep=0pt] at (0,0)
      {\ifcase\i\relax\includegraphics[width=\cw cm]{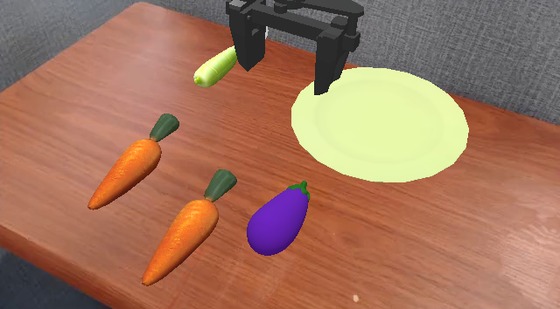}\or\includegraphics[width=\cw cm]{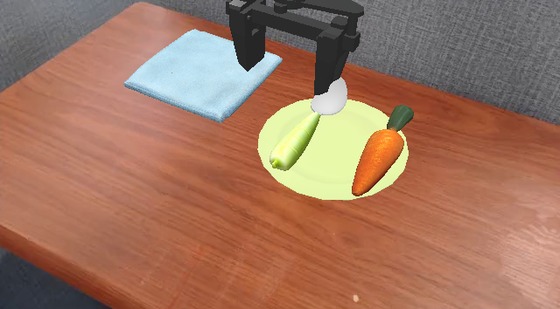}\or\includegraphics[width=\cw cm]{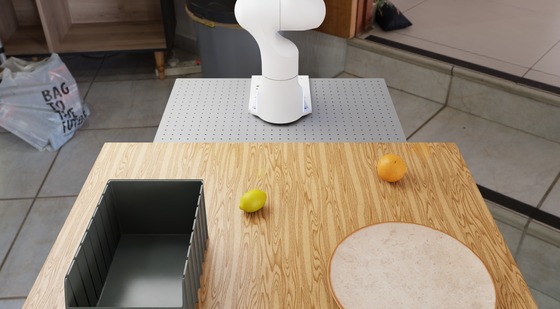}\or\includegraphics[width=\cw cm]{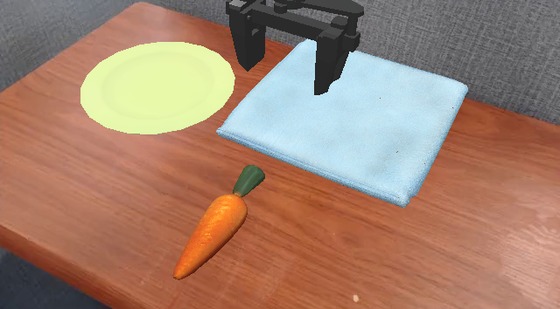}\or\includegraphics[width=\cw cm]{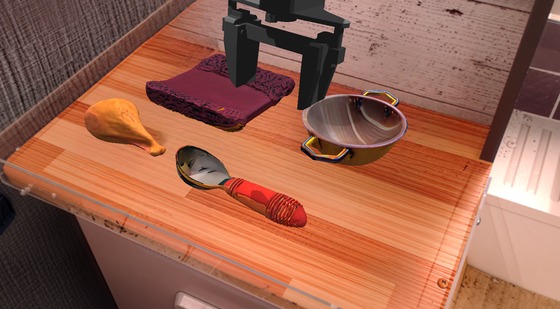}\or\includegraphics[width=\cw cm]{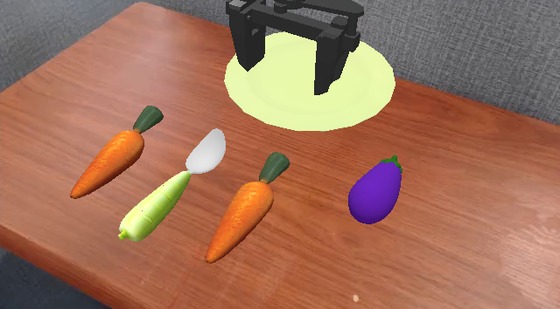}\or\includegraphics[width=\cw cm]{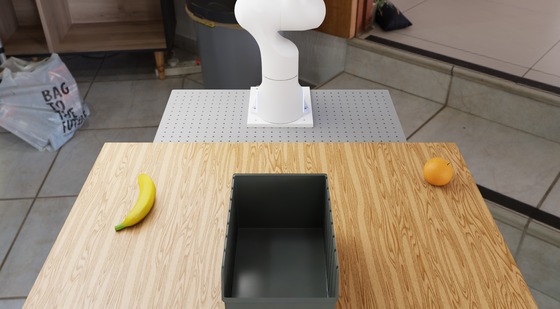}\or\includegraphics[width=\cw cm]{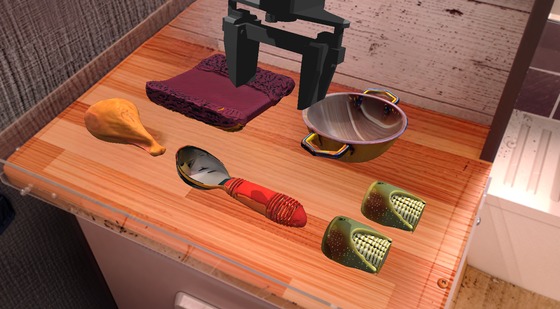}\fi};
    \draw[black!25, line width=0.3pt] (0,0) rectangle (\cw,-\ch);
    \node[anchor=north west, nm, text=black!85,
          inner sep=0pt] at (0.02,-\ch-0.10) {\name};
    \node[anchor=north west, ins, align=left, text width=3.28cm, inner sep=0pt]
      at (0.02,-\ch-0.30) {``\example''};
    \begin{scope}[x=\cw cm, y=-\ch cm]
      \csname anover\key\endcsname
    \end{scope}
  \end{scope}}

\end{tikzpicture}\fi}
\caption{\textbf{Manipulation task categories.} Example scenes and natural language instructions across diverse task structures. Visual overlays indicate target placements and task constraints.}
\label{fig:coverage}
\end{figure*}

We build \benchname{} on three established platforms: SimplerEnv~\citep{li2024simplerenv}, \robotarena{}~\citep{jangir2025robotarena}, and RoboLab~\citep{yang2026robolab}. Each suite---\emph{SimplerEnv-Long}, \emph{\robotarena{}-Long}, and \emph{RoboLab-Long}---keeps the original assets while adding four key features (Table~\ref{tab:benchmarks}): 
(1)~\textbf{controlled horizon scaling}, from 1 to 10 subgoals with \tasksIncremental{} tasks that extend a shorter task of the same scene or are extended by a longer one, and \shareSixPlus{}\% of tasks comprising 6 or more subgoals; 
(2)~\textbf{diverse task categories}, covering eight spatial and logical task structures (Figure~\ref{fig:coverage}); 
(3)~\textbf{explicit per-subgoal predicates}, checking every subgoal programmatically; and (4)~\textbf{multiple embodiments}, evaluating policies across multiple robot platforms.

\phantomsection\label{sec:benchmarks:setup}
\noindent\textbf{Problem setting.} We define a \emph{long-horizon task} as one that requires completing several subgoals in sequence from a single full instruction. Each \emph{subgoal} is a single pick-and-place action (for example, putting a carrot on a plate), the target location is its \emph{destination}, and the number of subgoals is the task's \emph{horizon}. An \emph{episode} is one execution rollout starting from a randomized scene layout.

The systems we evaluate control the robot with a low-level policy, a task-and-motion planner, or motion primitives. A low-level policy maps visual observations and a language instruction to robot control actions. Generalist low-level policies are trained on demonstration datasets (e.g., BridgeData and DROID) of short-horizon skills, like placing one object in a container (e.g., a plate, towel, bowl, or bin). We evaluate them without fine-tuning on data of the target tasks and environments.

\phantomsection\label{sec:benchmarks:axes}
\noindent\textbf{Task horizon.}
Task horizon is the number of subgoals an instruction requires the robot to complete. 
To isolate the effect of horizon, tasks form ladders: within one scene and task category, a longer task keeps all subgoals of a shorter one and adds more (for example, \emph{``put the carrot on the plate''} and \emph{``put the carrot and the spoon on the plate''}). Every task on such a ladder is \emph{incremental}. In total, \tasksIncremental{} of our \tasksTotalAll{} tasks are incremental (Table~\ref{tab:benchmarks}). Except in exclusion tasks, instructions explicitly name each target object rather than using vague collective phrases (like \emph{``put everything in the bin''}), ensuring that we test instruction following rather than random searching. Tasks scale from 1 up to 10 subgoals, which reaches the maximum object density that fits in the workspace without objects blocking basic grasps.\\
\textbf{Task categories.} 
Task categories define the spatial and logical structure of each instruction (Figure~\ref{fig:coverage}). Across the three suites, we evaluate eight task categories (every task: Appendix~\ref{sec:app:catalog}):
\textbf{single destination}, moving multiple objects to one container (e.g., \emph{``put the spoon and the carrot on the plate''});
\textbf{split destinations}, sorting objects into different containers (e.g., \emph{``put the lemon on the plate and the orange in the bin''});
\textbf{transfers}, moving objects out of one container into another (e.g., \emph{``move the carrot and the spoon from the plate to the blue towel''});
\textbf{relay routing}, moving an object through an intermediate stop (e.g., \emph{``first put the carrot on the plate, then on the towel''});
\textbf{revisitations}, returning to a container used earlier (e.g., \emph{``put the spoon on the cloth, then in the pot, then back on the cloth''});
\textbf{exclusions}, moving specific objects while leaving distractors alone (e.g., \emph{``put everything except the eggplant on the plate''});
\textbf{ordered sequences}, following a strict order (e.g., \emph{``first put the banana in the bin, then the orange''}); and
\textbf{duplicate instances}, picking specific items among identical visual copies (e.g., \emph{``put both corn cobs on the cloth''}).
Evaluating all eight categories allows us to measure how different geometric and logical constraints affect policy success (\S\ref{sec:exp:tasktype}).

\phantomsection\label{sec:benchmarks:construction}
\noindent\textbf{Scene construction.} Each scene is physically stable and within the robot's reachable workspace. A longer task adds more objects and a larger container for them; object sizes stay the same, so grasping difficulty does not grow with the horizon. Every layout is validated by scripted policies that complete all subgoals, so failures are caused by sequential execution rather than reachability issues.

\phantomsection\label{sec:benchmarks:metrics}
\noindent\textbf{Evaluation metrics.} We evaluate every subgoal with programmatic state checks. We report two main metrics: \textbf{Task Success Rate (\%)}, the percentage of episodes where all subgoals and constraints are satisfied at the end of the episode, and \textbf{Subgoal Progress (\%)}, the percentage of required subgoals completed, measuring partial progress on unfinished tasks. For every subgoal that is not completed, we also report a \textbf{Failure Cause}, such as a subgoal that was completed and later undone (\S\ref{sec:exp:failures}). The step budget grows with the number of subgoals: \budgetStepsPerGoal{} steps (\budgetSecondsPerGoal\,s at 5\,Hz) per subgoal on the WidowX suites, and a per-horizon budget of \budgetRobolabMin{} to \budgetRobolabMax{} steps (\budgetRobolabMinS\,s to \budgetRobolabMaxS\,s at 15\,Hz) on RoboLab-Long. The budget counts simulated time only: the simulation pauses while a system computes.

\section{Characterizing orchestrators}
\label{sec:orchestrator}

To analyze how different orchestration methods support long-horizon manipulation, we characterize their operation around four design decisions (Figure~\ref{fig:orchestrator}, Table~\ref{tab:decisions}): (1)~subgoal planning, (2)~trigger mechanisms, (3)~visual verification, and (4)~failure recovery. This characterization informs our experimental comparisons and analysis, helping us identify how decisions in each of these aspects affect task execution and performance. We divide the evaluated systems into two groups. \emph{Learned-policy} systems send text subgoals to a low-level policy. \emph{Tool selection} systems do not use a low-level policy. Instead, they call motion primitives or a task-and-motion planner. In addition to existing systems, we introduce \harness{} (Plan-See-Pick Agent), a tool selection system in which a VLM plans text subgoals and executes each subgoal by calling manipulation primitives that verify their own outcome (Appendix~\ref{app:harness}).
\newcommand{\lab}{\color{black!55}\footnotesize}
\begin{figure*}[!ht]
\centering
\definecolor{advcol}{HTML}{009988}
\definecolor{witcol}{HTML}{CC3311}
\ifdefined\LaTeXML\includegraphics[width=0.77\linewidth]{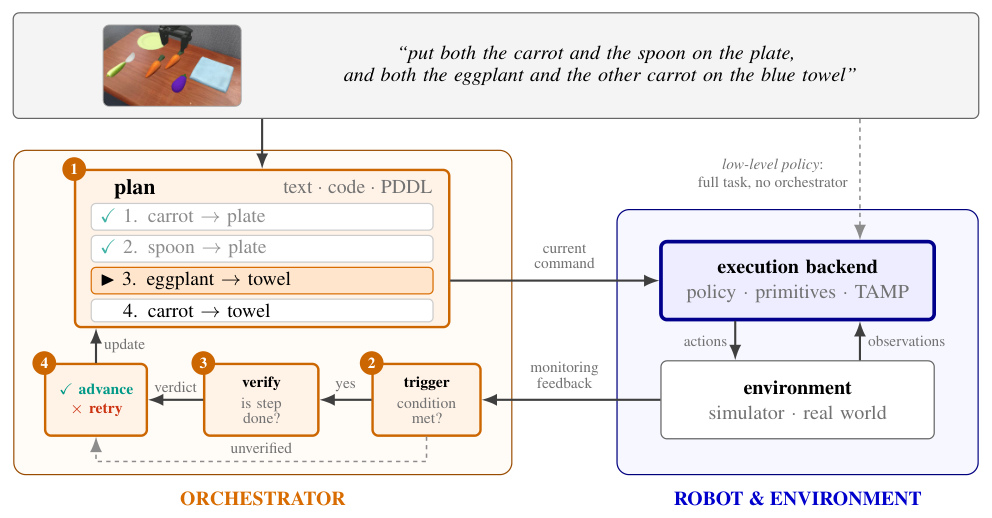}\else\begin{tikzpicture}[
  scale=0.72, transform shape,
  x=1cm, y=1cm, font=\small,
  >={Latex[length=1.8mm,width=1.4mm]},
  box/.style   = {draw=black!60, rounded corners=2pt, align=center,
                  inner xsep=4pt, inner ysep=4pt, fill=white},
  pol/.style   = {box, draw=blue!55!black, fill=blue!7, very thick},
  dec/.style   = {box, draw=orange!80!black, fill=orange!9, thick},
  lab/.style   = {text=black!55, font=\footnotesize},
  alab/.style  = {text=black!55, font=\scriptsize, align=center},
  chip/.style  = {rounded corners=1.5pt, inner xsep=4pt, inner ysep=2.0pt,
                  align=left, text width=5.2cm, fill=white, draw=black!25},
  badge/.style = {circle, fill=orange!80!black, text=white, minimum size=3.8mm, inner sep=0pt,
                  font=\fontsize{7}{7}\selectfont\bfseries},
  flow/.style  = {->, draw=black!75, line width=0.7pt},
  soft/.style  = {->, draw=black!45, line width=0.5pt, dashed, dash pattern=on 1.4pt off 1.4pt},
]

\node[dec, anchor=north west, minimum width=6.0cm, minimum height=2.52cm] (plan) at (0.9,-1.15) {};
\node[badge] at (plan.north west) {1};

\node[anchor=north west, font=\bfseries, inner sep=0pt] (plantitle) at ([xshift=18pt, yshift=-4.5pt]plan.north west) {plan};
\node[anchor=north east, lab, inner sep=0pt] at ([xshift=-8pt, yshift=-4.5pt]plan.north east) {text $\cdot$ code $\cdot$ PDDL};

\node[chip, text width=5.2cm, fill=white, draw=black!20, text=black!45, anchor=north] (c1) at ([yshift=-15.5pt]plan.north)
  {\textcolor{advcol!75}{$\checkmark$}~1. carrot $\rightarrow$ plate};
\node[chip, text width=5.2cm, fill=white, draw=black!20, text=black!45, anchor=north] (c2) at ([yshift=-2.0pt]c1.south)
  {\textcolor{advcol!75}{$\checkmark$}~2. spoon $\rightarrow$ plate};
\node[chip, text width=5.2cm, fill=orange!20, draw=orange!85!black, anchor=north] (c3) at ([yshift=-2.0pt]c2.south)
  {$\blacktriangleright$~3. eggplant $\rightarrow$ towel};
\node[chip, text width=5.2cm, fill=white, draw=black!20, anchor=north] (c4) at ([yshift=-2.0pt]c3.south)
  {\phantom{$\checkmark$}~4. carrot $\rightarrow$ towel};

\node[dec, anchor=center, text width=1.35cm, minimum height=1.15cm, font=\scriptsize] (rout) at (1.25,-4.85) {
  \textcolor{advcol}{$\checkmark$ \textbf{advance}}\\[1pt]
  \textcolor{witcol}{$\times$ \textbf{retry}}
};
\node[badge] at (rout.north west) {4};

\node[dec, anchor=center, text width=1.55cm, minimum height=1.15cm, font=\scriptsize] (ver) at (3.90,-4.85) {
  \textbf{verify}\\[1.5pt] {\color{black!55}\scriptsize is step\\[-1pt]done?}
};
\node[badge] at (ver.north west) {3};

\node[dec, anchor=center, text width=1.45cm, minimum height=1.15cm, font=\scriptsize] (trig) at (6.55,-4.85) {
  \textbf{trigger}\\[1.5pt] {\color{black!55}\scriptsize condition\\[-1pt]met?}
};
\node[badge] at (trig.north west) {2};

\draw[flow] (trig.west) -- node[alab, above, pos=0.50] {yes} (ver.east);
\draw[flow] (ver.west) -- node[alab, above, pos=0.50] {verdict} (rout.east);
\draw[flow] (rout.north) -- node[alab, right] {update} (plan.south -| rout.north);
\draw[soft] (trig.south) -- ++(0,-0.40) coordinate (unvermid)
            -- node[alab, above] {unverified} (rout.south |- unvermid)
            -- (rout.south);

\coordinate (c3level) at (0,0 |- c3.center);
\node[pol, anchor=center, text width=4.1cm, minimum height=1.25cm] (pi) at ([xshift=12.5cm]c3level) {
  \textbf{execution backend}\\[1.5pt]
  {\lab policy $\cdot$ primitives $\cdot$ TAMP}
};

\node[box, anchor=center, text width=4.1cm, minimum height=1.25cm] (env) at (12.5,-4.85) {
  \textbf{environment}\\[1.5pt]
  {\lab simulator $\cdot$ real world}
};

\draw[flow] ([xshift=-10mm]pi.south) -- node[alab, left] {actions} ([xshift=-10mm]env.north);
\draw[flow] ([xshift=10mm]env.north) -- node[alab, right] {observations} ([xshift=10mm]pi.south);

\begin{scope}[on background layer]
  \coordinate (orchtop) at (3.9,-0.85);
  \coordinate (orchbot) at (3.9,-6.05);
  \coordinate (exectop) at (12.5,-1.80);
  \coordinate (execbot) at (12.5,-6.05);

  \node[draw=orange!60!black, rounded corners=4pt, fill=orange!3,
        fit=(plan)(rout)(ver)(trig)(orchtop)(orchbot), inner xsep=10pt, inner ysep=0pt] (orchbox) {};
  \node[font=\bfseries\footnotesize, text=orange!85!black, anchor=north] at ([yshift=-0.14cm]orchbox.south)
    {ORCHESTRATOR};

  \node[draw=blue!50!black, rounded corners=4pt, fill=blue!3,
        fit=(pi)(env)(exectop)(execbot), inner xsep=14pt, inner ysep=0pt] (execbox) {};
  \node[font=\bfseries\footnotesize, text=blue!80!black, anchor=north] at ([yshift=-0.14cm]execbox.south)
    {ROBOT \& ENVIRONMENT};
\end{scope}

\coordinate (groupleft) at ($(orchbox.west)!0.5!(execbox.east) + (-6.3,0)$);
\node[inner sep=0pt, anchor=north west] (img) at (groupleft |- 0,1.16)
  {\phantom{\includegraphics[trim=0 50 0 55, clip, width=2.2cm]{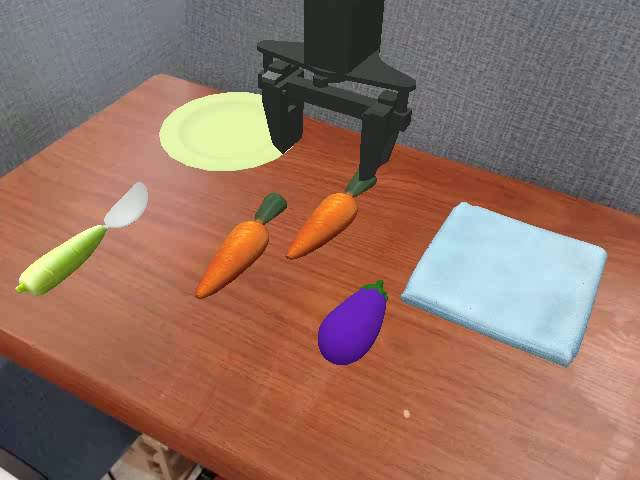}}};
\begin{scope}[on background layer]
  \node[box, fill=black!4, inner xsep=0pt, inner ysep=4pt,
        fit=(img)(orchbox.west |- img.center)(execbox.east |- img.center)] (banner) {};
\end{scope}
\begin{scope}
  \clip[rounded corners=2pt] (img.south west) rectangle (img.north east);
  \node[inner sep=0pt] at (img.center) {\includegraphics[trim=0 50 0 55, clip, width=2.2cm]{figures/frames/rollout_1.png}};
\end{scope}
\draw[black!35, line width=0.3pt, rounded corners=2pt] (img.south west) rectangle (img.north east);
\node[anchor=west, text width=9.6cm, align=center, font=\small\itshape] at ([xshift=0.8cm]img.east)
  {``put both the carrot and the spoon on the plate,\\
  and both the eggplant and the other carrot on the blue towel''};

\coordinate (channellab) at ($(orchbox.east)!0.5!(execbox.west)$);

\draw[flow] (banner.south -| plan.north) -- (plan.north);

\draw[flow] (plan.east |- pi.west) -- (pi.west);
\node[alab, above=2pt] at (channellab |- pi.west) {current\\[-1pt]command};

\draw[flow] (env.west) -- (trig.east);
\node[alab, above=2pt] at (channellab |- env.west) {monitoring\\[-1pt]feedback};

\coordinate (direntry) at ([xshift=1.0cm]pi.north);
\draw[soft] (banner.south -| direntry) -- node[alab, left=2pt, pos=0.45] {\emph{low-level policy}:\\full task, no orchestrator} (direntry);

\end{tikzpicture}\fi
\caption{\textbf{Overview of the orchestrator architecture.} The top box shows the task input (initial camera view and full instruction). Orange numbered boxes mark the four orchestrator decisions evaluated in our study (Table~\ref{tab:decisions}). The blue box denotes the execution backend. Dashed lines show baselines without orchestration (\emph{low-level policy}) or without verification (\emph{unverified}).}
\label{fig:orchestrator}
\end{figure*}

\noindent\textbf{1. Subgoal planning.} A plan decomposes a full instruction into sequential single-subgoal instructions (for example, decomposing \emph{``put the carrot and the spoon on the plate''} into subgoal 1: \emph{``put the carrot on the plate''} and subgoal 2: \emph{``put the spoon on the plate''}). In our experiments, the planner is always a highly capable frontier Vision-Language Model (VLM). The only exception is the privileged verifier, which takes the ground-truth subgoals from the simulator. Our learned-policy variants and \harness{} generate the full list of text subgoals once, at the start of the episode (\emph{Text subgoals}). VoLo~\citep{chen2026volo}, which we run with the low-level policy as its only execution tool, also generates a list of text subgoals from the first frame and replans the remaining ones during the episode. The other external systems use different plan representations: \capx{}~\citep{fu2026capx} generates a Python program, \tiptop{}~\citep{shen2026tiptop} generates symbolic goal predicates for its planner, and \gptpolicy{}~\citep{cheng2026incontextrobotlearningvlm} selects only the next tool call. \emph{Low-level policy} is the no-orchestrator baseline and passes the full instruction to the low-level policy without generating a subgoal plan.\\
\noindent\textbf{2. Trigger mechanisms.} Because the low-level policy runs continuously without knowing when to stop, the orchestrator needs a trigger condition to decide when to switch or advance to the next subgoal. We evaluate three subgoal-switching strategies:
(1)~\emph{Periodic trigger}: advances after a fixed number of steps; (2)~\emph{Proprioceptive trigger}: advances to the next subgoal when internal robot state indicates an attempt has finished, specifically when the gripper opens after a sustained close (Appendix~\ref{app:orch}); (3)~\emph{Privileged trigger}: a simulation-only diagnostic that checks ground-truth contacts at every step, advancing to the next subgoal the instant the current goal is reached; it is used by the \emph{Privileged verifier}. Tool selection systems execute blocking commands and make the next decision when a command completes, either after each primitive call (\emph{Primitive completion}) or after the full program (\emph{Program completion}). \tiptop{} executes its plan open-loop, without a trigger.\\
\noindent\textbf{3. Visual verification.} When a trigger fires, a verifier determines whether the subgoal succeeded. Across our controlled variants and the privileged verifier, we use three verification settings: no verifier (\emph{None}), which advances unconditionally; visual verification (\emph{Visual (VLM)}), which queries a vision-language model using the latest camera observation and subgoal description; and simulator-state verification (\emph{Simulator state}), used by the privileged verifier to check the task predicate directly. \harness{} verifies each primitive geometrically from the depth image (\emph{Geometric state}), for example by checking that a grasped object has left its initial position. \capx{} and \gptpolicy{} query the VLM with the new camera image.\\
\noindent\textbf{4. Failure recovery.} After verification, the recovery mechanism determines whether to advance, retry, or replan. In our retry-based variants, a completed subgoal is removed from the queue and execution advances; an incomplete subgoal is retried in place. When a plan contains duplicate subgoals for interchangeable objects, an incomplete subgoal moves to the end of the queue so one difficult object does not consume the episode budget. The privileged verifier retries immediately. Without a verifier, the orchestrator always advances to the next subgoal (\emph{Advance}). To replan (\emph{Replan}), the VLM generates a new plan from the current camera image: VoLo replans the remaining subgoals, \capx{} generates a new program, and \gptpolicy{} selects a new tool call. \harness{} combines both mechanisms: it retries failed subgoals and replans after all subgoals have been attempted (Appendix~\ref{app:harness}).

\begin{table*}[!ht]
\centering
\caption{\textbf{Overview of the evaluated systems according to the proposed characterization.} Each configuration maps to the orchestrator decisions in Figure~\ref{fig:orchestrator}. Top: systems using a low-level policy backend. Bottom: systems using alternative execution backends.}
\label{tab:decisions}
\scriptsize
\setlength{\tabcolsep}{2pt}
\raggedright
\begin{tabular*}{\linewidth}{@{\extracolsep{\fill}} llllll @{}}
\toprule
System & Backend & Plan & Trigger & Verification & Recovery \\
\midrule
Low-level policy                & Learned policy    & None                  & None                 & None            & None \\
Periodic trigger                & Learned policy    & Text subgoals         & Periodic             & None            & Advance \\
Proprioceptive trigger          & Learned policy    & Text subgoals         & Proprioceptive       & None            & Advance \\
Proprioceptive verifier         & Learned policy    & Text subgoals         & Proprioceptive       & Visual (VLM)    & Retry \\
VoLo~\citep{chen2026volo}       & Learned policy    & Text subgoals         & Periodic             & Visual (VLM)    & Replan \\
Privileged verifier             & Learned policy    & Ground-truth subgoals & Privileged           & Simulator state & Retry \\
\midrule
\capx{}~\citep{fu2026capx}      & Motion primitives & Python program        & Program completion   & Visual (VLM)    & Replan \\
\tiptop{}~\citep{shen2026tiptop}& Symbolic TAMP     & Goal predicates       & None                 & None            & None \\
\harness{}                      & Motion primitives & Text subgoals         & Primitive completion & Geometric state & Retry, replan \\
\gptpolicy{}~\citep{cheng2026incontextrobotlearningvlm} & Motion primitives & Next tool call & Primitive completion & Visual (VLM) & Replan \\
\bottomrule
\end{tabular*}
\end{table*}

\section{Experiments}
\label{sec:experiments}

We first study how task success scales with task horizon (\S\ref{sec:exp:horizon}). Then we investigate the dominant failures in different systems (\S\ref{sec:exp:failures}) and whether simulation predicts real-world performance (\S\ref{sec:exp:realrobot}).

\noindent\textbf{Experimental setup and low-level policies.}
On \emph{SimplerEnv-Long} and \emph{\robotarena{}-Long} we evaluate a $\pi_0$-like~\citep{black2024pi0} low-level policy that uses Qwen3-VL~\citep{qwen3vl} as backbone and is trained on BridgeData~v2~\citep{walke2023bridgedata}.
On \emph{RoboLab-Long} we use the public \pifive{}~\citep{intelligence2025pi05} fine-tuned on DROID~\citep{khazatsky2024droid} (Appendix~\ref{app:policy_training}).
For periodic-trigger variants, we use \periodicStepsBridge{} steps (5.4\,s at 5\,Hz) on the WidowX suites and \periodicStepsRobolab{} steps (16.7\,s at 15\,Hz) on RoboLab-Long. In both cases, the interval is set to the median time the privileged verifier takes to complete a subgoal.
Unless noted otherwise, visual verifiers use Gemini 3.7 Flash. \gptpolicy{} uses GPT-6 Astra. In simulation, the tool selection systems use the ground-truth depth image and camera calibration from the simulator. In real-world experiments, depth is estimated with MoGe-2~\citep{wang2025moge2} from calibrated cameras.
For the simulation results we report 95\% Wilson confidence intervals over episodes, and test differences between systems paired over tasks (Appendix~\ref{app:per-suite}).

\subsection{Task Horizon Scaling}
\label{sec:exp:horizon}
\begin{figure*}[!ht]
\centering
\includegraphics[width=0.66\linewidth]{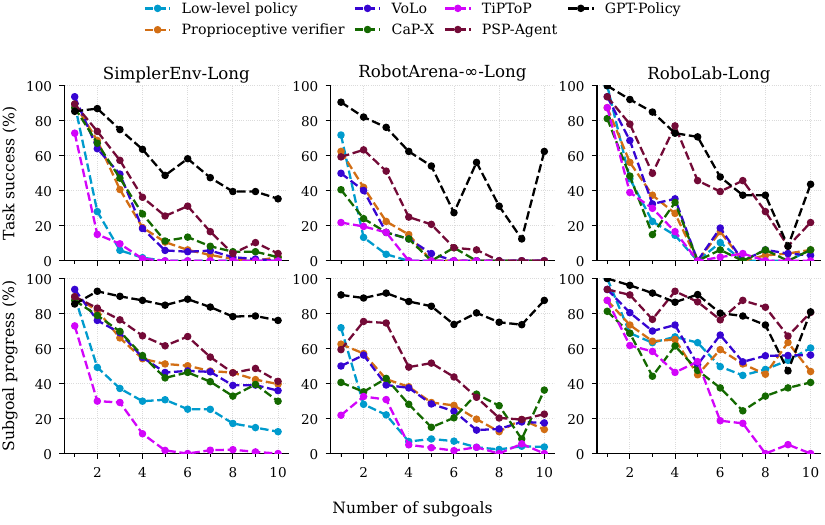}
\caption{\textbf{Performance scaling across task horizon.} Task success rate (top) and subgoal progress (bottom) as the horizon increases from 1 to 10 subgoals across all three benchmark suites (all ten systems: Tables~\ref{tab:suite-horizon-success} and~\ref{tab:suite-horizon-progress}).}
\label{fig:ladder}
\end{figure*}

On single-subgoal tasks, the low-level policy reaches a similar success rate as the \emph{Proprioceptive verifier} (\introFlatOne\% vs.\ \introVerOne\%). As task horizon grows, the low-level policy rapidly collapses to \introFlatTwo\% at 2 subgoals and \introFlatFive\% at 5 subgoals.
In contrast, orchestrator-based methods maintain higher success rates with increasing task horizon with \emph{Proprioceptive verifier} achieving \introVerTwo\% at 2 subgoals, \introVerFive\% at 5 subgoals and retaining partial progress up to 10 subgoals (Figure~\ref{fig:ladder}, Table~\ref{tab:comp-overall}).
The Privileged verifier serves as a privileged reference for the specific low-level policy being evaluated. The best learned-policy orchestration methods, VoLo and the \emph{Proprioceptive verifier}, decrease the gap of the low-level policy to this reference from 18.6 to 4.2--4.5 percentage points.
Tool selection systems, which use ground-truth depth in simulation, are not limited by the low-level policy. \gptpolicy{} achieves the highest success rate on every suite (\gridGptAll\% overall), and \harness{} is second (\gridHarnessAll\%) while using the same VLM as the learned-policy orchestrators. Both exceed the privileged reference of the learned-policy systems, and on 5-subgoal tasks \harness{} still succeeds in \introHarnessFive\% of episodes, against \introOracleFive\% for the privileged verifier. \capx{} performs on par with the best learned-policy methods (\gridCapxAll\%), while \tiptop{} (\gridTiptopAll\%) does not improve over the low-level policy.

\begin{table}[!ht]
\centering
\setlength{\tabcolsep}{3pt}
\caption{
    \textbf{Task success rate (\%) of every system.}
    Top: per suite, over all tasks (1 to 10 subgoals).
    Middle: per task category.
    Bottom: all tasks.
    Middle and bottom combine the episodes of the three suites.
    $\pm$ covers the 95\% Wilson interval.
    \textcolor{bestlearned}{Green} marks the best learned-policy method, \textcolor{bestscripted}{blue} the best tool selection method (\capx{}, \tiptop{}, \harness{} or \gptpolicy{}).
    Privileged verifier is a privileged reference for learning methods using this specific low-level policy.
}
\label{tab:comp-overall}
\small
\resizebox{\linewidth}{!}{%
\begin{tabular}{lc|cccc|c|cccc}
\toprule
Tasks & \shortstack{Low-level\\policy} & \shortstack{Periodic\\trigger} & \shortstack{Proprioceptive\\trigger} & \shortstack{Proprioceptive\\verifier} & VoLo & \shortstack{Privileged\\verifier} & \capx{} & \tiptop{} & \harness{} & \gptpolicy{} \\
\midrule
SimplerEnv-Long & 8.7\,\smallpm{1.7} & 17.4\,\smallpm{2.2} & 21.5\,\smallpm{2.3} & 25.2\,\smallpm{2.4} & \textcolor{bestlearned}{25.5\,\smallpm{2.4}} & 29.5\,\smallpm{2.5} & 28.7\,\smallpm{2.5} & 6.7\,\smallpm{1.5} & 37.1\,\smallpm{2.7} & \textcolor{bestscripted}{\textbf{60.3\,\smallpm{2.7}}} \\
\robotarena{}-Long & 10.9\,\smallpm{3.6} & 15.2\,\smallpm{4.0} & 16.2\,\smallpm{4.1} & \textcolor{bestlearned}{24.2\,\smallpm{4.6}} & 21.3\,\smallpm{4.4} & 25.5\,\smallpm{4.6} & 16.2\,\smallpm{4.1} & 11.2\,\smallpm{3.6} & 39.9\,\smallpm{5.0} & \textcolor{bestscripted}{\textbf{66.8\,\smallpm{4.9}}} \\
RoboLab-Long & 19.6\,\smallpm{4.5} & 25.6\,\smallpm{4.8} & 28.1\,\smallpm{4.9} & 25.6\,\smallpm{4.8} & \textcolor{bestlearned}{29.0\,\smallpm{4.9}} & 34.4\,\smallpm{5.1} & 20.7\,\smallpm{4.5} & 17.3\,\smallpm{4.3} & 51.4\,\smallpm{5.2} & \textcolor{bestscripted}{\textbf{62.8\,\smallpm{5.2}}} \\
\midrule
Single destination & 14.9\,\smallpm{2.9} & 23.5\,\smallpm{3.4} & 25.0\,\smallpm{3.4} & \textcolor{bestlearned}{31.0\,\smallpm{3.6}} & \textcolor{bestlearned}{31.0\,\smallpm{3.6}} & 33.8\,\smallpm{3.7} & 23.5\,\smallpm{3.4} & 12.9\,\smallpm{2.8} & 44.0\,\smallpm{3.8} & \textcolor{bestscripted}{\textbf{61.8\,\smallpm{3.7}}} \\
Transfer & 16.7\,\smallpm{6.9} & 14.6\,\smallpm{6.7} & 14.6\,\smallpm{6.7} & 15.3\,\smallpm{6.8} & \textcolor{bestlearned}{27.1\,\smallpm{7.8}} & 26.4\,\smallpm{7.7} & 43.8\,\smallpm{8.2} & 10.4\,\smallpm{6.1} & 36.1\,\smallpm{8.1} & \textcolor{bestscripted}{\textbf{68.8\,\smallpm{8.0}}} \\
Split destination & 7.1\,\smallpm{3.5} & 11.5\,\smallpm{4.1} & 17.6\,\smallpm{4.7} & 17.2\,\smallpm{4.7} & \textcolor{bestlearned}{20.6\,\smallpm{5.0}} & 24.3\,\smallpm{5.2} & 16.2\,\smallpm{4.6} & 12.2\,\smallpm{4.2} & 50.7\,\smallpm{5.7} & \textcolor{bestscripted}{\textbf{60.5\,\smallpm{5.7}}} \\
Relay & 2.3\,\smallpm{2.4} & 10.9\,\smallpm{4.0} & 15.5\,\smallpm{4.5} & \textcolor{bestlearned}{21.4\,\smallpm{5.0}} & 16.1\,\smallpm{4.6} & 27.3\,\smallpm{5.3} & 32.2\,\smallpm{5.4} & 0.3\,\smallpm{1.5} & 24.3\,\smallpm{5.1} & \textcolor{bestscripted}{\textbf{71.7\,\smallpm{5.3}}} \\
Revisit & 0.0\,\smallpm{2.2} & 1.2\,\smallpm{3.0} & 1.2\,\smallpm{3.0} & 3.0\,\smallpm{3.8} & \textcolor{bestlearned}{3.6\,\smallpm{4.0}} & 4.8\,\smallpm{4.4} & 8.9\,\smallpm{5.3} & 0.0\,\smallpm{2.2} & 6.0\,\smallpm{4.7} & \textcolor{bestscripted}{\textbf{43.5\,\smallpm{7.6}}} \\
Exclusion & 0.6\,\smallpm{2.7} & 31.0\,\smallpm{7.4} & \textcolor{bestlearned}{39.9\,\smallpm{7.6}} & 39.3\,\smallpm{7.5} & 32.7\,\smallpm{7.4} & 42.3\,\smallpm{7.6} & 31.0\,\smallpm{7.4} & 10.7\,\smallpm{5.6} & 48.8\,\smallpm{7.5} & \textcolor{bestscripted}{\textbf{60.1\,\smallpm{7.6}}} \\
Ordered & 22.2\,\smallpm{7.5} & 28.5\,\smallpm{7.9} & 29.2\,\smallpm{7.9} & 35.4\,\smallpm{8.1} & \textcolor{bestlearned}{38.2\,\smallpm{8.1}} & 39.6\,\smallpm{8.2} & 27.1\,\smallpm{7.8} & 12.5\,\smallpm{6.4} & 56.2\,\smallpm{8.2} & \textcolor{bestscripted}{\textbf{65.3\,\smallpm{8.1}}} \\
Duplicates & 29.7\,\smallpm{8.4} & 24.2\,\smallpm{8.1} & \textcolor{bestlearned}{31.2\,\smallpm{8.5}} & 30.5\,\smallpm{8.4} & \textcolor{bestlearned}{31.2\,\smallpm{8.5}} & 33.6\,\smallpm{8.6} & 25.8\,\smallpm{8.2} & 11.7\,\smallpm{6.7} & 52.3\,\smallpm{8.6} & \textcolor{bestscripted}{\textbf{58.6\,\smallpm{8.7}}} \\
\midrule
Overall & 11.0\,\smallpm{1.4} & 18.4\,\smallpm{1.7} & 21.7\,\smallpm{1.8} & 25.0\,\smallpm{1.9} & \textcolor{bestlearned}{25.3\,\smallpm{1.9}} & 29.6\,\smallpm{2.0} & 25.0\,\smallpm{1.9} & 9.4\,\smallpm{1.3} & 40.1\,\smallpm{2.2} & \textcolor{bestscripted}{\textbf{62.0\,\smallpm{2.1}}} \\
\bottomrule
\end{tabular}
}
\end{table}

\phantomsection\label{sec:exp:tasktype}
Trends also differ across task categories.
On \emph{Exclusion} and \emph{Relay} tasks, the low-level policy nearly always fails while the Proprioceptive verifier approaches the privileged verifier.
On \emph{Single destination},  \emph{Split destination}, \emph{Transfer} and \emph{Ordered} the gain of VoLo is also steady, but smaller.
On \emph{Duplicates} the low-level policy already approaches the privileged verifier and orchestration provides only tiny improvement.
\emph{Revisit}, whose tasks mostly have 6 or more subgoals, resists both: it needs intermediate transfers where execution errors compound, and reaches only \catRevisitAll\% on average over the learned-policy systems (Table~\ref{tab:comp-overall}).

\begin{figure}[!ht]
\centering
\includegraphics[width=0.78\linewidth]{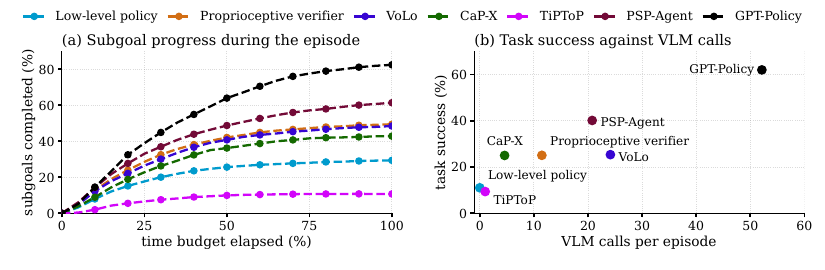}
\caption{\textbf{Robot time and compute.} (a) Share of all subgoals completed as the simulated time budget elapses. (b) Task success against VLM calls per episode. Both combine all suites (per suite: Table~\ref{tab:efficiency}; tokens and API cost: Appendix~\ref{app:efficiency}).}
\label{fig:efficiency}
\end{figure}

To assess the efficiency we analyze the \emph{elapsed time} and the \emph{compute} used by each method.
We calculate \emph{elapsed time} as the percentage of the simulated time budget needed to complete a task and \emph{compute} as the number of VLM calls per episode (Figure~\ref{fig:efficiency}). We do not compare wall-clock time, because it depends on the hardware of each run and on queueing at shared VLM endpoints (Appendix~\ref{app:efficiency}).
Most progress happens early: the \emph{Proprioceptive verifier} completes \effHalfVerifier\% of all subgoals in the first half of the budget and \effEndVerifier\% by the end, indicating that a longer budget would help little. These results suggest that the end-to-end low-level policy limits the returns to test-time scaling: VoLo makes more VLM calls than the \emph{Proprioceptive verifier} but achieves similar success, while \harness{} and \gptpolicy{} pair higher call counts with higher success using motion primitives.

\subsection{Failure Analysis}
\label{sec:exp:failures}

\begin{figure}[!ht]
\centering
\includegraphics[width=0.76\textwidth]{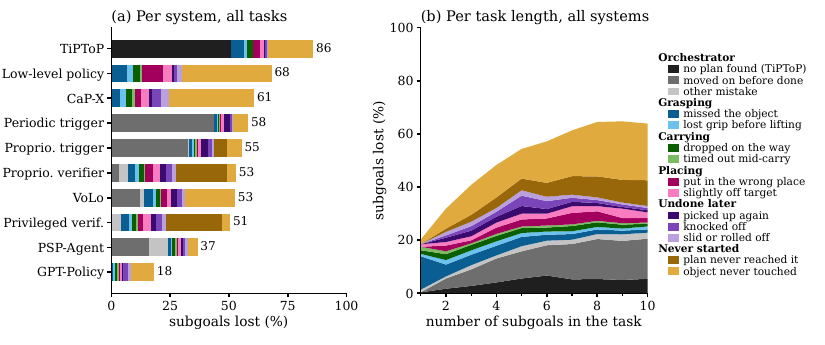}
\caption{\textbf{Why subgoals fail.} Share of subgoals not completed at the end of the episode, by cause (Appendix~\ref{sec:app:failures}). A loss that an orchestrator decision made final is drawn as an orchestrator cause (grey; \capx{} and \gptpolicy{} log no decisions), every other loss by what happened to the object. (a) One bar per system, over all its episodes. (b) One point per task length, over all ten systems. Every value is computed within each suite and then averaged over the three suites (per suite: Figure~\ref{fig:failure-suites}).}
\label{fig:failure-horizon}
\end{figure}

Next, we investigate the dominant failure modes for each method and try to understand whether learned-policy methods are bottlenecked by the orchestrator or the low-level policy.

\phantomsection\label{sec:exp:vlm}
To evaluate orchestrator reasoning independently from execution noise, we benchmark vision-language models offline on \tasksTotalAll{} scenes and \decompBenchStateN{} annotated frames across four capabilities (Table~\ref{tab:vlmbench}): plan decomposition, subgoal verification, scene state tracking, and subgoal switching. Gemini 3.7 Flash and GPT-5.6 Sol lead plan decomposition at \decompBenchExactFlash\%.
Gemini 3.7 Flash also leads state tracking (\decompBenchTrackFlash\%), the hardest capability for most models, and serves as our default orchestrator VLM with the highest mean accuracy over the four capabilities (\decompBenchMeanFlash\%).

\begin{table}[!ht]
\centering
\renewcommand{\arraystretch}{0.90}
\setlength{\tabcolsep}{2.5pt}
\caption{\textbf{Offline evaluation of Vision-Language Models across orchestrator capabilities.} Scores in \%; $\pm$ covers the 95\% Wilson interval, widened for the correlation of frames from the same scene. TP/TN rates measure detecting finished vs.\ unfinished subgoals. Tokens: generated per query over the four capabilities, reasoning included (Table~\ref{tab:vlm-tokens}).}
\label{tab:vlmbench}
\scriptsize
\begin{tabular}{lcccccc}
\toprule
 & \multicolumn{1}{c}{Planning} & \multicolumn{2}{c}{Subgoal completion} & \multicolumn{1}{c}{State tracking} & \multicolumn{1}{c}{Next subgoal} & \multicolumn{1}{c}{Tokens} \\
\cmidrule(lr){2-2}\cmidrule(lr){3-4}\cmidrule(lr){5-5}\cmidrule(lr){6-6}\cmidrule(lr){7-7}
Model & exact match (\%) & TP rate (\%) & TN rate (\%) & all correct (\%) & valid (\%) & per query \\
\midrule
Qwen3-VL-2B & 75.2\,\smallpm{7.6} & 68.2\,\smallpm{7.0} & 34.1\,\smallpm{5.9} & 9.3\,\smallpm{4.0} & 58.9\,\smallpm{5.3} & -- \\
Qwen3-VL-8B & 91.0\,\smallpm{5.8} & 69.0\,\smallpm{6.8} & 79.9\,\smallpm{5.7} & 47.9\,\smallpm{5.5} & 72.3\,\smallpm{4.4} & -- \\
Qwen3.5-9B & 86.9\,\smallpm{6.5} & 44.4\,\smallpm{6.3} & 96.9\,\smallpm{2.5} & 11.7\,\smallpm{3.7} & 63.2\,\smallpm{4.8} & 5{,}173 \\
Qwen3.8-27B & 91.0\,\smallpm{5.8} & \textbf{76.2\,\smallpm{6.3}} & 84.8\,\smallpm{4.3} & 58.2\,\smallpm{4.8} & 79.3\,\smallpm{4.1} & 1{,}904 \\
Gemini 3.1 Pro & 89.7\,\smallpm{6.0} & 54.8\,\smallpm{6.6} & 96.0\,\smallpm{2.9} & 68.7\,\smallpm{4.3} & 75.2\,\smallpm{4.5} & 583 \\
Gemini 3.7 Flash & \textbf{93.1\,\smallpm{5.4}} & 59.4\,\smallpm{6.6} & 96.6\,\smallpm{2.6} & \textbf{70.8\,\smallpm{4.0}} & 81.4\,\smallpm{4.2} & 412 \\
Gemini Robotics-ER 2 & 90.3\,\smallpm{5.9} & 54.0\,\smallpm{6.5} & \textbf{97.8\,\smallpm{2.3}} & 69.6\,\smallpm{4.2} & 76.6\,\smallpm{4.1} & 510 \\
GPT-5.6 Sol & \textbf{93.1\,\smallpm{5.4}} & 72.8\,\smallpm{6.3} & 86.4\,\smallpm{4.3} & 56.4\,\smallpm{4.7} & 82.0\,\smallpm{4.3} & 282 \\
GPT-6 Astra & 91.7\,\smallpm{5.7} & 71.1\,\smallpm{6.2} & 86.4\,\smallpm{4.4} & 53.2\,\smallpm{5.1} & \textbf{85.4\,\smallpm{4.0}} & 203 \\
Claude Opus 5 & 90.3\,\smallpm{5.9} & 68.6\,\smallpm{6.6} & 87.9\,\smallpm{4.1} & 61.0\,\smallpm{4.7} & 79.8\,\smallpm{4.4} & 400 \\
\bottomrule
\end{tabular}
\end{table}

On average \decMovedTrigger--\decMovedPeriodic\% of subgoals fail because an orchestrator without verification moves on before the subgoal is completed and never returns to it (Figure~\ref{fig:failure-horizon}; Table~\ref{tab:orchestrator-decisions} in the appendix).
This failure extends to \harness{}, which incorrectly marks \decMovedHarness\% of its subgoals as completed.
With the introduction of verification mechanisms, this failure mode decreases to \decMovedVerifier\% and orchestrator-related failures comprise only \decAllVerifier\% (\decAllVolo\% for VoLo).
This shifts the primary limitation of learned-policy systems to the low-level policy’s capabilities, with failure to manipulate subgoal target objects emerging as the main bottleneck (Low-level policy and VoLo).
This is followed by a failure to reach later subgoals in the plan (Proprioceptive verifier and Privileged verifier), because the method keeps retrying earlier ones.
We note that for these learned-policy methods, successful episodes on average complete within 50\% of the time budget (Figure~\ref{fig:efficiency}; Appendix~\ref{app:efficiency}), which indicates that the allocated time budget is sufficient.
Instead qualitative inspection reveals that the low-level policy ends up consuming most of the budget moving aimlessly and trying to reach the target object.
Long tasks therefore fail because progress stalls: half of the failed episodes complete no subgoal after the first \stallLastMedian\% of the budget.
Reaching the target object is also the primary failure mode in tool selection systems such as \capx{} and \gptpolicy{}, while \tiptop{} fails to find a plan for \fateNoPlanTiptop\% of its subgoals.

\subsection{Real-World Experiments}
\label{sec:exp:realrobot}

To test whether \benchname{} is indicative of real-world performance, we repeat the comparison on physical WidowX and Franka robots.
Due to the especially high costs and time requirements, we evaluate only a subset of the methods on task horizons ranging from 1 to 5.
We observe a similar trend to simulation: the low-level policy's success rate collapses much faster than orchestrator-based methods.
Furthermore, we observe preserved rankings of the methods. This is an encouraging result for using \benchname{} as a reliable rank-measuring tool for long-horizon tasks.

\begin{figure}[!ht]
\centering
\begin{minipage}[c]{0.47\linewidth}
\centering
\renewcommand{\arraystretch}{0.82}
\setlength{\tabcolsep}{1.5pt}
\scriptsize
\begin{tabular}{lcccc}
\toprule
Task category & Subgoals & \shortstack{Low-level\\policy} & \shortstack{Prop.\\verifier} & \harness{} \\
\midrule
\multicolumn{5}{l}{\emph{WidowX}} \\
\quad Single destination & 1 & 66.7 & 75.0 & \textbf{100.0} \\
\quad Single destination & 2--3 & 0.0 & 33.3 & \textbf{87.5} \\
\quad Split destination & 2--3 & 0.0 & 29.2 & \textbf{83.3} \\
\quad Exclusion & 2 & 0.0 & 58.3 & \textbf{91.7} \\
\quad Ordered & 2 & 8.3 & 33.3 & \textbf{41.7} \\
\quad Duplicates & 2 & 0.0 & 0.0 & \textbf{58.3} \\
\midrule
\multicolumn{5}{l}{\emph{Franka}} \\
\quad Single destination & 5 & 0.0 & 8.3 & \textbf{25.0} \\
\quad Exclusion & 2 & 0.0 & 83.3 & \textbf{100.0} \\
\quad Ordered & 3 & 16.7 & 83.3 & \textbf{100.0} \\
\bottomrule
\end{tabular}

\end{minipage}\hfill
\begin{minipage}[c]{0.51\linewidth}
\centering
\includegraphics[width=\linewidth]{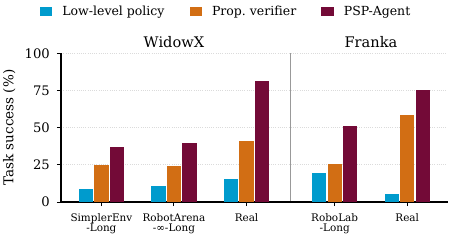}
\end{minipage}
\caption{\textbf{Real-robot evaluation.} Left: task success (\%) on the real robots per task category, 12 episodes per task and system. Right: overall task success in simulation (Table~\ref{tab:comp-overall}) and on the real robots. Both rank the three systems the same.}
\label{fig:realrobot}
\end{figure}

\section{Discussion and Conclusion}
\label{sec:discussion}
Our findings highlight three takeaways:
\textbf{(1)~Task horizon is a primary failure mode.} Low-level policies perform well on single-subgoal tasks, but collapse as horizon grows. High-level orchestration roughly triples overall multi-subgoal success.
\textbf{(2)~Physical execution is the main bottleneck.} VLMs decompose instructions accurately, but tasks often fail during execution even when given ground-truth subgoals.
\textbf{(3)~The two best systems execute with tool-based primitives.} \gptpolicy{} and \harness{} reach \gridGptAll\% and \gridHarnessAll\% success overall, against at most \gridOracleAll\% for the learned-policy systems.

\textbf{Limitations.} High-level VLM inference is slow, limiting reactive adjustments during rapid movements. Our evaluation focuses on tabletop pick-and-place rather than pushing, sliding, or articulated objects. Visual state tracking remains challenging under heavy clutter, and physical robot experiments cover fewer scenes than simulation.

\section*{Reproducibility Statement}
Section~\ref{sec:benchmarks} describes how the benchmark is built, and Appendix~\ref{sec:app:catalog} lists every task with its instruction, episodes and step budget. Section~\ref{sec:orchestrator} and Appendix~\ref{app:orch} specify every evaluated system and its settings, Appendix~\ref{sec:app:prompts} gives the prompts of our systems, and Appendix~\ref{app:policy_training} describes the low-level policies. We will release the long-horizon suites, the orchestrator code, the launch configurations and the offline VLM benchmark of Section~\ref{sec:exp:vlm}. Rollout videos are available on the project website.

\section*{AI Use Statement}
We used LLM-based coding and writing assistants. They helped implement the benchmark suites, the evaluated systems and the evaluation and analysis code; they helped analyze the results, review the methodology and check the paper's claims against the recorded results; and they helped draft, edit and format the text. The authors reviewed and tested all code, verified every number, claim and reference, and take full responsibility for the paper.

\section*{Ethics Statement}
This work involves no human subjects, personal data or crowdsourcing. We see no ethical concerns beyond the general safety considerations of deploying learned manipulation policies on physical robots.

\bibliography{references_arxiv}
\bibliographystyle{iclr2027_conference}

\clearpage
\appendix
\noindent\textbf{Appendix overview.} Appendix~\ref{app:policy_training} describes the low-level policies and compares \pifive{} with Cosmos-3; Appendix~\ref{app:orch} gives the configuration of every orchestrator and external system; Appendix~\ref{sec:app:prompts} prints the prompts; Appendix~\ref{sec:app:catalog} lists every task, its episodes and step budget; Appendix~\ref{app:realrobot} gives the real-robot results per task category; Appendix~\ref{app:per-suite} gives the results per suite and system, their uncertainty over tasks, and the construction of the offline VLM benchmark; Appendix~\ref{app:efficiency} covers robot time, VLM calls, tokens and cost; and Appendix~\ref{sec:app:failures} analyzes the failures of every system per suite.

\section{Low-Level Policy Architectures and Training Details}
\label{sec:appendix}
\label{app:policy_training}
\begin{table}[!ht]
\centering
\setlength{\tabcolsep}{5pt}
\caption{\textbf{The two DROID-trained low-level policies on RoboLab-Long.} Task success and subgoal progress (\%) of \pifive{}~\citep{intelligence2025pi05} and Cosmos-3~\citep{cosmos3} on the same tasks: with the low-level policy alone by number of subgoals (top), and under each orchestrator over all tasks (bottom); $\pm$ covers the 95\% Wilson interval, and a dash marks a configuration not run with Cosmos-3. The orchestrators in the main text use \pifive{}.}
\label{tab:base-policies}
\small
\sbox0{%
\begin{tabular}{lccccc}
\toprule
 & \multicolumn{2}{c}{Task success} & \multicolumn{2}{c}{Subgoal progress} & \multicolumn{1}{c}{} \\
\cmidrule(lr){2-3}\cmidrule(lr){4-5}
Tasks & \pifive{} & Cosmos-3 & \pifive{} & Cosmos-3 & Episodes \\
\midrule
\multicolumn{6}{l}{\emph{Low-level policy, by number of subgoals}} \\
1 subgoal & 100.0\,\smallpm{19.4} & 100.0\,\smallpm{19.4} & 100.0\,\smallpm{19.4} & 100.0\,\smallpm{19.4} & 16 \\
2 subgoals & 46.9\,\smallpm{12.1} & 12.5\,\smallpm{10.3} & 68.8\,\smallpm{8.5} & 39.8\,\smallpm{8.7} & 64 \\
3 subgoals & 22.5\,\smallpm{15.0} & 15.0\,\smallpm{14.1} & 63.3\,\smallpm{8.9} & 55.8\,\smallpm{8.9} & 40 \\
4 subgoals & 14.6\,\smallpm{12.6} & 12.5\,\smallpm{12.2} & 66.7\,\smallpm{6.9} & 49.5\,\smallpm{7.0} & 48 \\
5 subgoals & 0.0\,\smallpm{13.8} & 0.0\,\smallpm{13.8} & 63.3\,\smallpm{8.9} & 40.0\,\smallpm{8.9} & 24 \\
6 subgoals & 10.4\,\smallpm{11.8} & 6.2\,\smallpm{10.6} & 49.7\,\smallpm{5.7} & 44.4\,\smallpm{5.8} & 48 \\
7 subgoals & 0.0\,\smallpm{13.8} & 0.0\,\smallpm{13.8} & 44.6\,\smallpm{7.6} & 39.9\,\smallpm{7.6} & 24 \\
8 subgoals & 0.0\,\smallpm{10.7} & 0.0\,\smallpm{10.7} & 48.0\,\smallpm{6.1} & 32.0\,\smallpm{5.9} & 32 \\
9 subgoals & 0.0\,\smallpm{13.8} & 0.0\,\smallpm{13.8} & 53.2\,\smallpm{6.7} & 35.2\,\smallpm{6.6} & 24 \\
10 subgoals & 6.2\,\smallpm{13.9} & 0.0\,\smallpm{10.7} & 60.3\,\smallpm{5.5} & 25.6\,\smallpm{5.1} & 32 \\
\midrule
\multicolumn{6}{l}{\emph{All tasks, by orchestrator}} \\
Low-level policy & 19.6\,\smallpm{4.5} & 11.1\,\smallpm{3.7} & 56.6\,\smallpm{2.3} & 39.0\,\smallpm{2.3} & 352 \\
Periodic trigger & 25.6\,\smallpm{4.8} & -- & 57.1\,\smallpm{2.3} & -- & 352 \\
Proprioceptive trigger & 28.1\,\smallpm{4.9} & -- & 58.9\,\smallpm{2.3} & -- & 352 \\
Proprioceptive verifier & 25.6\,\smallpm{4.8} & -- & 56.1\,\smallpm{2.3} & -- & 352 \\
VoLo & 29.0\,\smallpm{4.9} & 25.6\,\smallpm{4.8} & 62.0\,\smallpm{2.3} & 52.5\,\smallpm{2.3} & 352 \\
Privileged verifier$^{\dagger}$ & 34.4\,\smallpm{5.1} & 22.2\,\smallpm{4.6} & 63.5\,\smallpm{2.2} & 59.3\,\smallpm{2.3} & 352 \\
\bottomrule
\end{tabular}
}\usebox0
\par\smallskip\parbox{\wd0}{\scriptsize $^{\dagger}$On tasks without a prescribed order, when a subgoal is not completed within a fixed step budget, the privileged verifier moves on to the next subgoal and returns to it later. The budget was set for \pifive{}, and Cosmos-3 needs more steps per subgoal than \pifive{}.}
\end{table}

We provide architecture, training, and deployment specifications for the base manipulation policies evaluated across our benchmarks:
\begin{itemize}
    \item \textbf{Bridge baseline (WidowX embodiment)}: Trained on the 43,851 trajectories of BridgeData~v2~\citep{walke2023bridgedata} that carry a language instruction. The model couples a \mbox{Qwen3-VL-2B} vision-language backbone (2.1B parameters) with a conditional flow-matching action expert (0.7B parameters, 2.8B in total). The action expert predicts end-effector deltas (translation and rotation) and a gripper command over an action chunk of 5 steps, of which the first 4 are executed before the next query. Training uses AdamW with a base learning rate of 5e-5, linear warmup for 250 steps, a batch size of 1024, and exponential moving average (EMA) checkpointing with decay 0.99 from step 20,000, trained for 30,000 steps. Control frequency is 5~Hz on SimplerEnv-Long and \robotarena{}-Long.
    \item \textbf{DROID baseline (Franka embodiment)}: Evaluates the public \pifive{} low-level policy~\citep{intelligence2025pi05} fine-tuned on the DROID dataset~\citep{khazatsky2024droid} ($\sim$76,000 demonstration trajectories with a 7-DoF Franka Panda robot), in its joint-position version. The model pairs a PaliGemma 3B vision-language backbone ($\sim$2.9B parameters) with a flow-matching action expert (3.35B parameters in total), predicting joint positions and the gripper state over action chunks of 15 steps, of which 8 are executed. Control frequency is 15~Hz on \robolab{}-Long.
    \item \textbf{Cosmos-3 baseline (Franka embodiment)}: Evaluates the public Cosmos3-Nano-Policy-DROID checkpoint~\citep{cosmos3}, the Cosmos-3 world model released as a policy for the DROID setup, on \robolab{}-Long. Like \pifive{}, it predicts joint positions and the gripper state from the camera views, the joint state, and the text instruction.
\end{itemize}

Table~\ref{tab:base-policies} compares the two DROID-trained policies on the same RoboLab-Long tasks. Cosmos-3 completes fewer subgoals than \pifive{} at almost every horizon, so the orchestrators in the main text use \pifive{}. Orchestration helps both policies: VoLo raises task success from \latestRlCosmos\% to \latestRlCosmosVolo\% with Cosmos-3 and from \latestRlFlat\% to \latestRlVolo\% with \pifive{}, and the privileged verifier reaches \latestRlCosmosOracle\% and \latestRlOracle\%.

\section{Orchestrator Configurations and External Systems}
\label{app:orch}
\label{app:external}
\paragraph{Periodic trigger.} It advances every \periodicStepsBridge{} steps on the WidowX suites and every \periodicStepsRobolab{} steps on \robolab{}-Long, and stays on the last subgoal.

\paragraph{Proprioceptive trigger.} The trigger tracks the normalized gripper aperture ($[0, 1]$ travel). An attempt ends when the gripper opens past $0.5$ after a close of at least 3 steps on the WidowX suites, and of at least 21 steps with a lift of 3\,cm on \robolab{}-Long. A cooldown of 8 steps (WidowX suites) or 57 steps (\robolab{}-Long) prevents re-triggering while the jaws settle. On \robolab{}-Long an attempt also ends after 300 steps without a release, which produces 68\% of its advances, and the plan restarts from its first subgoal after the last. The WidowX suites have no such limit.

\paragraph{Proprioceptive verifier.} It detects attempts in the same way, without the cooldown. A verification query is sent 5 steps (WidowX suites) or 35 steps (\robolab{}-Long) after a release, and after 45 or 300 steps without one. The verifier asks for an object count rather than a binary verdict, to disambiguate repeated identical objects: a subgoal is complete once the count reaches the number of times it occurs in the plan so far (one on revisit tasks). An incomplete subgoal is retried in place; on \robolab{}-Long, an incomplete subgoal with interchangeable duplicates moves to the end of the queue instead.

\paragraph{Privileged verifier.} It reads the simulator's success predicate for the current subgoal and advances once the predicate has held for a settle window, 3 steps on SimplerEnv-Long, 10 on \robotarena{}-Long and 120 on \robolab{}-Long (to filter contact chatter on container rims), with the gripper open. On \robolab{}-Long, a subgoal not completed within 600 steps is set aside and returned to later, except on ordered tasks; on the WidowX suites it is retried until the budget ends.

\paragraph{\harness{}.}
\label{app:harness}
\harness{} gives the VLM seven tools: \texttt{pick\_and\_place}, \texttt{pick}, \texttt{place}, \texttt{push}, \texttt{measure}, \texttt{grasp\_at} and \texttt{run\_code} (Appendix~\ref{sec:app:prompts}). The VLM plans the task from the current image as a list of goals, each naming one object and one destination. The first call for a goal is \texttt{pick\_and\_place} on that object and destination; after a failure, the VLM chooses the next call from the goal, the tool catalog and the reports of the calls so far. Each tool returns whether it succeeded and, on failure, a short description of the cause (for example, that the gripper closed without an object). A goal is given up for the current pass after two failed calls that move the arm. After a pass over all goals, the VLM plans again from the new image and the reported outcomes; an empty plan is checked once more, and a second empty plan ends the episode. To localize an object, a tool fits an upright box to the depth points inside the bounding box the VLM returns for its name. A grasp is checked with depth on the WidowX suites (the object has left its start position) and with the jaw width on \robolab{}-Long, and a release is checked with depth over the destination on SimplerEnv-Long and \robolab{}-Long.

\paragraph{External systems.} We evaluate four external baselines and alternative execution backends:
\begin{itemize}
    \item \textbf{VoLo}~\citep{chen2026volo}: Evaluated in its default subgoal mode, with the low-level policy as its only execution tool (the full system can also call action primitives). The VLM splits the task into an ordered list of subgoals from the first frame, checks progress every 20 steps on SimplerEnv-Long, 15 on \robotarena{}-Long and 150 on \robolab{}-Long, and replans the remaining task when a subgoal fails, at most three times.
    \item \textbf{\capx{}}~\citep{fu2026capx}: We evaluate the multi-turn variant of \capx{} with high-level primitives. A VLM generates a Python program over six primitives: \texttt{get\_object\_pose}, \texttt{sample\_grasp\_pose}, \texttt{goto\_pose}, \texttt{open\_gripper}, \texttt{close\_gripper}, and \texttt{home\_pose}. \texttt{get\_object\_pose} segments the named object with SAM3 in the scene camera, lifts the mask into 3D with the calibrated depth, and fits a box, rather than reading privileged simulator state; grasps are computed from the box. Unlike the release, the grasp is taken from straight above the box center rather than sampled with Contact-GraspNet, the box extends down to the support surface, and on \robolab{}-Long the arm ramps to each target within a clamped workspace. After each program, a second VLM query compares the frames before and after execution and decides whether to regenerate the program or finish.
    \item \textbf{\tiptop{}}~\citep{shen2026tiptop}: Performs open-loop task-and-motion planning from an initial scene perception. A VLM extracts 2D object bounding boxes and goal predicates, SAM-2 segments masks, metric depth reconstructs point clouds, M2T2 proposes grasp candidates, RANSAC estimates support planes, and cuTAMP with cuRobo solves for collision-free trajectory plans. Its planner finds a plan for most single-subgoal tasks but fails more often as subgoals are added: no plan is found for \tiptopNoPlanOne\% of its subgoals at one subgoal and \tiptopNoPlanFive\% at five.
    \item \textbf{\gptpolicy{}}~\citep{cheng2026incontextrobotlearningvlm}: A VLM agent that chooses one tool call at a time. We use its agent loop and tools (\texttt{move\_to}, \texttt{move\_eef\_chunk}, \texttt{set\_gripper}, \texttt{locate\_point}, \texttt{done}, and \texttt{give\_up}), with our description of the robot in its prompt. We answer \texttt{locate\_point}, for up to 8 pixels per call, with the calibrated depth camera. We disable \texttt{check\_path}, because the simulators do not provide the robot's inverse kinematics.
\end{itemize}

\paragraph{VLM settings.} In simulation, every hosted VLM query uses Gemini 3.7 Flash (\texttt{gemini-3.7-flash}), except those of \gptpolicy{}, which uses GPT-6 Astra (\texttt{gpt-6-astra}); all were queried between 10 and 24 September 2026. The Proprioceptive verifier, \harness{}, VoLo and \tiptop{} decode at temperature $0.0$, \capx{} at $1.0$, and \gptpolicy{} at the provider's default. Generation is capped at 2048 tokens for verification, for \harness{}'s planning and tool choice (1024 for grounding) and for \gptpolicy{}, at 8192 for VoLo and \tiptop{}, and at 20,480 for \capx{}. Reasoning effort is the provider's default, except \emph{low} for \tiptop{}. A verification query generates about 220 tokens, reasoning included.

\section{Vision-Language Prompts and Tool Catalog}
\label{sec:app:prompts}
We print the prompts of the two systems we built, as they ran: the verification query of the \emph{Proprioceptive verifier}, and the prompt templates and tool catalog of \harness{}. Words in angle brackets are filled in at run time. The external systems use their own prompts: VoLo and \tiptop{} those of their releases, \capx{} those of its multi-turn variant, and \gptpolicy{} its agent prompt with our description of the robot.
\paragraph{Completion verification.} Sent by the \emph{Proprioceptive verifier} with the current subgoal and frame on all three suites.
\begin{promptbox}
You are checking the CURRENT state of a robot tabletop scene from its camera views. The subtask below moves ONE object to a goal location, and the scene may hold SEVERAL interchangeable objects of that same kind. Count how many DISTINCT objects of the kind the subtask names are in the subtask's END STATE RIGHT NOW -- at the goal location the subtask asks for, regardless of how they got there. An object counts as placed when any part of it rests on the goal location, INCLUDING when partially hidden by the robot arm or other objects. An object the gripper is still holding does NOT count, however close to the goal location it is held: the release is not yet a placement and it is the release that most often fails. An object away from the goal location, or lifted well above it, is not counted. Count each physical object at most once, however many views show it. First write one short line listing what you counted, then output STRICT JSON: \{``count'': \textless{}integer\textgreater{}\}.
\end{promptbox}

\paragraph{Verification, container suffix.} Appended to every verification query.
\begin{promptbox}
When a goal location is a container whose interior you cannot see into, do not look for the object inside it. Instead account for every object of that kind in the scene: if an object of that kind is no longer resting anywhere else in the scene, and the gripper is not holding it, it is in the container and counts as placed. Objects still resting elsewhere do not count.
\end{promptbox}

\paragraph{Verification, multi-view suffix.} Appended when more than one camera is sent, which is the case on \robolab{}-Long (main and wrist views).
\begin{promptbox}
Several camera views of the SAME scene at the SAME instant are provided, each labelled. They do not show different moments. A view may not contain the goal location at all -- judge from whichever view shows it, and treat evidence from a close-up view as authoritative over a distant one where the goal location is small or occluded.
\end{promptbox}

\paragraph{\harness{}, planning.} Sent with the current frame at the start of the episode and after every pass over the plan. Goals already worked on are listed with the tools' reports in the form below.
\begin{promptbox}
You are planning for a \textless{}robot\textgreater{} working on a tabletop, seen from its camera. Task: \textless{}task\textgreater{}

\textless{}attempted\textgreater{}Break the task into the smallest goals that each move ONE object to ONE destination. Name every object by its colour and kind as it LOOKS in the image (``the yellow lemon'', ``the blue bin''), never by a role or a bare noun, because those names are what will be pointed at in the image later and several objects may be of the same kind. Only include goals the task actually asks for. An object the task does not mention must be left where it is, even when it sits among the ones that move. When the task states an order (``first ... then ...''), keep the goals in the order the task gives them, and list a goal TWICE if the task asks for the same object to be moved twice. Otherwise order the goals so that anything placed on an open surface, where a later reach could knock it off, comes after anything dropped into a container. Leave out any goal the image shows is already met, and return an empty list of goals when the whole task is done. Give each goal as a sentence, plus the object it moves and where that object goes, named the same way. Output STRICT JSON and nothing else: \{``goals'': [\{``goal'': ``...'', ``object'': ``...'', ``destination'': ``...''\}]\}
\end{promptbox}

\paragraph{\harness{}, attempted goals.} Filled into the planning prompt from the second pass on.
\begin{promptbox}
Already worked on, with what the executor reported. The image is the evidence of what actually happened, so ask for one of these again ONLY if the picture shows it did not take -- repeating a goal that took sends the arm back over it and knocks it over: \textless{}lines\textgreater{}
\end{promptbox}

\paragraph{\harness{}, re-check.} Appended to the planning prompt, which is sent again, when a plan comes back empty; a second empty plan ends the episode.
\begin{promptbox}
Look again before stopping. The episode is NOT over, and stopping early throws the rest of it away: on a two-goal SimplerEnv unit 57\% of the steps were spent standing still after the agent decided it was finished, while a third of the goals were still unmet. Name anything the task asks for that is not done, however small. Where the destination HIDES what goes into it -- a bin, a box, a pot you cannot see the bottom of -- the executor's report is the evidence and the picture is not. A goal reported done there IS done, and asking for it again sends the hand back in to take the object out. Where the destination is OPEN -- a plate, a cloth, a patch of table -- the picture is the evidence and the report is not. If the object is not lying on it, that goal is NOT done, whatever was reported: a released object rolls off, and a later reach knocks it off. Return an empty list only when every part of the task is either visible in the picture or reported done into something you cannot see inside.
\end{promptbox}

\paragraph{\harness{}, tool choice.} Sent with the current frame for every tool call after the first one of a goal; the catalog is listed below.
\begin{promptbox}
You are controlling a \textless{}robot\textgreater{}. The image is the scene right now. Goal: \textless{}goal\textgreater{} The gripper is \textless{}hand\textgreater{} right now.

Available tools: \textless{}catalog\textgreater{}

\textless{}history\textgreater{}Choose ONE tool call that makes progress on this goal. pick\_and\_place is the cheap path and handles most goals in one call. When it reports something the hand could not do, do not repeat it: the tools below it are yours -- measure the thing, then take hold of it where YOU judge the jaws will close, or push it there along the surface when it is too wide to hold at all, and place it yourself. run\_code drives the same API when a goal needs something none of the tools expresses. If a previous attempt failed, do not repeat it unchanged: name the OBJECT differently, since that name is what gets pointed at in the image and it may not match what is visible. The destination is what the goal asks for -- describe it differently if it was not found, but never put the object somewhere else. Output STRICT JSON and nothing else: \{``tool'': ``\textless{}name\textgreater{}'', ``args'': \{...\}\}
\end{promptbox}

\paragraph{\harness{}, grounding.} Sent with the current frame whenever a tool locates an object or a destination by name.
\begin{promptbox}
Point out \textless{}name\textgreater{} in this image. Draw it TIGHTLY around \textless{}name\textgreater{} alone. Anything lying beside or touching it must be OUTSIDE the box, even where that means cutting the box in close. Then add one box for EVERY other object lying near it, each with its own label, so \textless{}name\textgreater{} can be told apart from them. A name can say WHICH one it means by where it lies or what it is beside. Box the one it points at, and box the others of the same kind as well, so they can be told apart. Never reply with an empty list because the wording is not an exact name. Reply with ONLY a JSON list of boxes, each [ymin, xmin, ymax, xmax] normalised to 0-1000: [\{``box\_2d'': [y0, x0, y1, x1], ``label'': ``...''\}] Return an empty list [] only if nothing of that kind is in the picture at all.
\end{promptbox}

\paragraph{\harness{}, tool catalog.} The \emph{catalog} of the tool-choice prompt: each tool's signature and description.
\begin{promptbox}
pick\_and\_place(object\_name: str, destination\_name: str): Move one object to one destination. This is the whole of most goals; prefer it over calling pick and place separately.

pick(object\_name: str): Grasp an object and lift it clear of the surface. Name it the way it looks, with its colour (``the yellow lemon''), so the right one is grounded when several are similar. The jaws close on the widest part of it they can span, since that is the part a slightly wide approach still catches; use \textasciigrave{}grasp\_at\textasciigrave{} to take hold somewhere else.

place(destination\_name: str): Put the object already in the gripper onto or into a destination, named the same way. This needs a pick to have succeeded first; to move something still on the table, use pick\_and\_place. Releases above the destination's own top surface and on a clear part of it, so it works for a plate, for a cloth, and for the inside of a bin.

push(object\_name: str, destination\_name: str): Slide something along the surface to a destination with the closed hand, without lifting it. This is how to move what the jaws cannot take hold of: a pot or a bowl too wide to span, something too heavy, something flat with nothing to grip. It cannot put one thing INTO another or stack it on top, and it drags whatever stands in the way, so lift when you can and push when you cannot.

measure(object\_name: str, surface: bool = False): What the cameras and depth say about one thing, so you can decide where to take hold of it yourself. Reports where it is, how big it is, how wide and how thick it is along its length section by section, and what the hand can actually close on. The jaws hold a prism, not a scoop, so the widest part of a thing is not always the part to take it by. Pass surface=True for something to put things ON, like a plate or a cloth, which is too flat to ground otherwise.

grasp\_at(object\_name: str, x\_m: float, y\_m: float, z\_m: float, yaw\_deg: Optional[float] = None): Take hold of an object at a point YOU choose, then lift it. Use it when \textasciigrave{}pick\textasciigrave{} reports it cannot take something -- a pot too wide to span has a handle, a bowl has a rim, a spoon has a stem -- and \textasciigrave{}measure\textasciigrave{} to find where those are. \textasciigrave{}yaw\_deg\textasciigrave{} turns the hand about the vertical; leave it out to keep the object's own angle. Everything else is as \textasciigrave{}pick\textasciigrave{}: the jaws open on the way, close on the point, and the object is checked and lifted.

run\_code(code: str): Run Python against this same API when a goal needs something the other tools do not cover. \textasciigrave{}pick\textasciigrave{}, \textasciigrave{}place\textasciigrave{}, \textasciigrave{}pick\_and\_place\textasciigrave{}, \textasciigrave{}measure\textasciigrave{}, \textasciigrave{}grasp\_at\textasciigrave{}, \textasciigrave{}go\_home\textasciigrave{}, \textasciigrave{}locate(name, surface=False)\textasciigrave{} (returns an object with \textasciigrave{}center\_m\textasciigrave{}, \textasciigrave{}extent\_m\textasciigrave{}, \textasciigrave{}top\_m\textasciigrave{}, \textasciigrave{}yaw\_rad\textasciigrave{} and \textasciigrave{}cross\_section\_m()\textasciigrave{}, the width and height along its length; pass surface=True for something to put things ON, like a plate or a cloth, which is too flat to ground otherwise), \textasciigrave{}move\_to(position\_m, yaw\_rad)\textasciigrave{}, \textasciigrave{}open\_gripper\textasciigrave{}, \textasciigrave{}close\_gripper\textasciigrave{} and \textasciigrave{}np\textasciigrave{} are already defined. Every skill returns a report with \textasciigrave{}ok\textasciigrave{}, True when it finished, and \textasciigrave{}text\textasciigrave{}. Print what you learn.
\end{promptbox}

\clearpage
\section{Benchmark Suites and Complete Task Catalogs}
\label{sec:app:catalog}
We provide complete task specifications for all \tasksTotalAll{} evaluation tasks across the three benchmark suites: 54 tasks in SimplerEnv-Long (Table~\ref{tab:catalog-simpler}), 47 tasks in \robotarena{}-Long (Table~\ref{tab:catalog-ra}), and 44 tasks in \robolab{}-Long (Table~\ref{tab:catalog-robolab}); one \robotarena{}-Long task, \texttt{scene45\_all\_3}, repeats another (Table~\ref{tab:catalog-ra}). Table~\ref{tab:tasktypes} summarizes the eight task categories with representative instructions.

\begin{table}[H]
\centering
\caption{\textbf{The eight kinds of task}, with an example instruction for each. The kind of task decides what one instruction has to express, and so what an orchestrator can add. \emph{Tasks} counts the tasks of that kind across the three suites; Figure~\ref{fig:coverage} shows one frame per kind, with what its instruction asks for drawn on the scene.}
\label{tab:tasktypes}
\small
\begin{tabular}{l p{0.25\textwidth} p{0.32\textwidth} c}
\toprule
Task category & Requirement & Example instruction & Tasks \\
\midrule
Single destination & every object goes to the same destination & \emph{``put the carrot and the spoon on the plate''} & 44 \\
Transfer & every object starts on one destination and must be moved to another & \emph{``move the carrot and the spoon from the plate to the blue towel''} & 8 \\
Split destination & different objects go to different destinations & \emph{``put the carrot on the plate and put the spoon on the blue towel''} & 25 \\
Relay & objects pass through an intermediate destination & \emph{``first put the carrot on the plate, then put the carrot on the blue towel''} & 20 \\
Revisit & objects return to a destination they have already left & \emph{``first put the spoon on the cloth, then put the spoon in the pot, then put the spoon on the cloth''} & 13 \\
Exclusion & some objects must be left where they are & \emph{``put everything except the eggplant on the plate''} & 13 \\
Ordered & the placements must occur in a stated order & \emph{``first put the banana in the bin, then put the orange in the bin''} & 14 \\
Duplicates & identical objects cannot be told apart in language & \emph{``put two spoons on the plate and put one spoon on the blue towel''} & 8 \\
\bottomrule
\end{tabular}
\end{table}

\paragraph{Episodes and step budgets.} Each episode of a task starts from its own initial layout: SimplerEnv-Long draws 24 seeded layouts per task, \robotarena{}-Long shifts the objects of one authored layout by 8 fixed offsets of up to 3\,cm, and \robolab{}-Long jitters every object's position and yaw at reset. The step budget is \budgetStepsPerGoal{} steps per subgoal on the WidowX suites. On \robolab{}-Long it is 900, 1350, 2700, 3600, 4050, 4500 and 4950 steps for 1, 2, 3 to 6, 7, 8, 9 and 10 subgoals. In relay and revisit tasks a later leg of an object counts only once its earlier legs are done, and an ordered task fails when its objects are placed out of order. The primitives and plans of the tool selection systems execute as ordinary control steps, so their motions count against the same budget. Each task counts one run of each system, with all its episodes. A run that stopped before recording its results, or in which a check of the \robolab{}-Long videos for unrendered scenes flagged an episode, was repeated; where a task was nevertheless measured twice, only one of the runs is counted. This happened for \dupTasks{} tasks of a system, \dupTasksVolo{} of them VoLo's on \robolab{}-Long. Summed over each system's repeated tasks, the counted runs never succeed more often than the others, and only for VoLo does the choice move a suite's success by more than half a point: \latestRlVolo\% on \robolab{}-Long with its later runs counted, and \latestRlVoloEarly\% with the earlier ones. In total, the paper reports \compEpisodesTotal{} simulated episodes: \episodesPerEval{} for each of the ten systems of Table~\ref{tab:comp-overall} and \compEpisodesCosmos{} for the three Cosmos-3 configurations of Table~\ref{tab:base-policies}.

\begingroup\scriptsize\renewcommand{\arraystretch}{0.92}
\begin{longtable}{@{}>{\raggedright\arraybackslash}p{0.29\textwidth} c p{0.56\textwidth}@{}}
\caption{\textbf{Every SimplerEnv-Long task}, with the instruction the policy is given, grouped by kind of task and ordered by subgoal count.}\ifdefined\LaTeXML\else\label{tab:catalog-simpler}\fi\\
\toprule
Task & Subgoals & Instruction \\
\midrule
\endfirsthead
\multicolumn{3}{@{}l}{\emph{SimplerEnv-Long, continued}} \\
\toprule
Task & Subgoals & Instruction \\
\midrule
\endhead
\bottomrule
\endfoot
\multicolumn{3}{@{}l}{\emph{Single destination}} \\
\quad \texttt{all1\_plate} & 1 & ``put the carrot on the plate.'' \\
\quad \texttt{all1\_towel} & 1 & ``put the spoon on the blue towel.'' \\
\quad \texttt{all2\_plate} & 2 & ``put the carrot and the spoon on the plate'' \\
\quad \texttt{all2\_towel} & 2 & ``put the carrot and the spoon on the blue towel'' \\
\quad \texttt{all3\_plate} & 3 & ``put the carrot, the spoon and the eggplant on the plate'' \\
\quad \texttt{all3\_towel} & 3 & ``put the carrot, the spoon and the eggplant on the blue towel'' \\
\quad \texttt{all4\_plate} & 4 & ``put both carrots, the spoon and the eggplant on the plate'' \\
\quad \texttt{all4\_towel} & 4 & ``put both carrots, the spoon and the eggplant on the blue towel'' \\
\quad \texttt{all5\_plate} & 5 & ``put both carrots, both spoons and the eggplant on the plate'' \\
\quad \texttt{all5\_towel} & 5 & ``put both carrots, both spoons and the eggplant on the blue towel'' \\
\quad \texttt{all6\_plate} & 6 & ``put both carrots, both spoons and both eggplants on the plate'' \\
\quad \texttt{all6\_towel} & 6 & ``put both carrots, both spoons and both eggplants on the blue towel'' \\
\quad \texttt{all7\_plate} & 7 & ``put the three carrots, both spoons and both eggplants on the plate'' \\
\quad \texttt{all7\_towel} & 7 & ``put the three carrots, both spoons and both eggplants on the blue towel'' \\
\quad \texttt{all8\_plate} & 8 & ``put the three carrots, the three spoons and both eggplants on the plate'' \\
\quad \texttt{all8\_towel} & 8 & ``put the three carrots, the three spoons and both eggplants on the blue towel'' \\
\quad \texttt{all9\_plate} & 9 & ``put the three carrots, the three spoons and the three eggplants on the plate'' \\
\quad \texttt{all9\_towel} & 9 & ``put the three carrots, the three spoons and the three eggplants on the blue towel'' \\
\quad \texttt{all10\_plate} & 10 & ``put the four carrots, the three spoons and the three eggplants on the plate'' \\
\quad \texttt{all10\_towel} & 10 & ``put the four carrots, the three spoons and the three eggplants on the blue towel'' \\
\addlinespace
\multicolumn{3}{@{}l}{\emph{Transfer}} \\
\quad \texttt{move\_plate\_to\_towel\_2} & 2 & ``move the carrot and the spoon from the plate to the blue towel'' \\
\quad \texttt{move\_plate\_to\_towel\_3} & 3 & ``move the carrot, the spoon and the eggplant from the plate to the blue towel'' \\
\quad \texttt{move\_towel\_to\_plate\_3} & 3 & ``move the carrot, the spoon and the eggplant from the blue towel to the plate'' \\
\quad \texttt{move\_plate\_to\_towel\_4} & 4 & ``move both carrots, the spoon and the eggplant from the plate to the blue towel'' \\
\quad \texttt{move\_plate\_to\_towel\_5} & 5 & ``move both carrots, both spoons and the eggplant from the plate to the blue towel'' \\
\addlinespace
\multicolumn{3}{@{}l}{\emph{Split destination}} \\
\quad \texttt{split2} & 2 & ``put the carrot on the plate and put the spoon on the blue towel'' \\
\quad \texttt{split3} & 3 & ``put the carrot and the spoon on the plate and put the eggplant on the blue towel'' \\
\quad \texttt{split4} & 4 & ``put one carrot and the spoon on the plate and put the eggplant and one carrot on the blue towel'' \\
\quad \texttt{split5} & 5 & ``put one carrot, one spoon and the eggplant on the plate and put one carrot and one spoon on the blue towel'' \\
\quad \texttt{split6} & 6 & ``put one carrot, one spoon and one eggplant on the plate and put one carrot, one spoon and one eggplant on the blue towel'' \\
\quad \texttt{split7} & 7 & ``put two carrots, one spoon and one eggplant on the plate and put one spoon, one eggplant and one carrot on the blue towel'' \\
\addlinespace
\multicolumn{3}{@{}l}{\emph{Relay}} \\
\quad \texttt{relay2e} & 2 & ``first put the carrot on the plate, then put the carrot on the blue towel'' \\
\quad \texttt{relay3e} & 3 & ``first put the carrot on the plate, then put the carrot on the blue towel, then put the spoon on the plate'' \\
\quad \texttt{relay4e} & 4 & ``first put the carrot on the plate, then put the carrot on the blue towel, then put the spoon on the plate, then put the spoon on the blue towel'' \\
\quad \texttt{relay5e} & 5 & ``first put the carrot on the plate, then put the carrot on the blue towel, then put the spoon on the plate, then put the spoon on the blue towel, then put the eggplant on the plate'' \\
\quad \texttt{relay6e} & 6 & ``first put the carrot on the plate, then put the carrot on the blue towel, then put the spoon on the plate, then put the spoon on the blue towel, then put the eggplant on the plate, then put the eggplant on the blue towel'' \\
\quad \texttt{relay7e} & 7 & ``first put the carrot on the plate, then put the carrot on the blue towel, then put the spoon on the plate, then put the spoon on the blue towel, then put the eggplant on the plate, then put the eggplant on the blue towel, then put the carrot on the plate'' \\
\quad \texttt{relay8e} & 8 & ``first put the carrot on the plate, then put the carrot on the blue towel, then put the spoon on the plate, then put the spoon on the blue towel, then put the eggplant on the plate, then put the eggplant on the blue towel, then put the carrot on the plate, then put the carrot on the blue towel'' \\
\quad \texttt{relay9e} & 9 & ``first put the carrot on the plate, then put the carrot on the blue towel, then put the spoon on the plate, then put the spoon on the blue towel, then put the eggplant on the plate, then put the eggplant on the blue towel, then put the carrot on the plate, then put the carrot on the blue towel, then put the spoon on the plate'' \\
\quad \texttt{relay10e} & 10 & ``first put the carrot on the plate, then put the carrot on the blue towel, then put the spoon on the plate, then put the spoon on the blue towel, then put the eggplant on the plate, then put the eggplant on the blue towel, then put the carrot on the plate, then put the carrot on the blue towel, then put the spoon on the plate, then put the spoon on the blue towel'' \\
\addlinespace
\multicolumn{3}{@{}l}{\emph{Revisit}} \\
\quad \texttt{shuttle7} & 7 & ``first put the carrot on the plate, then put the spoon on the plate, then put the eggplant on the plate, then put the carrot on the blue towel, then put the spoon on the blue towel, then put the eggplant on the blue towel, then put the carrot on the plate'' \\
\quad \texttt{shuttle8} & 8 & ``first put the carrot on the plate, then put the spoon on the plate, then put the eggplant on the plate, then put the carrot on the blue towel, then put the spoon on the blue towel, then put the eggplant on the blue towel, then put the carrot on the plate, then put the spoon on the plate'' \\
\quad \texttt{shuttle9} & 9 & ``first put the carrot on the plate, then put the spoon on the plate, then put the eggplant on the plate, then put the carrot on the blue towel, then put the spoon on the blue towel, then put the eggplant on the blue towel, then put the carrot on the plate, then put the spoon on the plate, then put the eggplant on the plate'' \\
\quad \texttt{shuttle10} & 10 & ``first put the carrot on the plate, then put the spoon on the plate, then put the eggplant on the plate, then put the carrot on the blue towel, then put the spoon on the blue towel, then put the eggplant on the blue towel, then put the carrot on the plate, then put the spoon on the plate, then put the eggplant on the plate, then put the carrot on the blue towel'' \\
\addlinespace
\multicolumn{3}{@{}l}{\emph{Exclusion}} \\
\quad \texttt{except\_eggplant\_\{2,3,4,5\}} & 2--5 & ``put everything except the eggplant on the plate'' \\
\addlinespace
\multicolumn{3}{@{}l}{\emph{Ordered}} \\
\quad \texttt{ordered\_plate\_2} & 2 & ``first put the carrot on the plate, then put the spoon on the plate'' \\
\quad \texttt{ordered\_plate\_3} & 3 & ``first put the carrot on the plate, then put the spoon on the plate, then put the eggplant on the plate'' \\
\addlinespace
\multicolumn{3}{@{}l}{\emph{Duplicates}} \\
\quad \texttt{spoons2} & 2 & ``put one spoon on the plate and put one spoon on the blue towel'' \\
\quad \texttt{spoons3} & 3 & ``put two spoons on the plate and put one spoon on the blue towel'' \\
\quad \texttt{spoons4} & 4 & ``put two spoons on the plate and put two spoons on the blue towel'' \\
\quad \texttt{spoons5} & 5 & ``put three spoons on the plate and put two spoons on the blue towel'' \\
\end{longtable}\ifdefined\LaTeXML\label{tab:catalog-simpler}\fi
\endgroup

\begingroup\scriptsize\renewcommand{\arraystretch}{0.92}
\begin{longtable}{@{}>{\raggedright\arraybackslash}p{0.29\textwidth} c p{0.56\textwidth}@{}}
\caption{\textbf{Every \robotarena{}-Long task}, with the instruction the policy is given, grouped by kind of task and ordered by subgoal count. The \texttt{scene45\_all\_N} tasks give the instruction of \texttt{scene45\_Ngoal}, word for word for $N=2,3$ and with two objects named in the opposite order for $N=4$ to $6$. For $N=3$ to $6$ they also share its scene, so \texttt{scene45\_all\_3} repeats \texttt{scene45\_3goal}; \texttt{scene45\_all\_2} has a larger cloth and moved objects. In \texttt{scene54\_transfer\_2} both objects start on the table rather than on the cloth.}\ifdefined\LaTeXML\else\label{tab:catalog-ra}\fi\\
\toprule
Task & Subgoals & Instruction \\
\midrule
\endfirsthead
\multicolumn{3}{@{}l}{\emph{\robotarena{}-Long, continued}} \\
\toprule
Task & Subgoals & Instruction \\
\midrule
\endhead
\bottomrule
\endfoot
\multicolumn{3}{@{}l}{\emph{Single destination}} \\
\quad \texttt{scene36\_1goal} & 1 & ``put the fork in the pan.'' \\
\quad \texttt{scene45\_1goal} & 1 & ``put the blue bowl on the purple cloth.'' \\
\quad \texttt{scene54\_1goal} & 1 & ``put the spoon on the cloth.'' \\
\quad \texttt{scene76\_1goal} & 1 & ``put the metal pot on the yellow cloth.'' \\
\quad \texttt{scene36\_2goal} & 2 & ``put the fork and the corn cob in the pan'' \\
\quad \texttt{scene45\_2goal}, \texttt{scene45\_all\_2} & 2 & ``put the blue bowl and the sushi toy on the purple cloth'' \\
\quad \texttt{scene54\_2goal} & 2 & ``put the spoon and the chicken drumstick on the cloth'' \\
\quad \texttt{scene76\_2goal} & 2 & ``put the metal pot and the mushroom toy on the yellow cloth'' \\
\quad \texttt{scene45\_3goal}, \texttt{scene45\_all\_3} & 3 & ``put the blue bowl, the sushi toy and the banana on the purple cloth'' \\
\quad \texttt{scene50\_3goal} & 3 & ``put the stuffed animal, the blue block and the cheese in the pot'' \\
\quad \texttt{scene54\_3goal} & 3 & ``put the spoon, the chicken drumstick and the pot on the cloth'' \\
\quad \texttt{scene45\_4goal} & 4 & ``put the blue bowl, the sushi toy, the cheese and the banana on the purple cloth'' \\
\quad \texttt{scene45\_all\_4} & 4 & ``put the blue bowl, the sushi toy, the banana and the cheese on the purple cloth'' \\
\quad \texttt{scene45\_5goal} & 5 & ``put the blue bowl, the sushi toy, the cheese, the banana and the mushroom on the purple cloth'' \\
\quad \texttt{scene45\_all\_5} & 5 & ``put the blue bowl, the sushi toy, the banana, the cheese and the mushroom on the purple cloth'' \\
\quad \texttt{scene45\_6goal} & 6 & ``put the blue bowl, the sushi toy, the cheese, the banana, the mushroom and the corn cob on the purple cloth'' \\
\quad \texttt{scene45\_all\_6} & 6 & ``put the blue bowl, the sushi toy, the banana, the cheese, the mushroom and the corn cob on the purple cloth'' \\
\addlinespace
\multicolumn{3}{@{}l}{\emph{Transfer}} \\
\quad \texttt{scene54\_transfer\_2} & 2 & ``move the spoon and the chicken drumstick from the cloth to the pot'' \\
\addlinespace
\multicolumn{3}{@{}l}{\emph{Split destination}} \\
\quad \texttt{scene54\_split\_2} & 2 & ``put the spoon on the cloth and put the chicken drumstick in the pot'' \\
\quad \texttt{scene54\_split\_3} & 3 & ``put the spoon and the corn cob on the cloth and put the chicken drumstick in the pot'' \\
\addlinespace
\multicolumn{3}{@{}l}{\emph{Relay}} \\
\quad \texttt{scene54\_relay\_2} & 2 & ``first put the spoon on the cloth, then put the spoon in the pot'' \\
\quad \texttt{scene54\_route\_2} & 2 & ``first put the spoon in the pot, then put the spoon on the cloth'' \\
\quad \texttt{scene54\_relay\_4} & 4 & ``first put the spoon on the cloth, then put the spoon in the pot, then put the chicken drumstick on the cloth, then put the chicken drumstick in the pot'' \\
\quad \texttt{scene54\_route\_4} & 4 & ``first put the spoon in the pot, then put the spoon on the cloth, then put the chicken drumstick in the pot, then put the chicken drumstick on the cloth'' \\
\quad \texttt{scene54\_route\_6} & 6 & ``first put the spoon in the pot, then put the spoon on the cloth, then put the chicken drumstick in the pot, then put the chicken drumstick on the cloth, then put the corn cob in the pot, then put the corn cob on the cloth'' \\
\quad \texttt{scene54\_route\_7} & 7 & ``first put the spoon in the pot, then put the spoon on the cloth, then put the chicken drumstick in the pot, then put the chicken drumstick on the cloth, then put the corn cob in the pot, then put the corn cob on the cloth, then put the cheese on the cloth'' \\
\quad \texttt{scene54\_route\_8} & 8 & ``first put the spoon in the pot, then put the spoon on the cloth, then put the chicken drumstick in the pot, then put the chicken drumstick on the cloth, then put the corn cob in the pot, then put the corn cob on the cloth, then put the cheese in the pot, then put the cheese on the cloth'' \\
\addlinespace
\multicolumn{3}{@{}l}{\emph{Revisit}} \\
\quad \texttt{scene54\_revisit\_3} & 3 & ``first put the spoon on the cloth, then put the spoon in the pot, then put the spoon on the cloth'' \\
\quad \texttt{scene54\_revisit\_6} & 6 & ``first put the spoon on the cloth, then put the spoon in the pot, then put the spoon on the cloth, then put the chicken drumstick on the cloth, then put the chicken drumstick in the pot, then put the chicken drumstick on the cloth'' \\
\quad \texttt{scene54\_revisit\_7} & 7 & ``first put the spoon on the cloth, then put the spoon in the pot, then put the spoon on the cloth, then put the spoon in the pot, then put the spoon on the cloth, then put the spoon in the pot, then put the spoon on the cloth'' \\
\quad \texttt{scene54\_revisit\_8} & 8 & ``first put the spoon on the cloth, then put the spoon in the pot, then put the spoon on the cloth, then put the spoon in the pot, then put the spoon on the cloth, then put the spoon in the pot, then put the spoon on the cloth, then put the spoon in the pot'' \\
\quad \texttt{scene54\_revisit\_9} & 9 & ``first put the spoon on the cloth, then put the spoon in the pot, then put the spoon on the cloth, then put the chicken drumstick on the cloth, then put the chicken drumstick in the pot, then put the chicken drumstick on the cloth, then put the corn cob on the cloth, then put the corn cob in the pot, then put the corn cob on the cloth'' \\
\quad \texttt{scene54\_revisit\_10} & 10 & ``first put the spoon on the cloth, then put the spoon in the pot, then put the spoon on the cloth, then put the spoon in the pot, then put the spoon on the cloth, then put the spoon in the pot, then put the spoon on the cloth, then put the spoon in the pot, then put the spoon on the cloth, then put the spoon in the pot'' \\
\addlinespace
\multicolumn{3}{@{}l}{\emph{Exclusion}} \\
\quad \texttt{scene45\_except\_\{2,3,4,5,6\}} & 2--6 & ``put everything except the small can on the purple cloth'' \\
\quad \texttt{scene54\_except}, \texttt{scene54\_except\_3} & 2, 3 & ``put everything except the pot on the cloth'' \\
\addlinespace
\multicolumn{3}{@{}l}{\emph{Ordered}} \\
\quad \texttt{scene54\_2goal\_ordered} & 2 & ``first put the chicken drumstick on the cloth, then put the spoon on the cloth'' \\
\quad \texttt{scene76\_2goal\_ordered} & 2 & ``first put the metal pot on the yellow cloth, then put the mushroom toy on the yellow cloth'' \\
\quad \texttt{scene54\_3goal\_ordered} & 3 & ``first put the corn cob on the cloth, then put the spoon on the cloth, then put the chicken drumstick on the cloth'' \\
\addlinespace
\multicolumn{3}{@{}l}{\emph{Duplicates}} \\
\quad \texttt{scene54\_duplicates\_2} & 2 & ``put both corn cobs on the cloth'' \\
\quad \texttt{scene54\_duplicates\_3} & 3 & ``put the three corn cobs on the cloth'' \\
\end{longtable}\ifdefined\LaTeXML\label{tab:catalog-ra}\fi
\endgroup

\begingroup\scriptsize\renewcommand{\arraystretch}{0.92}
\begin{longtable}{@{}>{\raggedright\arraybackslash}p{0.29\textwidth} c p{0.56\textwidth}@{}}
\caption{\textbf{Every \robolab{}-Long task}, with the instruction the policy is given, grouped by kind of task and ordered by subgoal count. The suffix of \texttt{comp\_relay\_N} and \texttt{comp\_revisit\_N} counts objects, so these tasks have $2N$ and $3N$ subgoals.}\ifdefined\LaTeXML\else\label{tab:catalog-robolab}\fi\\
\toprule
Task & Subgoals & Instruction \\
\midrule
\endfirsthead
\multicolumn{3}{@{}l}{\emph{RoboLab-Long, continued}} \\
\toprule
Task & Subgoals & Instruction \\
\midrule
\endhead
\bottomrule
\endfoot
\multicolumn{3}{@{}l}{\emph{Single destination}} \\
\quad \texttt{comp\_bin\_1} & 1 & ``put the banana in the bin.'' \\
\quad \texttt{comp\_plate\_1} & 1 & ``put the banana on the plate.'' \\
\quad \texttt{comp\_bin\_2} & 2 & ``put the banana and the orange in the bin'' \\
\quad \texttt{comp\_bin\_6} & 6 & ``put the banana, the orange, the lemon, the lime, the avocado and the onion in the bin'' \\
\quad \texttt{comp\_bin\_10} & 10 & ``put the banana, both oranges, the lemon, the lime, the avocado, the onion, the lychee, the pumpkin and the pomegranate in the bin'' \\
\addlinespace
\multicolumn{3}{@{}l}{\emph{Transfer}} \\
\quad \texttt{comp\_transfer\_2} & 2 & ``move the orange and the lemon from the plate to the bin'' \\
\quad \texttt{comp\_transfer\_3} & 3 & ``move the orange, the lemon and the lime from the plate to the bin'' \\
\addlinespace
\multicolumn{3}{@{}l}{\emph{Split destination}} \\
\quad \texttt{comp\_route\_2} & 2 & ``put the lemon on the plate and put the orange in the bin'' \\
\quad \texttt{comp\_route\_2\_binfirst} & 2 & ``put the orange in the bin and put the lemon on the plate'' \\
\quad \texttt{comp\_split\_plate\_2} & 2 & ``put the avocado on the plate and put the orange in the bin'' \\
\quad \texttt{comp\_route\_3} & 3 & ``put the lemon and the lime on the plate and put the orange in the bin'' \\
\quad \texttt{comp\_split\_plate\_3} & 3 & ``put the avocado and the banana on the plate and put the orange in the bin'' \\
\quad \texttt{comp\_route\_4} & 4 & ``put the lemon and the lime on the plate and put the orange and the avocado in the bin'' \\
\quad \texttt{comp\_split\_plate\_4} & 4 & ``put the avocado and the banana on the plate and put the orange and the lime in the bin'' \\
\quad \texttt{comp\_route\_5} & 5 & ``put the lemon, the lime and the banana on the plate and put the orange and the avocado in the bin'' \\
\quad \texttt{comp\_split\_plate\_5} & 5 & ``put the avocado and the banana on the plate and put the orange, the onion and the lychee in the bin'' \\
\quad \texttt{comp\_route\_6} & 6 & ``put the lemon, the lime and the banana on the plate and put the orange, the avocado and the onion in the bin'' \\
\quad \texttt{comp\_split\_plate\_6} & 6 & ``put the avocado, the banana and the lemon on the plate and put the orange, the lime and the onion in the bin'' \\
\quad \texttt{comp\_route\_7} & 7 & ``put the banana, the lemon and the lime on the plate and put the orange, the avocado, the onion and the lychee in the bin'' \\
\quad \texttt{comp\_split\_plate\_7} & 7 & ``put the avocado, the banana and the lemon on the plate and put the orange, the lime, the onion and the lychee in the bin'' \\
\quad \texttt{comp\_route\_8} & 8 & ``put the banana, the lemon and the lime on the plate and put the orange, the avocado, the onion, the lychee and the pumpkin in the bin'' \\
\quad \texttt{comp\_split\_plate\_8} & 8 & ``put the avocado, the banana, the lemon and the lime on the plate and put the orange, the onion, the lychee and the pumpkin in the bin'' \\
\quad \texttt{comp\_route\_9} & 9 & ``put the banana, the lemon and the lime on the plate and put both oranges, the avocado, the onion, the lychee and the pumpkin in the bin'' \\
\quad \texttt{comp\_route\_10} & 10 & ``put the banana, the lemon and the lime on the plate and put both oranges, the avocado, the onion, the lychee, the pumpkin and the pomegranate in the bin'' \\
\addlinespace
\multicolumn{3}{@{}l}{\emph{Relay}} \\
\quad \texttt{comp\_relay\_2} & 4 & ``first put the lemon on the plate, then put the lemon in the bin, then put the lime on the plate, then put the lime in the bin'' \\
\quad \texttt{comp\_relay\_3} & 6 & ``first put the lemon on the plate, then put the lemon in the bin, then put the lime on the plate, then put the lime in the bin, then put the banana on the plate, then put the banana in the bin'' \\
\quad \texttt{comp\_relay\_4} & 8 & ``first put the lemon on the plate, then put the lemon in the bin, then put the lime on the plate, then put the lime in the bin, then put the banana on the plate, then put the banana in the bin, then put the orange on the plate, then put the orange in the bin'' \\
\quad \texttt{comp\_relay\_5} & 10 & ``first put the lemon on the plate, then put the lemon in the bin, then put the lime on the plate, then put the lime in the bin, then put the banana on the plate, then put the banana in the bin, then put the orange on the plate, then put the orange in the bin, then put the avocado on the plate, then put the avocado in the bin'' \\
\addlinespace
\multicolumn{3}{@{}l}{\emph{Revisit}} \\
\quad \texttt{comp\_revisit\_1} & 3 & ``first put the lemon in the bin, then put the lemon on the plate, then put the lemon in the bin'' \\
\quad \texttt{comp\_revisit\_2} & 6 & ``first put the lemon in the bin, then put the lemon on the plate, then put the lemon in the bin, then put the lime in the bin, then put the lime on the plate, then put the lime in the bin'' \\
\quad \texttt{comp\_revisit\_3} & 9 & ``first put the lemon in the bin, then put the lemon on the plate, then put the lemon in the bin, then put the lime in the bin, then put the lime on the plate, then put the lime in the bin, then put the banana in the bin, then put the banana on the plate, then put the banana in the bin'' \\
\addlinespace
\multicolumn{3}{@{}l}{\emph{Exclusion}} \\
\quad \texttt{comp\_bin\_2\_distractor} & 2 & ``put everything except the red block in the bin'' \\
\quad \texttt{comp\_bin\_4\_distractor} & 4 & ``put everything except the red block and the blue block in the bin'' \\
\addlinespace
\multicolumn{3}{@{}l}{\emph{Ordered}} \\
\quad \texttt{comp\_ordered\_bin\_2} & 2 & ``first put the banana in the bin, then put the orange in the bin'' \\
\quad \texttt{comp\_ordered\_bin\_3} & 3 & ``first put the banana in the bin, then put the lemon in the bin, then put the orange in the bin'' \\
\quad \texttt{comp\_ordered\_bin\_4} & 4 & ``first put the orange in the bin, then put the lemon in the bin, then put the lime in the bin, then put the banana in the bin'' \\
\quad \texttt{comp\_ordered\_bin\_5} & 5 & ``first put the avocado in the bin, then put the orange in the bin, then put the lemon in the bin, then put the lime in the bin, then put the banana in the bin'' \\
\quad \texttt{comp\_ordered\_bin\_6} & 6 & ``first put the avocado in the bin, then put the orange in the bin, then put the lemon in the bin, then put the lime in the bin, then put the onion in the bin, then put the banana in the bin'' \\
\quad \texttt{comp\_ordered\_bin\_7} & 7 & ``first put the avocado in the bin, then put the orange in the bin, then put the lemon in the bin, then put the lime in the bin, then put the onion in the bin, then put the banana in the bin, then put the lychee in the bin'' \\
\quad \texttt{comp\_ordered\_bin\_8} & 8 & ``first put the orange in the bin, then put the pumpkin in the bin, then put the lychee in the bin, then put the lemon in the bin, then put the lime in the bin, then put the banana in the bin, then put the onion in the bin, then put the avocado in the bin'' \\
\quad \texttt{comp\_ordered\_bin\_9} & 9 & ``first put the onion in the bin, then put the orange in the bin, then put the pumpkin in the bin, then put the lychee in the bin, then put the lemon in the bin, then put the lime in the bin, then put the banana in the bin, then put the avocado in the bin, then put the orange in the bin'' \\
\quad \texttt{comp\_ordered\_bin\_10} & 10 & ``first put the pumpkin in the bin, then put the onion in the bin, then put the lime in the bin, then put the avocado in the bin, then put the lemon in the bin, then put the pomegranate in the bin, then put the lychee in the bin, then put the banana in the bin, then put the orange in the bin, then put the orange in the bin'' \\
\addlinespace
\multicolumn{3}{@{}l}{\emph{Duplicates}} \\
\quad \texttt{comp\_identical\_2} & 2 & ``put both oranges in the bin'' \\
\quad \texttt{comp\_identical\_4} & 4 & ``put the four limes in the bin'' \\
\end{longtable}\ifdefined\LaTeXML\label{tab:catalog-robolab}\fi
\endgroup

\section{Real-Robot Results per Task Category}
\label{app:realrobot}
Table~\ref{tab:realrobot} breaks the real-robot results of Section~\ref{sec:exp:realrobot} down by task category.

\begin{table}[!ht]
\centering
\renewcommand{\arraystretch}{0.90}
\caption{\textbf{Real-robot evaluation per task category.} Subgoal progress (\%) and task success rate (\%) on the WidowX (Bridge baseline) and Franka (DROID baseline) physical platforms, 12 episodes per task and system (3 layouts, 4 trials each). \harness{} ran a month after the other two systems on the same tasks, with layouts restored from the same reference images. On the Franka single-destination task the low-level policy and the Proprioceptive verifier ran with 150 steps per subgoal (950 steps, with 200 more for the first subgoal) and \harness{} with 300 (1,700 steps); every \harness{} success there placed its fourth object after step 1,200. \harness{} ran that task twice, with no reason for the repeat recorded, and its first run (0/12 successes, 38.3\% progress) is not included. On the real robots the Proprioceptive verifier used Gemini Robotics-ER~2 for planning and verification, and \harness{} Gemini 3.7 Flash. $^\dagger$Run with \harness{} only, and not part of the \emph{All multi-subgoal} row.}
\label{tab:realrobot}
\scriptsize
\begin{tabular}{lcccccccc}
\toprule
 & \multicolumn{2}{c}{} & \multicolumn{2}{c}{Low-level policy} & \multicolumn{2}{c}{Proprioceptive verifier} & \multicolumn{2}{c}{\harness{}} \\
\cmidrule(lr){4-5}\cmidrule(lr){6-7}\cmidrule(lr){8-9}
Task category & Subgoals & Episodes & Progress & Success & Progress & Success & Progress & Success \\
\midrule
\multicolumn{9}{l}{\emph{Embodiment A (WidowX, Bridge baseline)}} \\
\quad 1 subgoal, control & 1 & 24 & 66.7 & 66.7 & 75.0 & 75.0 & 100.0 & \textbf{100.0} \\
\quad Single destination & 2--3 & 24 & 18.3 & 0.0 & 56.7 & 33.3 & 95.0 & \textbf{87.5} \\
\quad Split destination & 2--3 & 24 & 21.7 & 0.0 & 63.3 & 29.2 & 86.7 & \textbf{83.3} \\
\quad Exclusion & 2 & 12 & 8.3 & 0.0 & 62.5 & 58.3 & 95.8 & \textbf{91.7} \\
\quad Ordered & 2 & 12 & 29.2 & 8.3 & 50.0 & 33.3 & 70.8 & \textbf{41.7} \\
\quad Duplicates & 2 & 12 & 8.3 & 0.0 & 37.5 & 0.0 & 70.8 & \textbf{58.3} \\
\quad \emph{All multi-subgoal} & 2--3 & 84 & 18.2 & 1.2 & 56.2 & 31.0 & 86.5 & \textbf{76.2} \\
\midrule
\multicolumn{9}{l}{\emph{Embodiment B (Franka, DROID baseline)}} \\
\quad Single destination & 5 & 12 & 65.0 & 0.0 & 65.0 & 8.3 & 56.7 & \textbf{25.0} \\
\quad Exclusion & 2 & 12 & 0.0 & 0.0 & 91.7 & 83.3 & 100.0 & \textbf{100.0} \\
\quad Ordered & 3 & 12 & 38.9 & 16.7 & 88.9 & 83.3 & 100.0 & \textbf{100.0} \\
\quad \emph{All multi-subgoal} & 2--5 & 36 & 44.2 & 5.6 & 77.5 & 58.3 & 78.3 & \textbf{75.0} \\
\quad Split destination$^\dagger$ & 3 & 12 & -- & -- & -- & -- & 97.2 & 91.7 \\
\quad Relay$^\dagger$ & 4 & 12 & -- & -- & -- & -- & 93.8 & 91.7 \\
\bottomrule
\end{tabular}
\end{table}

\section{Results per Suite and System}
\label{app:per-suite}
The main text combines some results over the three suites or draws only some of the systems. This appendix gives every one of them per suite and for all ten systems, over the same episodes. Table~\ref{tab:suite-horizon-success} and Table~\ref{tab:suite-horizon-progress} are the numbers behind Figure~\ref{fig:ladder}, Table~\ref{tab:suite-category-success} splits the task categories of Table~\ref{tab:comp-overall} by suite, and Table~\ref{tab:vlmbench-suites} splits the offline VLM benchmark of Table~\ref{tab:vlmbench} by the suite each scene comes from, without intervals, since each suite holds about a third of the samples. Failure causes per suite are in Appendix~\ref{sec:app:failures} and efficiency per suite in Appendix~\ref{app:efficiency}.

\paragraph{Uncertainty over tasks.} The tasks are the benchmark, so the intervals of the tables cover the randomness of the episodes on these tasks. For a fixed set of tasks the Wilson interval is conservative, because the variance within tasks is smaller than the variance of all episodes taken as one sample. A claim that one system is better than another should also hold across tasks. Since all systems run the same tasks, we test each such difference paired: tasks are resampled within each suite (10,000 bootstrap draws) and both systems are scored on the same resampled tasks, so the difficulty of a task counts for both. The overall differences the main text draws, in points with their 95\% intervals, are \gptpolicy{} over \harness{} \bootDiffGptHarness{}, \harness{} over the privileged verifier \bootDiffHarnessOracle{}, the privileged verifier over the \emph{Proprioceptive verifier} \bootDiffOracleVer{} and over VoLo \bootDiffOracleVolo{}, the \emph{Proprioceptive verifier} over the \emph{Proprioceptive trigger} \bootDiffVerTrigger{} and over the low-level policy \bootDiffVerFlat{}. VoLo and the \emph{Proprioceptive verifier} (\bootDiffVoloVer{}), \capx{} and the \emph{Proprioceptive verifier} (\bootDiffCapxVer{}), and the low-level policy and \tiptop{} (\bootDiffFlatTiptop{}) are not separated.

\paragraph{Offline VLM benchmark.} Plan decomposition is scored on \tasksTotalAll{} scenes, the reset frame and instruction of every task; a plan is an exact match when its object and destination pairs are exactly those of the task's ground-truth subgoals, with their multiplicity, in any order. Requiring their order on the \tasksOrderDependent{} ordered, relay and revisit tasks changes no score in Table~\ref{tab:vlmbench}. Verification, state tracking and next-subgoal selection use \decompBenchStateN{} frames recorded in rollouts of 143 of these tasks, mostly of the privileged verifier, \capx{} and \tiptop{}, labelled by the simulator's per-subgoal predicates. The TP and TN rates are the shares of finished and unfinished subgoals judged correctly; state tracking asks about every subgoal of a frame and counts the frame as correct only when all answers are; a next subgoal is valid when it names a subgoal that is still unfinished, and on those \tasksOrderDependent{} tasks the first unfinished one. Table~\ref{tab:vlmbench-suites} splits these scores by suite.

\begin{table}[!ht]
\centering
\setlength{\tabcolsep}{3pt}
\caption{\textbf{Task success rate (\%) per suite, system and number of subgoals} (Figure~\ref{fig:ladder}, top).}
\label{tab:suite-horizon-success}
\scriptsize
\resizebox{\linewidth}{!}{%
\begin{tabular}{lcccc|cccc|c|c|r}
\toprule
Tasks & \shortstack{Low-level\\policy} & \shortstack{Periodic\\trigger} & \shortstack{Proprioceptive\\trigger} & \shortstack{Proprioceptive\\verifier} & VoLo & \capx{} & \tiptop{} & \harness{} & \gptpolicy{} & \shortstack{Privileged\\verifier} & Episodes \\
\midrule
\multicolumn{12}{l}{\textbf{SimplerEnv-Long}} \\
1 subgoal & 89.6 & 91.7 & 91.7 & 89.6 & \textbf{93.8} & 87.5 & 72.9 & 89.6 & 85.4 & 85.4 & 48 \\
2 subgoals & 28.1 & 56.2 & 62.5 & 68.8 & 64.1 & 67.2 & 15.1 & 74.0 & \textbf{87.0} & 74.5 & 192 \\
3 subgoals & 6.0 & 29.6 & 36.1 & 40.7 & 49.5 & 47.2 & 9.7 & 57.4 & \textbf{75.0} & 50.5 & 216 \\
4 subgoals & 1.8 & 4.8 & 11.9 & 19.0 & 18.5 & 26.8 & 1.2 & 36.3 & \textbf{63.7} & 22.6 & 168 \\
5 subgoals & 0.0 & 0.6 & 7.7 & 10.7 & 6.0 & 11.3 & 0.0 & 25.6 & \textbf{48.8} & 11.9 & 168 \\
6 subgoals & 0.0 & 0.0 & 2.1 & 6.2 & 5.2 & 13.5 & 0.0 & 31.2 & \textbf{58.3} & 17.7 & 96 \\
7 subgoals & 0.0 & 0.0 & 0.8 & 3.3 & 5.8 & 8.3 & 0.0 & 16.7 & \textbf{47.5} & 3.3 & 120 \\
8 subgoals & 0.0 & 0.0 & 0.0 & 1.0 & 2.1 & 5.2 & 0.0 & 4.2 & \textbf{39.6} & 6.2 & 96 \\
9 subgoals & 0.0 & 0.0 & 1.0 & 0.0 & 1.0 & 5.2 & 0.0 & 10.4 & \textbf{39.6} & 3.1 & 96 \\
10 subgoals & 0.0 & 0.0 & 0.0 & 2.1 & 0.0 & 2.1 & 0.0 & 4.2 & \textbf{35.4} & 1.0 & 96 \\
\emph{All tasks} & 8.7 & 17.4 & 21.5 & 25.2 & 25.5 & 28.7 & 6.7 & 37.1 & \textbf{60.3} & 29.5 & 1,296 \\
\midrule
\multicolumn{12}{l}{\textbf{\robotarena{}-Long}} \\
1 subgoal & 71.9 & 62.5 & 56.2 & 62.5 & 50.0 & 40.6 & 21.9 & 59.4 & \textbf{90.6} & 68.8 & 32 \\
2 subgoals & 13.4 & 25.9 & 31.2 & 42.0 & 40.2 & 24.1 & 19.6 & 63.4 & \textbf{82.1} & 40.2 & 112 \\
3 subgoals & 3.8 & 8.8 & 8.8 & 22.5 & 16.2 & 16.2 & 16.2 & 51.2 & \textbf{76.2} & 23.8 & 80 \\
4 subgoals & 0.0 & 2.5 & 2.5 & 15.0 & 12.5 & 12.5 & 0.0 & 25.0 & \textbf{62.5} & 20.0 & 40 \\
5 subgoals & 0.0 & 0.0 & 0.0 & 0.0 & 4.2 & 0.0 & 0.0 & 20.8 & \textbf{54.2} & 8.3 & 24 \\
6 subgoals & 0.0 & 0.0 & 0.0 & 0.0 & 0.0 & 7.5 & 0.0 & 7.5 & \textbf{27.5} & 0.0 & 40 \\
7 subgoals & 0.0 & 0.0 & 0.0 & 0.0 & 0.0 & 0.0 & 0.0 & 6.2 & \textbf{56.2} & 0.0 & 16 \\
8 subgoals & 0.0 & 0.0 & 0.0 & 0.0 & 0.0 & 0.0 & 0.0 & 0.0 & \textbf{31.2} & 0.0 & 16 \\
9 subgoals & 0.0 & 0.0 & 0.0 & 0.0 & 0.0 & 0.0 & 0.0 & 0.0 & \textbf{12.5} & 0.0 & 8 \\
10 subgoals & 0.0 & 0.0 & 0.0 & 0.0 & 0.0 & 0.0 & 0.0 & 0.0 & \textbf{62.5} & 0.0 & 8 \\
\emph{All tasks} & 10.9 & 15.2 & 16.2 & 24.2 & 21.3 & 16.2 & 11.2 & 39.9 & \textbf{66.8} & 25.5 & 376 \\
\midrule
\multicolumn{12}{l}{\textbf{RoboLab-Long}} \\
1 subgoal & \textbf{100.0} & \textbf{100.0} & \textbf{100.0} & 87.5 & 93.8 & 81.2 & 87.5 & 93.8 & \textbf{100.0} & 87.5 & 16 \\
2 subgoals & 46.9 & 68.8 & 73.4 & 56.2 & 68.8 & 48.4 & 39.1 & 78.1 & \textbf{92.2} & 78.1 & 64 \\
3 subgoals & 22.5 & 20.0 & 30.0 & 37.5 & 32.5 & 15.0 & 30.0 & 50.0 & \textbf{85.0} & 40.0 & 40 \\
4 subgoals & 14.6 & 22.9 & 29.2 & 27.1 & 35.4 & 33.3 & 16.7 & \textbf{77.1} & 72.9 & 43.8 & 48 \\
5 subgoals & 0.0 & 0.0 & 12.5 & 0.0 & 0.0 & 0.0 & 0.0 & 45.8 & \textbf{70.8} & 8.3 & 24 \\
6 subgoals & 10.4 & 14.6 & 14.6 & 16.7 & 18.8 & 6.2 & 2.1 & 39.6 & \textbf{47.9} & 18.8 & 48 \\
7 subgoals & 0.0 & 0.0 & 0.0 & 0.0 & 0.0 & 0.0 & 4.2 & \textbf{45.8} & 37.5 & 0.0 & 24 \\
8 subgoals & 0.0 & 6.2 & 0.0 & 3.1 & 6.2 & 6.2 & 0.0 & 28.1 & \textbf{37.5} & 15.6 & 32 \\
9 subgoals & 0.0 & 0.0 & 0.0 & 4.2 & 4.2 & 0.0 & 0.0 & \textbf{8.3} & \textbf{8.3} & 4.2 & 24 \\
10 subgoals & 6.2 & 6.2 & 0.0 & 6.2 & 3.1 & 6.2 & 0.0 & 21.9 & \textbf{43.8} & 9.4 & 32 \\
\emph{All tasks} & 19.6 & 25.6 & 28.1 & 25.6 & 29.0 & 20.7 & 17.3 & 51.4 & \textbf{62.8} & 34.4 & 352 \\
\bottomrule
\end{tabular}
}
\end{table}

\begin{table}[!ht]
\centering
\setlength{\tabcolsep}{3pt}
\caption{\textbf{Task success rate (\%) per suite, system and task category.} Table~\ref{tab:comp-overall} combines the three suites; here each suite is shown on its own.}
\label{tab:suite-category-success}
\scriptsize
\resizebox{\linewidth}{!}{%
\begin{tabular}{lcccc|cccc|c|c|r}
\toprule
Tasks & \shortstack{Low-level\\policy} & \shortstack{Periodic\\trigger} & \shortstack{Proprioceptive\\trigger} & \shortstack{Proprioceptive\\verifier} & VoLo & \capx{} & \tiptop{} & \harness{} & \gptpolicy{} & \shortstack{Privileged\\verifier} & Episodes \\
\midrule
\multicolumn{12}{l}{\textbf{SimplerEnv-Long}} \\
Single destination & 9.6 & 18.8 & 22.7 & 26.9 & 27.1 & 23.8 & 11.7 & 41.2 & \textbf{54.4} & 30.0 & 480 \\
Transfer & 9.2 & 8.3 & 7.5 & 9.2 & 22.5 & 44.2 & 1.7 & 25.8 & \textbf{67.5} & 21.7 & 120 \\
Split destination & 12.5 & 15.3 & 21.5 & 23.6 & 25.7 & 21.5 & 8.3 & 41.7 & \textbf{61.1} & 29.2 & 144 \\
Relay & 3.2 & 13.0 & 19.0 & 25.5 & 18.1 & 39.8 & 0.0 & 23.1 & \textbf{74.5} & 28.2 & 216 \\
Revisit & 0.0 & 0.0 & 0.0 & 0.0 & 1.0 & 9.4 & 0.0 & 8.3 & \textbf{39.6} & 1.0 & 96 \\
Exclusion & 0.0 & 35.4 & 46.9 & 46.9 & 44.8 & 35.4 & 6.2 & 55.2 & \textbf{70.8} & 47.9 & 96 \\
Ordered & 39.6 & 64.6 & 62.5 & 72.9 & 81.2 & 62.5 & 22.9 & \textbf{87.5} & 81.2 & 85.4 & 48 \\
Duplicates & 12.5 & 10.4 & 14.6 & 17.7 & 15.6 & 15.6 & 0.0 & 40.6 & \textbf{47.9} & 21.9 & 96 \\
\emph{All tasks} & 8.7 & 17.4 & 21.5 & 25.2 & 25.5 & 28.7 & 6.7 & 37.1 & \textbf{60.3} & 29.5 & 1,296 \\
\midrule
\multicolumn{12}{l}{\textbf{\robotarena{}-Long}} \\
Single destination & 15.8 & 23.7 & 19.7 & 34.2 & 32.2 & 13.8 & 9.2 & 46.1 & \textbf{79.6} & 36.2 & 152 \\
Transfer & 0.0 & 0.0 & 0.0 & 0.0 & 0.0 & 50.0 & 25.0 & \textbf{100.0} & 62.5 & 0.0 & 8 \\
Split destination & 0.0 & 12.5 & 0.0 & 18.8 & 25.0 & 25.0 & 31.2 & \textbf{100.0} & 56.2 & 18.8 & 16 \\
Relay & 0.0 & 3.6 & 1.8 & 5.4 & 5.4 & 17.9 & 1.8 & 19.6 & \textbf{55.4} & 10.7 & 56 \\
Revisit & 0.0 & 0.0 & 0.0 & 0.0 & 0.0 & 12.5 & 0.0 & 4.2 & \textbf{58.3} & 2.1 & 48 \\
Exclusion & 1.8 & 16.1 & 26.8 & 30.4 & 8.9 & 8.9 & 10.7 & 26.8 & \textbf{32.1} & 26.8 & 56 \\
Ordered & 20.8 & 4.2 & 20.8 & 33.3 & 37.5 & 29.2 & 16.7 & 66.7 & \textbf{95.8} & 33.3 & 24 \\
Duplicates & 68.8 & 43.8 & 62.5 & 50.0 & 62.5 & 25.0 & 62.5 & 75.0 & \textbf{100.0} & 50.0 & 16 \\
\emph{All tasks} & 10.9 & 15.2 & 16.2 & 24.2 & 21.3 & 16.2 & 11.2 & 39.9 & \textbf{66.8} & 25.5 & 376 \\
\midrule
\multicolumn{12}{l}{\textbf{RoboLab-Long}} \\
Single destination & 75.0 & 80.0 & 72.5 & 67.5 & 72.5 & 57.5 & 42.5 & 70.0 & \textbf{82.5} & 70.0 & 40 \\
Transfer & \textbf{81.2} & 68.8 & 75.0 & 68.8 & 75.0 & 37.5 & 68.8 & \textbf{81.2} & \textbf{81.2} & 75.0 & 16 \\
Split destination & 2.2 & 7.4 & 15.4 & 10.3 & 14.7 & 9.6 & 14.0 & 54.4 & \textbf{60.3} & 19.9 & 136 \\
Relay & 0.0 & 9.4 & 15.6 & 21.9 & 21.9 & 6.2 & 0.0 & 40.6 & \textbf{81.2} & 50.0 & 32 \\
Revisit & 0.0 & 8.3 & 8.3 & 20.8 & 20.8 & 0.0 & 0.0 & 0.0 & \textbf{29.2} & 25.0 & 24 \\
Exclusion & 0.0 & 56.2 & 43.8 & 25.0 & 43.8 & 81.2 & 37.5 & 87.5 & \textbf{93.8} & 62.5 & 16 \\
Ordered & 11.1 & 12.5 & 9.7 & 11.1 & 9.7 & 2.8 & 4.2 & 31.9 & \textbf{44.4} & 11.1 & 72 \\
Duplicates & 93.8 & 87.5 & \textbf{100.0} & 87.5 & 93.8 & 87.5 & 31.2 & \textbf{100.0} & 81.2 & 87.5 & 16 \\
\emph{All tasks} & 19.6 & 25.6 & 28.1 & 25.6 & 29.0 & 20.7 & 17.3 & 51.4 & \textbf{62.8} & 34.4 & 352 \\
\bottomrule
\end{tabular}
}
\end{table}

\begin{table}[!ht]
\centering
\setlength{\tabcolsep}{3pt}
\caption{\textbf{Subgoal progress (\%) per suite, system and number of subgoals} (Figure~\ref{fig:ladder}, bottom).}
\label{tab:suite-horizon-progress}
\scriptsize
\resizebox{\linewidth}{!}{%
\begin{tabular}{lcccc|cccc|c|c|r}
\toprule
Tasks & \shortstack{Low-level\\policy} & \shortstack{Periodic\\trigger} & \shortstack{Proprioceptive\\trigger} & \shortstack{Proprioceptive\\verifier} & VoLo & \capx{} & \tiptop{} & \harness{} & \gptpolicy{} & \shortstack{Privileged\\verifier} & Episodes \\
\midrule
\multicolumn{12}{l}{\textbf{SimplerEnv-Long}} \\
1 subgoal & 89.6 & 91.7 & 91.7 & 89.6 & \textbf{93.8} & 87.5 & 72.9 & 89.6 & 85.4 & 85.4 & 48 \\
2 subgoals & 49.2 & 73.7 & 77.9 & 81.8 & 76.0 & 78.9 & 29.9 & 83.1 & \textbf{92.7} & 83.6 & 192 \\
3 subgoals & 37.2 & 62.0 & 65.9 & 66.0 & 69.0 & 69.8 & 29.2 & 76.4 & \textbf{89.8} & 70.2 & 216 \\
4 subgoals & 29.9 & 43.3 & 48.2 & 54.2 & 54.9 & 56.0 & 11.3 & 67.3 & \textbf{87.5} & 54.5 & 168 \\
5 subgoals & 30.7 & 38.2 & 46.9 & 51.2 & 46.3 & 43.2 & 1.8 & 61.5 & \textbf{84.8} & 49.0 & 168 \\
6 subgoals & 25.3 & 41.0 & 48.1 & 50.2 & 47.2 & 46.4 & 0.0 & 66.8 & \textbf{88.2} & 59.5 & 96 \\
7 subgoals & 25.2 & 31.4 & 42.3 & 47.1 & 46.7 & 41.1 & 1.9 & 55.1 & \textbf{83.7} & 50.4 & 120 \\
8 subgoals & 17.2 & 31.1 & 37.6 & 46.2 & 38.9 & 32.8 & 2.2 & 46.1 & \textbf{78.3} & 44.1 & 96 \\
9 subgoals & 14.8 & 34.5 & 35.0 & 42.2 & 39.1 & 39.5 & 0.9 & 48.6 & \textbf{78.6} & 39.2 & 96 \\
10 subgoals & 12.5 & 26.8 & 32.6 & 39.7 & 35.9 & 29.9 & 0.0 & 41.2 & \textbf{76.0} & 43.8 & 96 \\
\emph{All tasks} & 25.3 & 39.9 & 45.8 & 51.0 & 48.3 & 45.9 & 7.1 & 58.2 & \textbf{83.3} & 52.4 & 1,296 \\
\midrule
\multicolumn{12}{l}{\textbf{\robotarena{}-Long}} \\
1 subgoal & 71.9 & 62.5 & 56.2 & 62.5 & 50.0 & 40.6 & 21.9 & 59.4 & \textbf{90.6} & 68.8 & 32 \\
2 subgoals & 28.1 & 51.3 & 51.8 & 57.1 & 56.2 & 35.3 & 32.6 & 75.4 & \textbf{88.8} & 53.1 & 112 \\
3 subgoals & 22.1 & 40.4 & 44.2 & 42.5 & 39.2 & 42.9 & 30.8 & 74.6 & \textbf{91.7} & 47.9 & 80 \\
4 subgoals & 6.9 & 23.1 & 25.0 & 38.1 & 37.5 & 28.1 & 5.0 & 49.4 & \textbf{86.9} & 40.0 & 40 \\
5 subgoals & 8.3 & 29.2 & 30.0 & 29.2 & 28.3 & 15.0 & 3.3 & 51.7 & \textbf{84.2} & 30.0 & 24 \\
6 subgoals & 7.1 & 21.2 & 22.1 & 27.5 & 24.2 & 20.4 & 1.7 & 43.8 & \textbf{73.8} & 18.8 & 40 \\
7 subgoals & 3.6 & 11.6 & 11.6 & 19.6 & 13.4 & 33.9 & 3.6 & 32.1 & \textbf{80.4} & 17.9 & 16 \\
8 subgoals & 2.3 & 9.4 & 9.4 & 12.5 & 14.1 & 27.3 & 0.0 & 20.3 & \textbf{75.0} & 15.6 & 16 \\
9 subgoals & 4.2 & 16.7 & 8.3 & 19.4 & 18.1 & 8.3 & 5.6 & 19.4 & \textbf{73.6} & 16.7 & 8 \\
10 subgoals & 3.8 & 8.8 & 8.8 & 13.8 & 17.5 & 36.2 & 0.0 & 22.5 & \textbf{87.5} & 8.8 & 8 \\
\emph{All tasks} & 13.5 & 28.3 & 28.9 & 33.7 & 31.8 & 29.5 & 12.6 & 50.2 & \textbf{83.4} & 32.7 & 376 \\
\midrule
\multicolumn{12}{l}{\textbf{RoboLab-Long}} \\
1 subgoal & \textbf{100.0} & \textbf{100.0} & \textbf{100.0} & 87.5 & 93.8 & 81.2 & 87.5 & 93.8 & \textbf{100.0} & 87.5 & 16 \\
2 subgoals & 68.8 & 77.3 & 84.4 & 73.4 & 80.5 & 68.8 & 61.7 & 90.6 & \textbf{96.1} & 87.5 & 64 \\
3 subgoals & 63.3 & 54.2 & 70.0 & 64.2 & 70.0 & 44.2 & 58.3 & 76.7 & \textbf{91.7} & 72.5 & 40 \\
4 subgoals & 66.7 & 65.6 & 69.3 & 65.1 & 73.4 & 61.5 & 46.4 & \textbf{92.7} & 86.5 & 77.6 & 48 \\
5 subgoals & 63.3 & 49.2 & 60.8 & 45.0 & 50.8 & 47.5 & 52.5 & 86.7 & \textbf{90.8} & 60.8 & 24 \\
6 subgoals & 49.7 & 50.0 & 57.3 & 59.4 & 67.7 & 37.5 & 18.8 & 76.4 & \textbf{80.2} & 55.2 & 48 \\
7 subgoals & 44.6 & 52.4 & 50.6 & 51.2 & 52.4 & 24.4 & 17.3 & \textbf{87.5} & 78.6 & 49.4 & 24 \\
8 subgoals & 48.0 & 56.2 & 51.6 & 45.3 & 55.9 & 32.8 & 0.0 & \textbf{83.6} & 73.4 & 58.6 & 32 \\
9 subgoals & 53.2 & 56.0 & 56.9 & 63.4 & 56.0 & 37.5 & 5.1 & \textbf{67.1} & 47.2 & 53.2 & 24 \\
10 subgoals & 60.3 & 56.2 & 48.8 & 46.9 & 56.2 & 40.6 & 0.0 & 80.6 & \textbf{80.9} & 67.5 & 32 \\
\emph{All tasks} & 56.6 & 57.1 & 58.9 & 56.1 & 62.0 & 42.4 & 22.4 & \textbf{81.6} & 78.7 & 63.5 & 352 \\
\bottomrule
\end{tabular}
}
\end{table}

\begin{table}[!ht]
\centering
\renewcommand{\arraystretch}{0.90}
\setlength{\tabcolsep}{4pt}
\caption{\textbf{Offline evaluation of VLMs per suite.} Bold marks the best model within each suite.}
\label{tab:vlmbench-suites}
\scriptsize
\begin{tabular}{lccccc}
\toprule
 & \multicolumn{1}{c}{Planning} & \multicolumn{2}{c}{Subgoal completion} & \multicolumn{1}{c}{State tracking} & \multicolumn{1}{c}{Next subgoal} \\
\cmidrule(lr){2-2}\cmidrule(lr){3-4}\cmidrule(lr){5-5}\cmidrule(lr){6-6}
Model & exact match (\%) & TP rate (\%) & TN rate (\%) & all correct (\%) & valid (\%) \\
\midrule
\multicolumn{6}{l}{\textbf{SimplerEnv-Long scenes}} \\
Qwen3-VL-2B & 61.1 & 52.1 & 40.4 & 10.6 & 39.9 \\
Qwen3-VL-8B & 92.6 & 58.5 & 86.5 & 58.6 & 65.2 \\
Qwen3.5-9B & 96.3 & 45.7 & \textbf{97.1} & 13.6 & 60.6 \\
Qwen3.8-27B & 92.6 & \textbf{63.8} & 92.3 & \textbf{71.2} & 79.3 \\
Gemini 3.1 Pro & 96.3 & 47.9 & \textbf{97.1} & 66.2 & 69.2 \\
Gemini 3.7 Flash & 96.3 & 52.1 & 96.2 & 67.2 & 77.8 \\
Gemini Robotics-ER 2 & 96.3 & 43.6 & \textbf{97.1} & 65.2 & 71.7 \\
GPT-5.6 Sol & 94.4 & 55.3 & 89.4 & 66.7 & 83.3 \\
GPT-6 Astra & \textbf{98.1} & 55.3 & 90.4 & 64.6 & \textbf{84.8} \\
Claude Opus 5 & 96.3 & 60.6 & 90.4 & 69.7 & 83.8 \\
\midrule
\multicolumn{6}{l}{\textbf{\robotarena{}-Long scenes}} \\
Qwen3-VL-2B & 80.9 & 71.2 & 40.9 & 8.5 & 60.1 \\
Qwen3-VL-8B & 85.1 & 80.8 & 73.9 & 44.1 & 70.7 \\
Qwen3.5-9B & 80.9 & 42.5 & 94.8 & 12.2 & 70.2 \\
Qwen3.8-27B & 83.0 & \textbf{82.2} & 77.4 & 50.0 & 75.5 \\
Gemini 3.1 Pro & 74.5 & 42.5 & 92.2 & 64.9 & 71.3 \\
Gemini 3.7 Flash & 85.1 & 47.9 & 95.7 & \textbf{70.2} & 79.3 \\
Gemini Robotics-ER 2 & 76.6 & 42.5 & \textbf{97.4} & 68.6 & 72.9 \\
GPT-5.6 Sol & \textbf{87.2} & 76.7 & 78.3 & 55.3 & 76.6 \\
GPT-6 Astra & 80.9 & 71.2 & 82.6 & 54.3 & \textbf{80.3} \\
Claude Opus 5 & 76.6 & 80.8 & 84.3 & 58.0 & 71.8 \\
\midrule
\multicolumn{6}{l}{\textbf{RoboLab-Long scenes}} \\
Qwen3-VL-2B & 86.4 & 86.1 & 20.2 & 8.5 & 79.3 \\
Qwen3-VL-8B & 95.5 & 70.8 & 79.8 & 39.8 & 82.2 \\
Qwen3.5-9B & 81.8 & 44.4 & \textbf{99.0} & 9.1 & 58.6 \\
Qwen3.8-27B & \textbf{97.7} & 86.1 & 85.6 & 52.3 & 83.3 \\
Gemini 3.1 Pro & \textbf{97.7} & 76.4 & \textbf{99.0} & \textbf{75.6} & 86.2 \\
Gemini 3.7 Flash & \textbf{97.7} & 80.6 & 98.1 & \textbf{75.6} & 87.9 \\
Gemini Robotics-ER 2 & \textbf{97.7} & 79.2 & \textbf{99.0} & \textbf{75.6} & 86.2 \\
GPT-5.6 Sol & \textbf{97.7} & \textbf{91.7} & 92.3 & 46.0 & 86.2 \\
GPT-6 Astra & 95.5 & \textbf{91.7} & 86.5 & 39.2 & \textbf{91.4} \\
Claude Opus 5 & \textbf{97.7} & 66.7 & 89.4 & 54.5 & 83.9 \\
\bottomrule
\end{tabular}
\end{table}

\section{Efficiency and Token Usage}
\label{app:efficiency}
\begin{table}[!ht]
\centering
\setlength{\tabcolsep}{3pt}
\caption{\textbf{Efficiency of every system, per suite.} Subgoals completed are shares of all subgoals, measured when half of the simulated time budget has elapsed and at its end (the latter is the subgoal progress of Figure~\ref{fig:ladder}).}
\label{tab:efficiency}
\small
\resizebox{\linewidth}{!}{%
\begin{tabular}{lcccc|cccc|c|c}
\toprule
 & \shortstack{Low-level\\policy} & \shortstack{Periodic\\trigger} & \shortstack{Proprioceptive\\trigger} & \shortstack{Proprioceptive\\verifier} & VoLo & \capx{} & \tiptop{} & \harness{} & \gptpolicy{} & \shortstack{Privileged\\verifier} \\
\midrule
\multicolumn{11}{l}{\emph{VLM calls per episode}} \\
\quad SimplerEnv-Long & 0.0 & 0.0 & 0.0 & 12.7 & 24.8 & 3.8 & 1.0 & 18.0 & 50.1 & 0.0 \\
\quad \robotarena{}-Long & 0.0 & 0.0 & 0.0 & 9.3 & 26.4 & 5.6 & 1.0 & 20.2 & 41.7 & 0.0 \\
\quad RoboLab-Long & 0.0 & 0.0 & 0.0 & 9.3 & 19.2 & 6.3 & 1.0 & 31.8 & 70.8 & 0.0 \\
\addlinespace
\multicolumn{11}{l}{\emph{VLM calls per completed subgoal}} \\
\quad SimplerEnv-Long & 0.0 & 0.0 & 0.0 & 4.9 & 10.1 & 1.6 & 2.8 & 6.1 & 11.8 & 0.0 \\
\quad \robotarena{}-Long & 0.0 & 0.0 & 0.0 & 7.3 & 22.2 & 5.1 & 2.1 & 10.7 & 13.3 & 0.0 \\
\quad RoboLab-Long & 0.0 & 0.0 & 0.0 & 3.2 & 6.0 & 2.9 & 0.9 & 7.5 & 17.4 & 0.0 \\
\addlinespace
\multicolumn{11}{l}{\emph{Input tokens per episode (thousands)}} \\
\quad SimplerEnv-Long & 0.0 & 0.0 & 0.0 & 17.2 & 70.5 & 7.5 & 2.0 & 27.3 & 357.8 & 0.0 \\
\quad \robotarena{}-Long & 0.0 & 0.0 & 0.0 & 12.6 & 75.0 & 12.9 & 2.0 & 31.2 & 293.5 & 0.0 \\
\quad RoboLab-Long & 0.0 & 0.0 & 0.0 & 23.6 & 91.6 & 15.3 & 2.1 & 48.5 & 508.8 & 0.0 \\
\addlinespace
\multicolumn{11}{l}{\emph{Generated tokens per episode (thousands)}} \\
\quad SimplerEnv-Long & 0.0 & 0.0 & 0.0 & 2.7 & 7.4 & 3.1 & 0.5 & 10.4 & 5.7 & 0.0 \\
\quad \robotarena{}-Long & 0.0 & 0.0 & 0.0 & 1.8 & 6.6 & 4.6 & 0.4 & 10.1 & 4.9 & 0.0 \\
\quad RoboLab-Long & 0.0 & 0.0 & 0.0 & 2.5 & 7.2 & 5.1 & 0.4 & 16.4 & 8.3 & 0.0 \\
\addlinespace
\multicolumn{11}{l}{\emph{API cost per episode (USD)}} \\
\quad SimplerEnv-Long & 0.000 & 0.000 & 0.000 & 0.046 & 0.161 & 0.034 & 0.007 & 0.119 & 3.861 & 0.000 \\
\quad \robotarena{}-Long & 0.000 & 0.000 & 0.000 & 0.032 & 0.162 & 0.054 & 0.006 & 0.123 & 3.177 & 0.000 \\
\quad RoboLab-Long & 0.000 & 0.000 & 0.000 & 0.054 & 0.191 & 0.061 & 0.006 & 0.196 & 5.501 & 0.000 \\
\addlinespace
\multicolumn{11}{l}{\emph{API cost per successful episode (USD)}} \\
\quad SimplerEnv-Long & 0.00 & 0.00 & 0.00 & 0.18 & 0.63 & 0.12 & 0.10 & 0.32 & 6.40 & 0.00 \\
\quad \robotarena{}-Long & 0.00 & 0.00 & 0.00 & 0.13 & 0.76 & 0.33 & 0.05 & 0.31 & 4.76 & 0.00 \\
\quad RoboLab-Long & 0.00 & 0.00 & 0.00 & 0.21 & 0.66 & 0.30 & 0.04 & 0.38 & 8.76 & 0.00 \\
\addlinespace
\multicolumn{11}{l}{\emph{Subgoals completed by half of the time budget (\%)}} \\
\quad SimplerEnv-Long & 21.3 & 37.8 & 42.3 & 44.1 & 40.3 & 38.4 & 6.0 & 42.9 & 60.9 & 47.8 \\
\quad \robotarena{}-Long & 10.5 & 25.4 & 24.4 & 24.6 & 26.7 & 19.4 & 12.6 & 37.9 & 60.2 & 24.4 \\
\quad RoboLab-Long & 53.1 & 52.6 & 50.2 & 48.4 & 53.6 & 41.6 & 22.3 & 78.4 & 78.1 & 55.9 \\
\addlinespace
\multicolumn{11}{l}{\emph{Subgoals completed by the end of the budget (\%)}} \\
\quad SimplerEnv-Long & 25.3 & 39.9 & 45.8 & 51.0 & 48.3 & 45.9 & 7.1 & 58.2 & 83.3 & 52.4 \\
\quad \robotarena{}-Long & 13.5 & 28.3 & 28.9 & 33.7 & 31.8 & 29.5 & 12.6 & 50.2 & 83.4 & 32.7 \\
\quad RoboLab-Long & 56.6 & 57.1 & 58.9 & 56.1 & 62.0 & 42.4 & 22.4 & 81.6 & 78.7 & 63.3 \\
\bottomrule
\end{tabular}
}
\end{table}

Section~\ref{sec:exp:horizon} compares systems by two costs that do not depend on the hardware a run used.

\paragraph{Robot time.} Every episode has a simulated time budget (Section~\ref{sec:benchmarks:metrics}). The simulation advances only when a system returns an action, so time spent computing, for example waiting for a VLM, does not use the budget. We record which subgoals are satisfied at every step and report the share of all subgoals completed when a given share of the budget has elapsed. A successful episode ends at success, and its subgoals count as completed for the rest of the budget. At the end of the budget this share is exactly the subgoal progress of Figure~\ref{fig:ladder}, which we check for every evaluated unit that recorded its episode log (all but one \robolab{}-Long task of the privileged verifier, whose 8 episodes are therefore absent from Tables~\ref{tab:efficiency}, \ref{tab:subgoal-causes} and~\ref{tab:orchestrator-decisions}). Successful episodes of the learned-policy systems end, on average, after \successBudgetLo--\successBudgetHi\% of their budget.

\paragraph{VLM calls.} We count the successful requests each episode sends to the VLM endpoint. Each request is recorded twice, in the evaluation log and in the usage log of the VLM client, and neither record is complete on its own. We match the two records by time and count a request found in either of them once. Requests refused by the endpoint's rate limit and then retried are not counted.

\paragraph{Why not wall-clock time.} Wall-clock time is logged, but it does not compare systems fairly. The same \pifive{} policy with no VLM calls takes \wallMsRtxLo--\wallMsRtxHi\,ms per step on an RTX PRO 6000 but \wallMsAhLo--\wallMsAhHi\,ms on an A100, several runs share one GPU, and VLM requests wait in queues shared with other jobs. No two systems ran on identical hardware.

\paragraph{Tokens and API cost.} Calls of different systems send different amounts of text and images and generate different amounts of text, so we also report the tokens of the counted calls, from each provider's own usage report, and their API cost at the providers' standard list prices on 25 September 2026: \$1.50 per million input tokens and \$7.50 per million generated tokens for Gemini 3.7 Flash, and \$10 and \$50 for GPT-6 Astra (Table~\ref{tab:efficiency}). Generated tokens include reasoning. The cost covers only the hosted VLM, not the GPU time of the low-level policy, of \tiptop{}'s planner or of the simulators. Over all suites together, an episode costs \$\effUsdTiptop{} with \tiptop{}, \$\effUsdCapx{} with \capx{}, \$\effUsdVerifier{} with the \emph{Proprioceptive verifier}, \$\effUsdHarness{} with \harness{}, \$\effUsdVolo{} with VoLo and \$\effUsdGpt{} with \gptpolicy{}. \gptpolicy{} succeeds most often, but costs \$\effUsdSuccGpt{} per successful episode (\$\effUsdGptTotal{} for one full evaluation) against \$\effUsdSuccHarness{} for \harness{}, and most of its cost comes from input tokens. The usage reports do not say how much input the provider served from its cache, so \gptpolicy{}'s cost is an upper bound.

\begin{table}[!ht]
\centering
\setlength{\tabcolsep}{4pt}
\caption{\textbf{Tokens per query of the offline benchmark.} Mean input and generated tokens (reasoning included) per scored query, from each API's own usage counts. A query is one scene for plan decomposition and one frame for the other capabilities; state tracking asks about each subgoal of the frame separately. Input counts follow each provider's own tokenizer and image accounting, so they compare within a provider only. The two local Qwen3-VL models report no usage.}
\label{tab:vlm-tokens}
\footnotesize
\begin{tabular}{lcccccccc}
\toprule
 & \multicolumn{2}{c}{Plan decomposition} & \multicolumn{2}{c}{Subgoal completion} & \multicolumn{2}{c}{State tracking} & \multicolumn{2}{c}{Next subgoal} \\
\cmidrule(lr){2-3}\cmidrule(lr){4-5}\cmidrule(lr){6-7}\cmidrule(lr){8-9}
Model & input & generated & input & generated & input & generated & input & generated \\
\midrule
Qwen3-VL-2B & -- & -- & -- & -- & -- & -- & -- & -- \\
Qwen3-VL-8B & -- & -- & -- & -- & -- & -- & -- & -- \\
Qwen3.5-9B & 655 & 2{,}951 & 875 & 2{,}404 & 3{,}445 & 10{,}325 & 653 & 3{,}339 \\
Qwen3.8-27B & 690 & 943 & 617 & 967 & 2{,}321 & 3{,}790 & 690 & 1{,}189 \\
Gemini 3.1 Pro & 1{,}474 & 663 & 1{,}727 & 231 & 6{,}965 & 970 & 1{,}471 & 529 \\
Gemini 3.7 Flash & 1{,}475 & 431 & 1{,}727 & 195 & 6{,}960 & 743 & 1{,}471 & 290 \\
Gemini Robotics-ER 2 & 1{,}476 & 518 & 1{,}729 & 219 & 6{,}987 & 886 & 1{,}473 & 424 \\
GPT-5.6 Sol & 682 & 147 & 610 & 151 & 2{,}270 & 591 & 674 & 138 \\
GPT-6 Astra & 682 & 91 & 610 & 110 & 2{,}270 & 475 & 675 & 53 \\
Claude Opus 5 & 903 & 119 & 784 & 253 & 2{,}930 & 884 & 885 & 133 \\
\bottomrule
\end{tabular}
\end{table}

\paragraph{Tokens of the offline benchmark.} Table~\ref{tab:vlm-tokens} gives the tokens each model uses on the offline benchmark of Section~\ref{sec:exp:vlm}. The counts come from each API's own usage report, matched to the scored queries of every run (at least \tokCoverMin\% of the queries of every model and capability). Each model ran the benchmark once. Next-subgoal queries were sampled at temperature 0.3, and Claude Opus 5 and GPT-6 Astra ran at the provider's default temperature, because their endpoint does not accept one.

\section{Diagnostic Failure Analysis}
\label{sec:app:failures}
Section~\ref{sec:exp:failures} summarizes where subgoals are lost. This appendix defines every cause and reports the causes for each benchmark separately.

\paragraph{Subgoal causes.} We reconstruct each episode from its logged object poses, per-subgoal success predicates, gripper state, and orchestrator events, and assign every subgoal exactly one outcome. A subgoal that counts toward the subgoal progress is \emph{completed}; every other subgoal receives the first of the following twelve causes that applies, grouped into five stages.
\begin{itemize}\itemsep0em
    \item \emph{Undone later}: the subgoal was done at some step but not at the end. The robot \emph{picked it up again} when the gripper took the object again; the object was \emph{knocked off} when a collision moved it while nothing held it, and \emph{slid or rolled off} otherwise.
    \item Otherwise, if its object was ever held, the last hold decides. A hold credited to another of the object's subgoals (relay and revisit tasks) belongs to that subgoal and is skipped. The robot \emph{lost grip before lifting} when the object left the gripper before rising 3\,cm. The object was \emph{dropped on the way} when it was lifted and released away from every destination, and \emph{timed out mid-carry} when time ran out while the robot still held it. An object set down within its own destination's footprint (plus 2\,cm) without being counted is \emph{slightly off target} (\emph{Placing}); one set down on a different destination was \emph{put in the wrong place}.
    \item Otherwise the object was never held. For orchestrators that follow a fixed plan in order, the \emph{plan never reached} the subgoal when it was never issued. The robot \emph{missed the object} (\emph{Grasping}) when the gripper closed next to it, nearer to it than to any other object, and held nothing. For \tiptop{} only, \emph{no plan was found} when its planner reported failure. Every remaining subgoal had its \emph{object never touched}. A later leg of an object whose earlier leg was never done counts as never touched, since every hold of that object served the earlier leg.
\end{itemize}
The completed count reproduces the reported subgoal progress exactly for every evaluated unit with an episode log, so the causes partition exactly the subgoals Figure~\ref{fig:ladder} counts as lost.

\begin{figure}[!ht]
\centering
\includegraphics[width=0.93\textwidth]{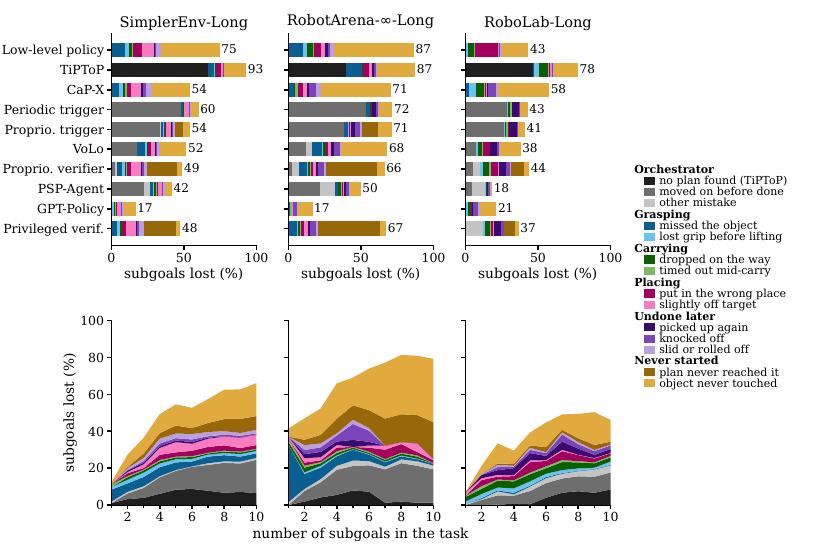}
\caption{\textbf{Why subgoals fail, per benchmark.} Share of each suite's subgoals not completed, split by cause. Top: one bar per system, over all tasks of the suite. Bottom: all ten systems together, by number of subgoals in the task; on \robotarena{}-Long each length from \faRaSparseFrom{} up rests on \faRaSparseTasks{} tasks. Systems keep the same order in every panel.}
\label{fig:failure-suites}
\end{figure}

\paragraph{How the figures are computed.} Every share is a count of subgoals divided by a count of subgoals. Within one suite, a system's bar (Figure~\ref{fig:failure-suites}, top) divides the subgoals it lost to each cause by all subgoals of all its episodes, over every task of the suite. A point at $N$ subgoals (Figure~\ref{fig:failure-suites}, bottom) combines the episodes of all ten systems on the suite's $N$-subgoal tasks. Figure~\ref{fig:failure-horizon} shows the unweighted mean of the three per-suite values, because SimplerEnv-Long holds more than three times as many subgoals as either other suite and would otherwise dominate. Table~\ref{tab:subgoal-causes} lists the per-suite values. Both figures and Table~\ref{tab:subgoal-causes} draw every lost subgoal once: as an orchestrator decision when one made the loss final (below), and otherwise by its cause. What grows most with horizon on every suite is orchestrator decisions and subgoals never started (Figure~\ref{fig:failure-suites}, bottom).

\begin{table}[!ht]
\centering
\renewcommand{\arraystretch}{0.72}
\setlength{\tabcolsep}{1.8pt}
\caption{\textbf{Where each system loses its subgoals}, as a share of all its subgoals (per suite, over all its tasks), split exactly as Figure~\ref{fig:failure-suites} draws it. \emph{Lost}: not completed at the end of the episode (one minus the subgoal progress). \emph{Orch.}: lost to an orchestrator mistake (Table~\ref{tab:orchestrator-decisions}); every other column counts only the remaining losses. The other columns abbreviate the causes defined above.}
\label{tab:subgoal-causes}
\scriptsize
\begin{tabular*}{\linewidth}{@{\extracolsep{\fill}}lccccccccccccc}
\toprule
 & \multicolumn{2}{c}{} & \multicolumn{2}{c}{Grasping} & \multicolumn{2}{c}{Carrying} & \multicolumn{2}{c}{Placing} & \multicolumn{3}{c}{Undone later} & \multicolumn{2}{c}{Never started} \\
\cmidrule(lr){4-5}\cmidrule(lr){6-7}\cmidrule(lr){8-9}\cmidrule(lr){10-12}\cmidrule(lr){13-14}
System & Lost & Orch. & Missed & \begin{tabular}[b]{@{}c@{}}Lost\\grip\end{tabular} & Dropped & \begin{tabular}[b]{@{}c@{}}Timed\\out\end{tabular} & \begin{tabular}[b]{@{}c@{}}Wrong\\place\end{tabular} & \begin{tabular}[b]{@{}c@{}}Off\\target\end{tabular} & \begin{tabular}[b]{@{}c@{}}Picked\\up\end{tabular} & \begin{tabular}[b]{@{}c@{}}Knocked\\off\end{tabular} & \begin{tabular}[b]{@{}c@{}}Slid\\off\end{tabular} & \begin{tabular}[b]{@{}c@{}}Not\\reached\end{tabular} & Untouched \\
\midrule
\multicolumn{14}{l}{\emph{SimplerEnv-Long}} \\
Low-level policy & 74.7 & -- & 9.1 & 3.3 & 1.7 & 0.7 & 6.1 & 8.3 & 0.6 & 0.9 & 3.3 & 0.0 & 40.6 \\
\tiptop{} & 92.9 & 66.6 & 4.0 & 1.1 & 0.7 & 0.0 & 3.4 & 1.3 & 0.1 & 0.2 & 0.3 & 0.0 & 15.2 \\
\capx{} & 54.1 & -- & 5.0 & 2.7 & 1.9 & 1.1 & 3.2 & 6.4 & 1.5 & 2.0 & 4.5 & 0.0 & 26.0 \\
Periodic trigger & 60.1 & 47.8 & 1.4 & 0.5 & 0.3 & 0.2 & 0.2 & 2.9 & 0.5 & 0.3 & 0.7 & 0.0 & 5.2 \\
Proprioceptive trigger & 54.2 & 34.2 & 1.5 & 0.7 & 0.3 & 0.2 & 0.5 & 3.3 & 0.6 & 0.8 & 1.3 & 5.9 & 4.7 \\
VoLo & 51.7 & 18.0 & 4.9 & 1.6 & 0.6 & 0.4 & 2.0 & 2.8 & 0.3 & 1.0 & 2.1 & 0.0 & 17.9 \\
Proprioceptive verifier & 49.0 & 4.0 & 3.5 & 1.8 & 0.7 & 0.4 & 3.2 & 6.8 & 0.5 & 1.4 & 2.3 & 21.1 & 3.5 \\
\harness{} & 41.8 & 26.5 & 2.3 & 0.7 & 1.2 & 0.9 & 0.8 & 2.2 & 0.6 & 0.4 & 1.7 & 0.0 & 4.5 \\
\gptpolicy{} & 16.7 & -- & 0.7 & 0.9 & 0.7 & 0.8 & 0.9 & 2.3 & 0.4 & 0.5 & 1.2 & 0.0 & 8.2 \\
\midrule
Privileged verifier & 47.6 & 0.0 & 3.8 & 2.3 & 1.0 & 0.3 & 2.5 & 7.1 & 0.4 & 1.8 & 3.4 & 21.7 & 3.3 \\
\addlinespace
\multicolumn{14}{l}{\emph{\robotarena{}-Long}} \\
Low-level policy & 86.5 & -- & 10.2 & 2.9 & 3.7 & 0.7 & 4.9 & 2.6 & 2.1 & 1.8 & 2.4 & 0.0 & 55.3 \\
\tiptop{} & 87.4 & 39.4 & 11.2 & 0.0 & 0.6 & 0.0 & 4.7 & 2.0 & 1.1 & 1.6 & 0.6 & 0.0 & 26.3 \\
\capx{} & 70.5 & -- & 3.9 & 0.0 & 0.5 & 2.3 & 3.6 & 2.2 & 1.6 & 4.7 & 3.5 & 0.0 & 48.2 \\
Periodic trigger & 71.7 & 53.8 & 2.4 & 0.1 & 0.5 & 0.3 & 0.3 & 0.6 & 2.6 & 1.4 & 0.6 & 0.0 & 9.1 \\
Proprioceptive trigger & 71.1 & 38.6 & 2.7 & 0.2 & 0.2 & 0.4 & 0.5 & 0.6 & 2.8 & 3.7 & 1.1 & 11.2 & 9.1 \\
VoLo & 68.2 & 16.4 & 6.7 & 0.8 & 1.8 & 0.3 & 1.3 & 1.8 & 2.4 & 4.0 & 0.9 & 0.0 & 31.8 \\
Proprioceptive verifier & 66.3 & 7.0 & 5.8 & 0.6 & 1.3 & 0.6 & 1.6 & 2.3 & 1.4 & 3.8 & 1.7 & 35.2 & 4.9 \\
\harness{} & 49.8 & 31.8 & 2.6 & 0.0 & 1.6 & 1.3 & 0.5 & 0.9 & 1.2 & 2.1 & 0.8 & 0.0 & 7.1 \\
\gptpolicy{} & 16.6 & -- & 0.8 & 0.1 & 0.4 & 1.0 & 0.4 & 0.2 & 0.5 & 2.5 & 0.8 & 0.0 & 10.0 \\
\midrule
Privileged verifier & 67.3 & 0.0 & 5.8 & 0.6 & 1.7 & 0.4 & 2.4 & 2.7 & 1.6 & 3.6 & 1.5 & 42.6 & 4.5 \\
\addlinespace
\multicolumn{14}{l}{\emph{RoboLab-Long}} \\
Low-level policy & 43.4 & -- & 0.2 & 1.7 & 4.1 & 0.8 & 15.4 & 0.7 & 0.2 & 0.8 & 1.0 & 0.0 & 18.4 \\
Low-level policy (Cosmos-3) & 61.0 & -- & 1.4 & 9.5 & 7.0 & 1.6 & 12.2 & 0.7 & 0.1 & 0.9 & 1.0 & 0.0 & 26.5 \\
\tiptop{} & 77.6 & 46.7 & 0.9 & 2.8 & 6.5 & 0.8 & 0.7 & 1.4 & 0.0 & 0.9 & 0.8 & 0.0 & 16.2 \\
\capx{} & 57.6 & -- & 2.3 & 5.1 & 5.2 & 0.7 & 0.6 & 0.8 & 1.2 & 5.4 & 1.4 & 0.0 & 35.0 \\
Periodic trigger & 42.9 & 28.9 & 0.0 & 0.7 & 1.0 & 1.2 & 0.1 & 0.1 & 4.8 & 0.6 & 0.2 & 0.0 & 5.3 \\
Proprioceptive trigger & 41.1 & 26.3 & 0.1 & 0.7 & 1.2 & 1.3 & 0.4 & 0.1 & 5.5 & 0.5 & 0.2 & 0.0 & 4.7 \\
VoLo & 38.0 & 7.0 & 0.1 & 1.4 & 3.0 & 1.0 & 4.2 & 0.3 & 4.7 & 1.6 & 0.6 & 0.0 & 14.1 \\
Proprioceptive verifier & 43.9 & 10.3 & 0.1 & 1.9 & 4.2 & 1.2 & 4.8 & 0.6 & 5.1 & 2.4 & 0.8 & 9.2 & 3.3 \\
\harness{} & 18.4 & 13.3 & 0.0 & 0.6 & 0.8 & 0.1 & 0.1 & 0.2 & 0.9 & 0.9 & 0.6 & 0.0 & 1.0 \\
\gptpolicy{} & 21.3 & -- & 0.6 & 1.3 & 2.0 & 0.2 & 0.2 & 0.4 & 0.3 & 3.6 & 1.2 & 0.0 & 11.4 \\
\midrule
Privileged verifier & 36.7 & 12.4 & 0.0 & 1.0 & 3.3 & 1.1 & 1.3 & 0.4 & 5.1 & 1.7 & 0.5 & 7.6 & 2.4 \\
\bottomrule
\end{tabular*}
\end{table}

\paragraph{Orchestrator decisions.} The causes above describe what happened to the object. Table~\ref{tab:orchestrator-decisions} adds the orchestrator decision that made a loss final, from the logged subgoal switches, VLM verdicts, plans and tool calls. The simulator's own success predicate at the moment of each switch tells whether the subgoal left behind was satisfied. VoLo's plan steps and \harness{}'s goals are free text, which we match to the ground-truth subgoals by their object and destination; the match covers at least \decPlanCoverMin\% of the completed subgoals of every suite and system, which a plan must have named. A lost subgoal gets the first decision that applies: its orchestrator's plan never named it, \tiptop{}'s planner found no plan, \harness{} had stopped making tool calls, the orchestrator left it unsatisfied after judging it done (\emph{moved on before done}) or after a \emph{not done} verdict or a stall (\emph{deferred}) and never returned, or it was never issued because the orchestrator was still commanding a satisfied subgoal when the budget ran out (\emph{stuck}). A completed subgoal the policy picked up again counts as \emph{re-issued} when the orchestrator was commanding it again at the time, having issued it while it was satisfied or kept it in force for 45 steps after. The success predicate we read at each switch agrees with the one the orchestrator logged at every switch, and every early advance of the \emph{Proprioceptive verifier} follows a VLM \emph{done} verdict that the simulator contradicts. The privileged verifier's deferrals use privileged success and are not VLM errors. Figures~\ref{fig:failure-horizon} and~\ref{fig:failure-suites} draw the two largest decisions, \emph{no plan found} (\tiptop{} only) and \emph{moved on before done}, and group the rest as \emph{other mistake}.

\begin{table}[!ht]
\centering
\renewcommand{\arraystretch}{0.72}
\setlength{\tabcolsep}{2.6pt}
\caption{\textbf{Subgoals lost to an orchestrator mistake}, as a share of all subgoals (per suite, over all its tasks), over the episodes of Table~\ref{tab:subgoal-causes}. Each lost subgoal is attributed to the orchestrator decision that made its loss final; the rest of \emph{Lost} was lost while the robot carried out a correct command. A dash marks a system that logs no orchestrator decisions (low-level policy, \capx{} and \gptpolicy{}). The columns are the decisions defined above; \emph{Stopped} (\harness{} only) means the agent made no tool call in the last 30\% of the episode.}
\label{tab:orchestrator-decisions}
\scriptsize
\begin{tabular}{lccccccccc}
\toprule
System & Lost & Orch. & Not planned & No plan & Stopped & Moved on & Deferred & Stuck & Re-issued \\
\midrule
\multicolumn{10}{l}{\emph{SimplerEnv-Long}} \\
Low-level policy & 74.7 & -- & -- & -- & -- & -- & -- & -- & -- \\
\tiptop{} & 92.9 & 66.6 & 0.0 & 66.6 & 0.0 & 0.0 & 0.0 & 0.0 & 0.0 \\
\capx{} & 54.1 & -- & -- & -- & -- & -- & -- & -- & -- \\
Periodic trigger & 60.1 & 47.8 & 0.0 & 0.0 & 0.0 & 47.8 & 0.0 & 0.0 & 0.0 \\
Proprioceptive trigger & 54.2 & 34.2 & 0.0 & 0.0 & 0.0 & 33.2 & 0.0 & 1.0 & 0.0 \\
VoLo & 51.7 & 18.0 & 0.5 & 0.0 & 0.0 & 17.5 & 0.0 & 0.0 & 0.1 \\
Proprioceptive verifier & 49.0 & 4.0 & 0.0 & 0.0 & 0.0 & 2.2 & 0.0 & 1.7 & 0.0 \\
\harness{} & 41.8 & 26.5 & 0.1 & 0.0 & 4.0 & 22.3 & 0.0 & 0.0 & 0.2 \\
\gptpolicy{} & 16.7 & -- & -- & -- & -- & -- & -- & -- & -- \\
\midrule
Privileged verifier & 47.6 & 0.0 & 0.0 & 0.0 & 0.0 & 0.0 & 0.0 & 0.0 & 0.0 \\
\addlinespace
\multicolumn{10}{l}{\emph{\robotarena{}-Long}} \\
Low-level policy & 86.5 & -- & -- & -- & -- & -- & -- & -- & -- \\
\tiptop{} & 87.4 & 39.4 & 0.0 & 39.4 & 0.0 & 0.0 & 0.0 & 0.0 & 0.0 \\
\capx{} & 70.5 & -- & -- & -- & -- & -- & -- & -- & -- \\
Periodic trigger & 71.7 & 53.8 & 0.0 & 0.0 & 0.0 & 53.7 & 0.0 & 0.0 & 0.1 \\
Proprioceptive trigger & 71.1 & 38.6 & 0.0 & 0.0 & 0.0 & 38.2 & 0.0 & 0.4 & 0.0 \\
VoLo & 68.2 & 16.4 & 4.3 & 0.0 & 0.0 & 12.1 & 0.0 & 0.0 & 0.0 \\
Proprioceptive verifier & 66.3 & 7.0 & 0.0 & 0.0 & 0.0 & 2.6 & 0.1 & 4.4 & 0.0 \\
\harness{} & 49.8 & 31.8 & 3.3 & 0.0 & 5.8 & 21.7 & 0.0 & 0.0 & 1.0 \\
\gptpolicy{} & 16.6 & -- & -- & -- & -- & -- & -- & -- & -- \\
\midrule
Privileged verifier & 67.3 & 0.0 & 0.0 & 0.0 & 0.0 & 0.0 & 0.0 & 0.0 & 0.0 \\
\addlinespace
\multicolumn{10}{l}{\emph{RoboLab-Long}} \\
Low-level policy & 43.4 & -- & -- & -- & -- & -- & -- & -- & -- \\
Low-level policy (Cosmos-3) & 61.0 & -- & -- & -- & -- & -- & -- & -- & -- \\
\tiptop{} & 77.6 & 46.7 & 0.0 & 46.7 & 0.0 & 0.0 & 0.0 & 0.0 & 0.0 \\
\capx{} & 57.6 & -- & -- & -- & -- & -- & -- & -- & -- \\
Periodic trigger & 42.9 & 28.9 & 0.0 & 0.0 & 0.0 & 28.9 & 0.0 & 0.0 & 0.0 \\
Proprioceptive trigger & 41.1 & 26.3 & 0.0 & 0.0 & 0.0 & 26.3 & 0.0 & 0.0 & 0.0 \\
VoLo & 38.0 & 7.0 & 0.0 & 0.0 & 0.0 & 7.0 & 0.0 & 0.0 & 0.1 \\
Proprioceptive verifier & 43.9 & 10.3 & 0.0 & 0.0 & 0.0 & 5.4 & 3.8 & 1.0 & 0.1 \\
\harness{} & 18.4 & 13.3 & 0.0 & 0.0 & 8.5 & 4.6 & 0.0 & 0.0 & 0.2 \\
\gptpolicy{} & 21.3 & -- & -- & -- & -- & -- & -- & -- & -- \\
\midrule
Privileged verifier & 36.7 & 12.4 & 0.0 & 0.0 & 0.0 & 0.0 & 12.3 & 0.1 & 0.0 \\
\bottomrule
\end{tabular}
\end{table}

\paragraph{Per-benchmark findings.} The three suites fail in visibly different ways (Figure~\ref{fig:failure-suites}). Misplacement is one example: on the WidowX suites, \fateNearMissWx\% of misplaced objects lie on their correct destination without credit, while on RoboLab-Long \pifive{} mostly picks the wrong container.
\begin{itemize}\itemsep0em
    \item \emph{SimplerEnv-Long.} \tiptop{} finds no plan for \fateNoPlanTiptopSe\% of its subgoals.
    \item \emph{\robotarena{}-Long.} The hardest suite: every system except \harness{} and \gptpolicy{} loses at least \fateLostVerifierRa\% of its subgoals, against \fateLostHarnessRa\% for \harness{} and \fateLostGptRa\% for \gptpolicy{}. The low-level policy misses the grasp on \fateMissFlatRa\% of its subgoals, and the \emph{Proprioceptive verifier} and privileged verifier never command \fateNeverCmdRa\% because they spend the budget retrying earlier subgoals.
    \item \emph{RoboLab-Long.} The Franka almost never misses a grasp (at most \fateMissCapxRl\% of subgoals for any system), so its physical losses come after the grasp. Without an orchestrator, \pifive{} sets \fateWrongFlatRl\% of its subgoals on the wrong container, and under the low-level policy alone and every orchestrator together \fateWrongDestPi\% of its misplaced objects land in the wrong container, against \fateWrongDestHarness\% for \harness{}, which locates each object with a VLM. Orchestrated configurations undo more of their completed subgoals, and the policy picks \fateRegraspOrchRl\% of those back up, against \fateRegraspFlatRl\% under the low-level policy alone. \capx{} never touches \fateOutOfTimeCapxRl\%, and Cosmos-3 loses its grip five times as often as \pifive{} (\fateSlipCosmosRl\% vs.\ \fateSlipPiRl\%). \harness{} loses only \fateLostHarnessRl\%.
\end{itemize}

\paragraph{Measurement precision and caveats.} Classifying precise placement versus near-target placement relies on oriented bounding box predicates in the simulator. Validating against logged axis-aligned extents reproduces the simulator's per-placement verdict for \faGeomAgreement\% of \faGeomGoals{} sampled subgoals. Disagreements go both ways (\faGeomFalsePos\% credited without the logged geometry placing the object, \faGeomFalseNeg\% placed by the geometry without credit). The second kind is what we label slightly off target, so part of those may be the geometry's error rather than the robot's.

\paragraph{Online verification accuracy.} Deployed online against simulator ground truth, the visual verifier achieves \faVerifierPrecision\% precision and \faVerifierRecall\% recall across \faVerifierCalls{} calls (\faVerPrecSe\%/\faVerRecSe\% on SimplerEnv-Long, \faVerPrecRa\%/\faVerRecRa\% on \robotarena{}-Long and \faVerPrecRl\%/\faVerRecRl\% on \robolab{}-Long). A missed completion makes the orchestrator keep commanding a subgoal that is already done, which uses up the budget: on \robotarena{}-Long, this leaves \decStuckVerifierRa\% of the verifier's subgoals never issued (Table~\ref{tab:orchestrator-decisions}).

\paragraph{Failures with every subgoal completed.} The subgoal view cannot see one failure: an episode that completes every subgoal but breaks the task's ordering or exclusion rule. This is \faConstraint\% of the \faFailures{} failed episodes of the ten systems of Table~\ref{tab:comp-overall}, and \faConstraintOrdered\% of the failed episodes of ordered tasks.

\end{document}